%% file: arxiv.tex
\documentclass[final]{cvpr}

\usepackage{times}
\usepackage{epsfig}
\usepackage{graphicx}
\usepackage{amsmath}
\usepackage{bbm}
\usepackage{amssymb}
\usepackage{newtxmath}
\usepackage{multirow, boldline, arydshln}
\usepackage{nopageno}

\usepackage[pagebackref=true,breaklinks=true,colorlinks,bookmarks=false]{hyperref}

\def\cvprPaperID{829} 
\def\confYear{CVPR 2021}

\begin{document}
\input{TEX/title.tex}
\input{TEX/abs.tex}
\input{TEX/body.tex}

{\small
\bibliographystyle{ieee_fullname}
\bibliography{egbib}
}
\clearpage
\input{TEX/appendix.tex}

\end{document}

%% file: TEX/title.tex
\vspace{-5mm}
\title{Learning to Track Instances without Video Annotations}
\vspace{-5mm}
\author{Yang Fu$^{1\ast}$\thanks{$\ast$ This work was done while Yang Fu was a research intern at NVIDIA}, Sifei Liu$^{2}$, Umar Iqbal$^{2}$, Shalini De Mello$^{2}$\\ Humphrey Shi$^{1,3\dagger}$\thanks{$\dagger$ corresponding author}, Jan Kautz$^{2}$ \\
$^{1}$University of Illinois at Urbana-Champaign, $^{2}$NVIDIA, $^{3}$University of Oregon
}
\renewcommand\footnotemark{}
\maketitle

%% file: TEX/abs.tex
\begin{abstract}
Tracking segmentation masks of multiple instances has been intensively studied, but still faces two fundamental challenges: 1) the requirement of large-scale, frame-wise annotation, and 2) the complexity of two-stage approaches.
To resolve these challenges, we introduce a novel semi-supervised framework by learning instance tracking networks with only a labeled image dataset and \emph{unlabeled} video sequences. With an instance contrastive objective, we learn an embedding to discriminate each instance from the others. We show that even when only trained with images, the learned feature representation is robust to instance appearance variations, and is thus able to track objects steadily across frames. We further enhance the tracking capability of the embedding by learning correspondence from unlabeled videos in a self-supervised manner.
In addition, we integrate this module into single-stage instance segmentation and pose estimation frameworks, which significantly reduce the computational complexity of tracking compared to two-stage networks. 
We conduct experiments on the YouTube-VIS and PoseTrack datasets. Without any video annotation efforts, our proposed method can achieve comparable or even better performance than most fully-supervised methods\footnote{Project page:  \url{https://oasisyang.github.io/projects/semi-track/index.html}}.
\vspace{-2mm}
\end{abstract}

%% file: TEX/body.tex
\section{Introduction}
\input{TEX/1_intro.tex}
\section{Related Work}~\label{related}
\input{TEX/2_related.tex}

\section{Proposed Method}\label{method}
\input{TEX/3_method.tex}

\section{Experiments}\label{exp}
\input{TEX/4_experiment.tex}
\section{Conclusion}
\input{TEX/5_conclusion.tex}

%% file: TEX/1_intro.tex
\begin{figure}[t]
	\centering
	\includegraphics[width=0.45\textwidth]{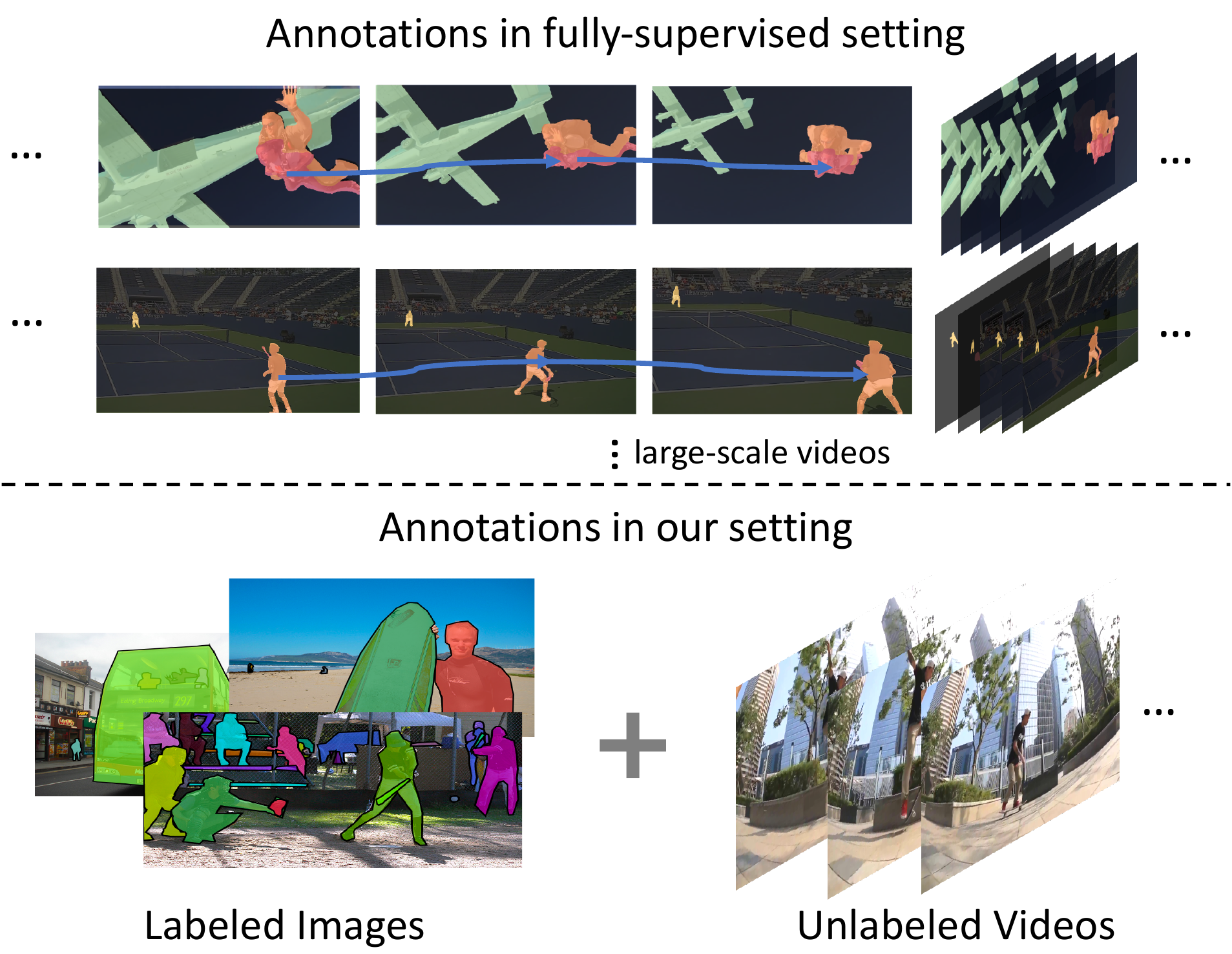}
	\caption{The annotations required for our proposed approach vs.~those for fully supervised approaches.}
	\label{fig:illus}
	\vspace{-3mm}
\end{figure}

In recent years, the vision community has rapidly improved the performance of instance segmentation at both the image and video levels as a core technique in autonomous driving.
The pipeline for segmenting instances from videos commonly includes: (i) segmentation on individual frame; and (ii) linking of each instance across frames for an entire video sequence.
Most existing approaches~\cite{Cao_SipMask_ECCV_2020, dong2019temporal, luiten2019video, yang2019video} employ fully-supervised learning that relies on dense annotations of instance segmentation masks and instance associations across video frames (see Fig.~\ref{fig:illus} top).
Since annotation of videos, especially in a per-frame manner requires excessive labor, the fully-supervised learning setting, however, becomes the major bottleneck for frame-wise video processing.


To reduce the dependence on labels, self-supervised tracking approaches have been developed to learn pixel-level video correspondences from large-scale unlabeled videos \cite{jabri2020space,li2019joint,wang2019learning}. The learned correspondences can be used to track any fine-grained attributes, \eg, segmentation masks, keypoints and textures, on a per-pixel basis. However, such self-supervised approaches aim to learn semantically-independent representations, \ie, they do not discriminate between object instances. Such approaches can be used for tracking only when ground truth attributes are annotated at keyframes, \eg, the 1st frame of any sequence \cite{davis_2017}; or when additional pre-trained instance segmentation models are provided.

In this paper, we consider a novel semi-supervised setting: we learn to track instances only with a labeled image dataset, and optionally, unlabeled video sequences. In other words, in addition to learning image-level instance segmentation, we also learn to associate instances across frames in a self-supervised manner. Our setting strikes a balance between the fully-supervised and the self-supervised ones.
With regards to it applications, our model can be seamlessly adapted and utilized for tracking objects on newly captured videos, \eg, traffic scene sequences during autonomous driving, without requiring any offline processing.

A typical way to learn tracking is to model instance association as a multi-class classification problem \cite{yang2019video}. Since we do not have the ground truth association labels, we instead learn a feature map that should be: (i) discriminative of different instances, and (ii) robust to appearance variation caused by motion of instances in videos.
Once learned, any object instance can be tracked by utilizing its feature embedding to search for the most similar one in the next frame.
To learn it with only labeled images, we introduce an Instance Contrastive (IC) objective defined densely on the embedding map. This objective encourages the pixel-level feature embedding to be consistent when being sampled from the same instance, while being less consistent for different ones.
In addition, we optimize a Maximum Entropy (ME) regularization to enforce that each instance, on being matched to others, exhibits a uniform distribution. With this constraint, when a new object enters a sequence, the model can easily detect it by comparing it with all existing instances
, and thus assign it a new instance label. 

In addition to using labeled images, we also discover when leveraging unlabeled videos, tracking performance can be further improved via self-supervised learning. In this work, we choose to learn self-supervised video correspondences. Specifically, we adopt a cycle-consistency loss by maximizing the likelihood of pixels returning to their original location on being propagated forward and backward along a stack of frames \cite{jabri2020space}. Since the feature embedding is utilized to construct the cross-frame affinity for propagation, it can be implicitly enhanced by enforcing this objective. Intuitively, video correspondence learning improves tracking performance by potentially encouraging the network to ``see’’ more instance appearance variations in time.

To further mitigate the data distribution shifts between labeled images, unlabeled videos, and testing videos, we introduce a self-supervised test-time adaptation strategy. Inspired by~\cite{sun2020test}, we enhance the model's tracking capability by keeping the self-supervised objective at the inference stage, and adapting it to any particular input sequence.

Instead of learning an independent network that separately produces the feature embedding for tracking, we integrate it as a head in to a bottom-up instance segmentation framework, \eg, SOLO~\cite{wang2019solo}. With labeled images, we jointly train the instance segmentation and the feature embedding parts of the network, enriching the original network with the new function of tracking.
We note that in addition to introducing a semi-supervised setting, we are also 
propose a bottom-up framework for tracking masks of multiple instances.
Finally, we also show that similar approaches can be generalized to the task of multiple human pose tracking, when building on top of a bottom-up human pose estimation network~\cite{wei2020point}. In summary, we conclude our contribution as the following:
\begin{itemize}
    \vspace{-2mm}
    \item A novel semi-supervised setting that can largely reduce the effort of labelling large-scale video datasets.
    \vspace{-5mm}
    \item An Instance Contrastive loss equipped with Maximum Entropy regularization to learn a feature embedding capable of tracking with only labeled images.
    \vspace{-2mm}
    \item A self-supervised video correspondence learning method that further improves tracking performance by leveraging unlabelled videos.
    \vspace{-2mm}
    \item Extensive experiments demonstrate that the proposed method performs on par if not better than most state-of-the-arts approaches, for both the video instance segmentation and pose tracking tasks.
\end{itemize}

%% file: TEX/2_related.tex
{\bf Video Instance Segmentation} is the joint task of detection, segmentation and tracking of object instances in videos. MaskTrack-RCNN~\cite{yang2019video} is the first attempt to address the video instance segmentation problem. It proposes a large-scale video dataset named YouTube-VIS for benchmarking video instance segmentation algorithms. MaskTrack RCNN extends Mask RCNN~\cite{he2017mask} with an additional tracking branch and achieves object association by object embedding and other cues, \ie, position and category. In addition, several methods from the Large-Scale Video Object Segmentation Challenge~\cite{challenge} achieve impressive results with large quantities of external data and complex algorithmic pipelines~\cite{dong2019temporal, luiten2019video, wang2019empirical, fu2020compfeat}. However, all these mentioned approaches heavily depend on video annotations, and to the best of our knowledge, our method is the first attempt at video instance segmentation without any video annotations.

\begin{figure*}[t]
	\centering
	\includegraphics[width=0.95\textwidth]{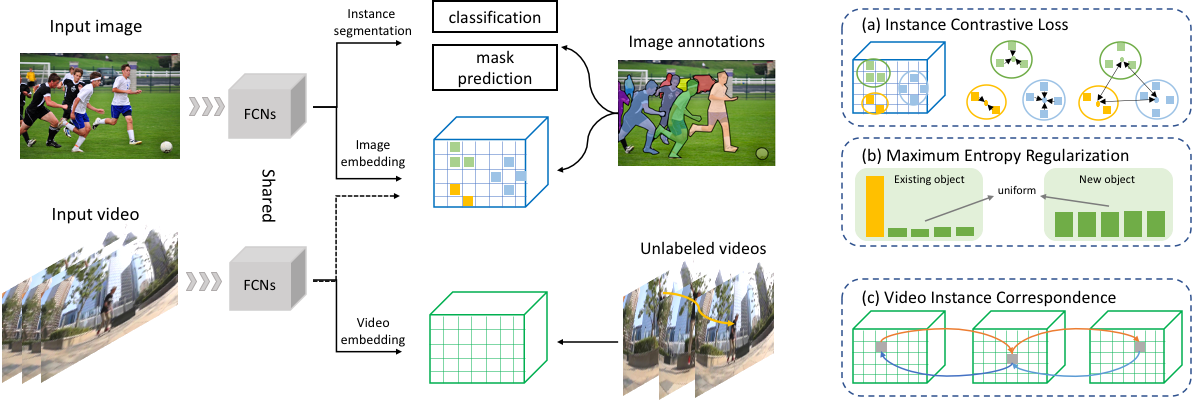}
	\caption{An overview of our proposed framework, which is built upon the bottom-up instance segmentation, \ie, classification and mask prediction heads. We propose image/video embedding heads. We train the image embedding branch with (a) an instance contrastive loss; (b) a maximum entropy regularization term using image annotations only; and train the video embedding branch via (c) self-supervised video correspondence learning. See Sec.~\ref{sec3.2}, \ref{sec3.3}, \ref{sec3.4} for more details.}
	\label{fig:frame}
	\vspace{-3mm}
\end{figure*}
{\bf Contrastive Learning} has recently received interest due to its success in self-supervised representation learning in the computer vision domain~\cite{chen2020simple, grill2020bootstrap, he2020momentum, oord2018representation}. These approaches follow a similar idea: pull together an anchor and a positive sample, meanwhile push apart the anchor from many negative samples. The positive sample is generated by a sets of data augmentations and the negative samples are randomly chosen from the mini-batch. The most widely used objective function is the InfoNEC~\cite{oord2018representation}, which encourages the mutual information between positive samples to be large while for negative samples, to be small. Recently, Khosla~\etal~\cite{khosla2020supervised} proposed a powerful contrastive loss that allows for multiple positives per anchor and proved its superior over traditional cross entropy under supervised setting. We borrow the similarity idea and propose the instance contrastive loss to effectively learn the instance embedding from image annotations.

{\bf Self-supervised Learning in Videos} aims to learn video-level representation by exploiting the frame redundancy. Some early work focus on representation learning from frames chronological order~\cite{misra2016shuffle, fernando2017self, wei2018learning}. For instance, Misra~\etal~\cite{misra2016shuffle} attempts to determine whether a sequence of frames from a video is placed in the correct temporal order, which can be used as a pretext task to improve some downstream tasks like action recognition. Besides, the colorization can be also treat as the supervision signal. 
Recently, several work~\cite{wang2019learning, li2019joint, jabri2020space} show that the cycle-consistency in time can be utilized as the supervisory signal for learning visual representations from video. The key idea is that: given any patch of an image at the first frame, then track it forward and backward, it should return its original position and the trajectory should be a circle. Different from the existing methods, the correspondence module in our framework focus on instance-level correspondence rather than pixel-level correspondence.

%% file: TEX/3_method.tex
We introduce our approach in this section. The overall framework is illustrated in Fig.~\ref{fig:frame}, which is built upon the bottom-up instance segmentation framework: SOLO~\cite{wang2019solo}. SOLO converts instance segmentation into two pixel-level classification tasks, \eg, instance classification and instance mask prediction. Specifically, the input image is divided into $s \times s$ grids, and if the instance's center falls into a grid cell, that grid cell is responsible for the above two tasks. We integrate a head that learns the proposed tracking embedding into it. The whole framework can be trained jointly and perform both instance segmentation in each frame, as well as tracking between frames. In this section, we mainly focus on how to learn the instance embedding.

We define the problem in Sec.~\ref{sec3.1}, and introduce how to utilize labeled images to learn a embedding for instance tracking through (i) an instance contrastive loss (IC) in Sec.~\ref{sec3.2}, and (ii) a maximum entropy (ME) regularization term in Sec.~\ref{sec3.3}.
We further improve its performance with unlabeled videos, as discussed in Sec.~\ref{sec3.4}.
\subsection{Problem Definition} \label{sec3.1}
In semi-supervised tracking, we have a labeled image dataset $\{X_\mathrm{Img}, Y_\mathrm{Img}\}$ where each individual image $x_\mathrm{Img}^{i}$ has its corresponding instance-level annotation $y_\mathrm{Img}^i$, including an instance category, a location (provided by a bounding box or a keypoint), and a mask.
Meanwhile, we also have an another video dataset $\{X_\mathrm{Vid}\}$ where no videos are annotated. The goal of semi-supervised tracking is to learn a feature representation 
that can effectively associate instances in $\{X_\mathrm{Vid}\}$ by only using the supervised information present in the image dataset.

\subsection{Instance Contrastive Loss} \label{sec3.2}

To learn a feature representation capable of tracking, we want to ensure that it is (i) discriminative of different instances, and (ii) consistent regardless of the variations present in videos. In addition, the feature representation should (iii) focus more on appearance rather than location, since objects can move in time.
Normally, such a feature embedding can be learned, \eg, via a side branch trained with labelled identities across frames as the supervision signal, as is evidenced in several existing works~\cite{Cao_SipMask_ECCV_2020, hwang2019pose, yang2019video}. 
%
Although no such annotations are accessible here, we find that instance-level annotation on images already provides sufficient information to achieve the above goals, \ie, to distinguish which pixels belong to the same instance, and which are from different ones. 
In the following, we propose to learn this via a contrastive learning framework.

%
We illustrate our network architecture in Fig.~\ref{fig:frame}: Other than the original classification and mask prediction heads in SOLO~\cite{wang2019solo}, we integrate our embedding network for tracking in parallel with them, as a third head. We equip it with the same sub-network structure and feature map resolution as the classification head at each level in FPN~\cite{lin2017feature} 
in order to make the network efficient and light-weight. We denote by $h(\cdot)$ the tracking head's mapping of the bottleneck representation to the tracking embedding, and by ${f}$ the output feature map.
%
%
We utilized the same grid-level instance labels that are assigned to the classification branch in SOLO and in several other works~\cite{kong1904foveabox, tian2019fcos, wang2019solo}: On the ground truth instance label images, we regard one pixel $(x,y)$ to belong to one instance if it falls into a range, controlled by scale factors $\varepsilon: (cx, cy, \varepsilon w, \varepsilon h)$, where  $(cx, cy)$, $w$, and $h$ denote the center of mass, width and height of the given ground truth mask. The instance assignment maps are down-sampled and rounded to fit the resolution of each level.
More details can be found in \cite{wang2019solo}. Similar to the classification head, the feature map is much smaller in size than the original image, e.g., $40\times 40$ at the most fine-grain level. We refer to each element as a grid cell.

With grid-level instance labels, we can directly extend the original formulation of contrastive learning~\cite{he2020momentum, tian2019contrastive}, based on InfoNCE~\cite{oord2018representation} to the instances of each image. With slight abuse of notation, for one query grid cell 
$x_q\in X$ with feature $f_q$ from the $i^\mathrm{th}$ instance $\Omega_i$, we sample another vector $f_p$ from the same instance as the positive sample, and all the other grid cells from different instance as the negative ones.
We thus optimize for the pixel $x_q$:
\vspace{-2mm}
\begin{equation}
    \mathcal{L}_q = -\log \frac{\exp(f_p^\top\cdot f_q)}{\sum_{k\in \Omega_{\bar{i}}}\exp(f_k^\top\cdot f_q)}, \quad p,q \in \Omega_i
    \label{eq:nce}
\end{equation}
where $\Omega_{\bar{i}}$ is the set of cells from all the other instances $\bar{i}$.
However, we found that \eqref{eq:nce} does not perform well in our case due to the highly long-tailed distribution of instances w.r.t.\ to their number of pixels. \textit{E.g.}, smaller instances will be insufficiently trained due to less positive samples.

\vspace{-2mm}
\paragraph{Center-Contra Losses.} We address the above issue by proposing a novel form of the loss: a combination of center and contrastive (Center-Contra) losses. We obtain the center representation $C_i$ of an instance $i$ by averaging all embedding features assigned with this instance, as $C_i = \frac{1}{N_i}\sum_{q\in \Omega_i}f_q$. 
Here $N_i$ represent the number of grid cells in $\Omega_i$. To force the embedding feature vectors of the same instance to be similar, we introduce the center loss that minimizes the L1 distance:
\begin{equation}
\begin{aligned}\label{eq:pull}
\mathcal{L}^\mathrm{center}_i = \sum_{q\in \Omega_i} \| C_i - f_q \|_1.
\end{aligned}
\end{equation}



Meanwhile, the embedding of different instances also need to be distinct from each others in order for the embedding to have a strong discriminative ability. Thus, we propose a contrast term by pushing the center representation of all the instances $\{C_i|i\in [1,K]\}$ further apart, where $K$ is the number of instances in an image.
In particular, we compute a dense similarity matrix:
\vspace{-2mm}
\begin{equation}
\label{eq:aff}
S(i, j) = \frac{\exp(C_i^\top\cdot C_j)}{\sum_{k=0}^K \exp(C_i^\top\cdot C_k)},
\end{equation}
To push apart instances, we need to encourage the elements on the diagonal of the matrix $S_{i,i}$ to be larger than the other off-diagonal elements $S_{i, j}, \forall j \neq i$. Thus, we maximize the self-matching likelihoods, where $\mathbf{CE}$ is the cross-entropy loss and $I$ is the identity matrix:
\vspace{-2mm}
\begin{equation}
\begin{aligned}\label{eq:contra} 
\mathcal{L}^\mathrm{contra} = \mathbf{CE}(S, I).
\end{aligned}
\end{equation}
Finally, we enforce IC losses by summing up the center losses of all instances, and combining them with the contrast term:
\vspace{-2mm}
\begin{equation}
    \mathcal{L}^\mathrm{IC} = \sum_{i=0}^{K}\mathcal{L}_{i}^\mathrm{center} + \lambda \mathcal{L}^\mathrm{contra}.
    \label{eq:IC}
\end{equation}
Compared to utilizing individual feature vectors, contrastive loss based on the center embedding in \eqref{eq:contra} effectively avoids the issue of highly-imbalanced size of instances.

\vspace{-2mm}
\paragraph{Tracking an Instance via the Embedding.} Given the learned embedding for tracking, we utilize the $\{C_i|i\in [1,K]\}$ as the prototype representations of instances to perform tracking, \ie, grid cells of the next frame are directly classified into $K$ classes by comparing against these prototypes through a softmax function, where the classification score indicates the instance associations. In addition, tracking can also be improved by leveraging information from the classification prediction branch, which is further discussed in Sec.~\ref{sec:eval}. 


\subsection{Maximum Entropy Regularization} \label{sec3.3}
\begin{figure}[t]
	\centering
	\includegraphics[width=0.48\textwidth]{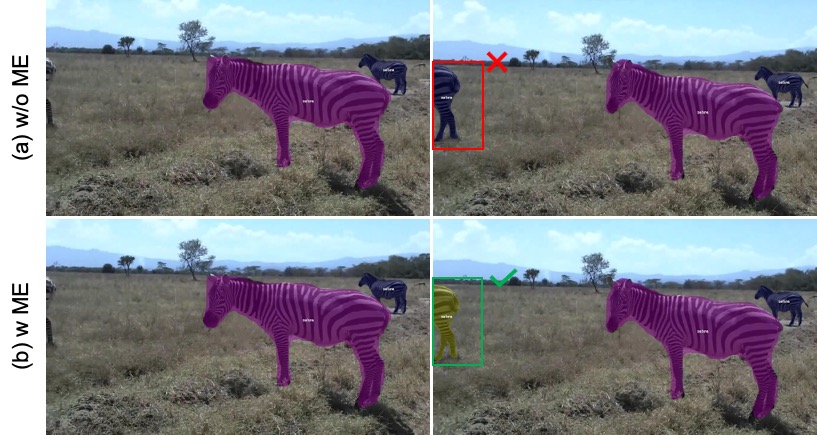}
	\caption{An illustration of failure case when a new object appears and the effectiveness of maximum entropy (ME) regularization. Row (a) and (b) are results without and with ME regularization. Best viewed in color and zoom in to see details.}
	\label{fig:fail}
	
\end{figure}

So far, our tracking approach is based on the assumption that any instance in the current frame also exists in the previous frame. It doesn't consider newly emerged objects. We observe that with the tracking procedure described in Sec.~\ref{sec3.2}, a new object is highly likely to exhibit a peaky  distribution for its similarity score when matched to all the instances in the previous frame. Consequently it will be incorrectly matched to an existing instance, \eg, see the dark black zebra in Fig.~\ref{fig:fail}, top.

To resolve this issue, we apply entropy maximization so that the model performs out-of-distribution detection~\cite{vyas2018out} -- which means ideally, a new object should not bear more resemblance to any of one existing instances in comparison to the others. Since we do not have video labels that annotate new objects in time, we exploit the existing image's labels by adding a ME term for all the instances in it: we increase the entropy measured for the similarity between the center embedding of each instance and all other instances. Reusing the similarity matrix $S$, the entropy is computed as: 
\vspace{-2mm}
\begin{equation}
    H=-\sum_i^K \sum_{j \neq i}^K S(i, j) \log (S(i,j)),
    \label{eq:ME}\vspace{-2mm}
\end{equation}
where $K$ is the number of instances and $S(i, j)$ is the probability of matching instance $i$ to $j$.
High entropy $H$ indicates uniform output probability. When enforced together with the IC term \eqref{eq:IC}, it encourages instances to be equally dissimilar to all other instances, see Fig.~\ref{fig:frame} (b).

When a new object is successfully detected, we follow the tracking strategy described in the previous section by comparing it to the existing $K$ objects (already detected in previous frames). Via ME, we enforce the similarity scores to be equally low for all existing instances as shown in Fig.~\ref{fig:frame} (b). Thus, it is easy to assign a new identity to a new object by setting a proper threshold such that all similarity scores are below it. Fig.~\ref{fig:fail} shows a comparison of the model without and with the proposed ME term.

\subsection{Self-supervised Video Correspondence} \label{sec3.4}

Although large-scale videos are hard to label, they are easy to acquire. Can we further improve our model by leveraging these videos? The answer is positive, but non-trivial: On the one hand, with a tracking embedding trained only with image collections, there is no guarantee that tracking of instances can be continuous and coherent over time. However, with videos we do not know the ground truth instance correspondences. Moreover, with videos we also need to address the domain gap that usually exists between image and videos.

To this end, we leverage self-supervised video correspondence learning~\cite{jabri2020space, li2019joint, wang2019learning} to regularize tracking of the predicted instances. We determine the valid grid cells (\ie, those belonging to any instances) through non-maxima suppression (NMS) on the matches with higher classification response (see inference in \cite{wang2019solo}) for more details). On the tracking embedding, we learn grid cell-level video correspondences in the valid grid cells only, \ie, within the regions containing instances, through a cycle consistency loss~\cite{jabri2020space,wang2019learning}. In detail, given a group of frames randomly sampled from one sequence, we compute cross-instance affinity $A \in \mathbb{R}^{P \times Q}$, where $P, Q$ are the numbers of valid instances in a pair of frames. Let $A_t^{t+1}(i, j)$ be the transition probability of the $i^\mathrm{th}$ instance at time $t$ being matched with the $j^\mathrm{th}$ instance at time $t+1$. We can formulate long-range correspondences by the chain rule:
\vspace{-2mm}
\begin{equation}
\begin{aligned}\label{eq:aff_cyc}
\bar{A}_{t}^{t+k} = \prod_{i=0}^{k-1} A_{t+i}^{t+i+1}.
\end{aligned}\vspace{-2mm}
\end{equation}
If we reverse this sequence and track the instances from $t+k$ to $t$, ideally, the $i^\mathrm{th}$ instance should return back to its original position in the first frame. Thus we have the following objective, where $I$ is the identity matrix:
\begin{equation}
\begin{aligned}\label{eq:cyc}
\mathcal{L}^\mathrm{cyc} = \mathbf{CE}(\bar{A}_{t}^{t+k}\bar{A}_{t+k}^{t}, I).
\end{aligned}
\end{equation}
We note that differently from~\cite{jabri2020space}, which needs to maintain a group of large affinity matrices (\ie, $N\times N$ where $N$ is the number of pixels), the dimensions of affinity in our case (\ie, number of valid grids) is much smaller and the module is more efficient.

In addition, we observe that when a domain gap between image and video datasets exists, \eg, COCO~\cite{lin2014microsoft} vs YouTube-VIS~\cite{yang2019video}, adopting the video objective~\eqref{eq:cyc} on the tracking embedding does not ensure convergence due the shared normalization. Therefore, we instead learn a video embedding using~\eqref{eq:cyc} with an additional head (see Fig.~\ref{fig:frame}, the dashed link is not used when domain gap exists). We found that with a shared backbone network, both the image embedding and the video embedding can be improved by self-supervised learning. During inference, we utilize the image embedding for tracking due to its superior performance.

\subsection{Test-time Adaptation}
Inspired by \cite{sun2019deep}, we can further mitigate the distribution shifts during the test-time: We still adopt the video embedding branch, and update the model weights by keeping the video correspondence loss in an online adaptation fashion. We find that the best performance can be achieved by updating the weights from the viedo correspondence branch 
as well as the backbone network (including the FPN Head~\cite{lin2017feature}).

%% file: TEX/4_experiment.tex
We evaluate our proposed method on two different instance-level tracking problems: video instance segmentation and multi-person pose tracking. 
\subsection{Datasets and Evaluation Metrics} \label{sec:eval}
\begin{table*}\setlength{\tabcolsep}{4pt}
\centering
\footnotesize
\begin{tabular} {l|c|c|ccc|c|c|c|c|c}
\hline
\multirow{2}{*}{Methods} & Video & With & Contrastive  & Max  & Video & \multirow{2}{*}{AP} & \multirow{2}{*}{AP$_{0.5}$} & \multirow{2}{*}{AP$_{0.75}$} & \multirow{2}{*}{AR$_1$} & \multirow{2}{*}{AR$_{10}$} \\ 
& Annotations & Embed & Loss & Entropy & Correspondence & & &  & & \\ \hline
MaskTrack-RCNN~\cite{yang2019video}& \checkmark & \checkmark &  &  &  & 29.0 & 47.5 & 32.2 & 28.7 & 32.4 \\ \hline
SOLO~\cite{wang2019solo} & & &  &  &  & 23.9 & 43.3 & 21.5 & 26.7 & 37.3 \\ 
SOLO-Track & & \checkmark & \checkmark &  &  & 28.4 & 50.0 & 30.4 & 27.6 & 34.4 \\
SOLO~Track & & \checkmark & \checkmark & \checkmark & & 29.7 & 52.8 & 29.9 & 30.7 & 34.9 \\
SOLO-Track & & \checkmark & \checkmark & \checkmark & \checkmark & {\bf 32.9} & {\bf 54.4} & {\bf 35.0} & {\bf 34.1} & {\bf 40.8} \\ \hline

\end{tabular}
\caption{Ablation study with different proposed components on YouTube-VIS validation set. The best results are highlighted in bold.}
\vspace{-3mm}
\label{ab:t1}
\end{table*}

\begin{table}\setlength{\tabcolsep}{5pt}
\centering
\footnotesize
\begin{tabular} {c|c|c|c|c|c}
\hline
\# frames & AP & AP$_{0.5}$ & AP$_{0.75}$ & AR$_1$ & AR$_{10}$ \\ \hline
2 & {\bf 32.9} & {\bf 54.4} & {\bf 35.0} & {\bf 34.1} & {\bf 40.8} \\
3 & 31.8 & 52.4 & 31.7 & 32.2 & 39.1 \\ 
4 & 30.9 & 51.6 & 30.9 & 31.7 & 38.4 \\ \hline
\end{tabular}
\caption{The performance of video instance segmentation with different number of frames in video correspondence model. The best results are highlighted in bold.}
\vspace{-5mm}
\label{ab2:frame}
\end{table}

{\bf YouTube-VIS~\cite{yang2019video}} is the first and largest dataset for video instance segmentation. In each video, objects with bounding boxes and masks are labeled manually every five frames and the identities cross different frames are annotated as well. Since only the validation set is available for evaluation, all results reported in this paper are evaluated on the validation set. It is important to note that for VIS, we only test on the videos whose categories overlap with COCO~\cite{lin2014microsoft}, which are 20 categories. We contacted the authors for the annotations of that sub validation set.

{\bf PoseTrack~\cite{andriluka2018posetrack}} is a large-scale benchmark for multi-person pose estimation and tracking. It contains challenging sequences of  people in dense crowds performing a wide range of activities. We conduct experiments only on PoseTrack 2018, where each person is annotated with 15 body joints, each one defined as a point and associated with a unique person id cross frames.

{\bf Evaluation Metrics}. 
For VIS, we use the metrics mentioned in~\cite{yang2019video}, which are average precision (AP) and average recall (AR) based on a spatio-temporal Intersection-over-Union (IoU) metric. 
For pose tracking, we evaluate our model via standard pose estimation~\cite{ruggero2017benchmarking} and tracking metrics~\cite{andriluka2018posetrack}, which are expressed by AP and multi-object tracking accuracy (MOTA), respectively. Unlike~\cite{girdhar2018detect, sun2019deep, wang2020combining}, we report MOTA along with its corresponding AP after post-processing videos. We apply post-processing to ignore some keypoints that are below a predefined confidence score. Note that it can lower the performance on AP but improve the performance on MOTA.

\vspace{-2mm}
\subsection{Implementation Details}
{\bf Training}. For both VIS and pose tracking, we first pre-train our model on the COCO dataset with the instance embedding head with the IC loss and ME regularization. In particular, we utilize SOLO and PointSetAnchor~\cite{wei2020point} as the base models for instance segmentation and pose estimation, respectively. The details of instance and keypoint embedding modules are described in supplementary materials.
Our model is implemented on MMDetection~\cite{chen2019mmdetection} and the whole framework is trained with 8 NVIDIA TITAN V100 GPUs until convergence.

{\bf Inference}. During evaluation, the testing video is processed frame by frame in an online fashion as described in~\cite{yang2019video}. More details can be found in supplementary materials.
To keep consistent with the previous approaches and improve the performance, we also apply a post-processing procedure introduced in~\cite{yang2019video}, which combines the initial prediction results with: detection confidence, bounding box IoU, category consistency, and similarity scores, etc. 
During the test-time training, each testing video is trained for 5 iterations with a sinlge NVIDIA TITAN V100 GPU.

\subsection{Ablation Study}
We conduct all ablation studies on the YouTube-VIS dataset. We believe that similar conclusions can also be drawn for pose tracking.

{\bf Baseline Model}. To the best of our knowledge, this is the first work to learn semi-supervised tracking using only image annotations, and hence it is important to establish a strong baseline model. In particular, we use MaskTrack RCNN~\cite{yang2019video} as the fully supervised baseline. It takes the pretrained MaskRCNN model and finetunes it on YouTube-VIS~\cite{yang2019video} with full video annotations, including instance categories, locations, masks and identities. The MaskTrack baseline is used to show how well our proposed semi-supervised method performs compared to fully supervised state-of-the-art methods. 
In addition, we also provide a bottom-up baseline based on SOLO, by training the task of instance segmentation without learning the tracking embedding. The objects are associated by spatial distance and category consistency.
It is clear that the SOLO baseline is less accurate than MaskTrack RCNN. 
The SOLO baseline is used to validate the effectiveness of each proposed component in our method.

{\bf Effectiveness of Instance Contrastive Loss}. To validate its effectiveness, we report the performance with, and without the embedding branch in Table~\ref{ab:t1}. With our proposed IC loss, the performance is improved by 4.5\%, 6.7\% and 8.9\% in AP, AP$_{0.5}$ and AP$_{0.75}$, respectively compared to the SOLO baseline. 
%
This improvement validates the previous claim that even only trained with labeled images, our method can learn discriminative representation with strong tracking capability.

{\bf Effectiveness of Maximum Entropy Regularization}. Besides strong distinguishing ability, a robust embedding also needs to discover new objects. However, as shown in Fig~\ref{fig:fail}, 
the embedding feature cannot distinguish between new and existing objects effectively by only using the IC loss. Thus, the maximum entropy (ME) regularization term is proposed to address this problem. 
%
As listed in Table~\ref{ab:t1}, the model with the ME regularization term can effectively boost performance on VIS. Specifically, it improves AP and AP$_{0.5}$ by 1.3\% and 2.8\%, and it achieves an AP of 29.7\%, which outperforms the fully supervised baseline of MaskTrack RCNN~\cite{yang2019video}.

{\bf Effectiveness of Video Correspondence}. 
We also show the effectiveness of self-supervisely learn video correspondence with unlabeled videos that are fairly cheap and easy to obtain. As listed in Table~\ref{ab:t1}, the proposed video correspondence model can improve performance significantly across all evaluation metrics. For instance, the gains in AP, AP$_{0.5}$ and AP$_{0.75}$ are 2.8\%, 1.6\% and 5.11\%, respectively. In addition, compared to the SOLO baseline, our final model improves the performance by 9.0\%, 11.1\% and 13.5\% for AP, AP$_{0.5}$ and AP$_{0.75}$, respectively. Furthermore, it also outperforms MaskTrack RCNN by a large margin. These improvements show that the video correspondence model can significantly enhance the tracking capability of our embedding representation.

{\bf Sequence Length}. So far we have validated all our proposed components. The video correspondence model especially brings a significant improvement, but the number of frames used to compute the cycle loss can affect its performance a lot. 
We can only perform the experiment with 2 to 4 frames due to limitations on GPU memory. From Table~\ref{ab2:frame}, it can be observed that the video correspondence model can achieve the best performance using only two frames. With increased number of frames, the performance on AP drops gradually from 32.9\% to 30.9\%. The degraded results may be caused by inclusion of noisy sampled with more frames. 
Since we do not have any annotations, we instead use category-level predictions to sample several positive instances. While the predictions are not exactly accurate, more frames can bring more noise, which leads to the worse performance. 

\begin{table}\setlength{\tabcolsep}{4pt}
\centering
\footnotesize
\begin{tabular} {l|c|c|c|c|c}
\hline
Methods & AP & AP$_{0.5}$ & AP$_{0.75}$ & AR$_1$ & AR$_{10}$ \\ \hline
\multicolumn{6}{c}{\bf Video + Image Annotations}\\ 
MaskTrack R-CNN~\cite{yang2019video} & 29.0 & 47.5 & 32.2 & 28.7 & 32.4 \\ 
SipMask~\cite{Cao_SipMask_ECCV_2020} & 24.1 & 42.0 & 26.0 & 26.2 & 28.6 \\ \cdashline{1-6}
\multicolumn{6}{c}{\bf Only Image Annotations}\\ 
Ours & 29.7 & 52.8 & 29.9 & 30.7 & 34.9 \\ 
Ours$^+$ & {\bf 32.9} & {\bf 54.4} & {\bf 35.0} & {\bf 34.1} & {\bf 40.8} \\
\hlineB{2.5}
\multicolumn{6}{c}{\bf After post-processing}\\ \hlineB{2.5}
 \multicolumn{6}{c}{\bf Video + Image Annotations}\\ 
IoUTracker+~\cite{yang2019video}& 29.4 & 48.5 & 30.6 & 32.1 & 34.2 \\
SeqTracker~\cite{yang2019video} &  31.8 & 52.2 & 35.8 & 32.2 & 34.4 \\
MaskTrack R-CNN~\cite{yang2019video} & 36.0 & 58.4 & {\bf 40.2} & 35.4 & 38.9 \\
SipMask~\cite{Cao_SipMask_ECCV_2020} & {\bf 37.7} & 57.8 & 38.0 & 37.4 & 40.3 \\ \cdashline{1-6}
\multicolumn{6}{c}{\bf Only Image Annotations}\\ 
Ours & 34.1 & 58.0 & 37.9 & 33.0 & 39.2 \\ 
Ours$^+$ & 37.4 & 59.7 & 39.1 & 36.4 & 43.8 \\ 
Ours$^*$ & {\bf 38.3} & {\bf 61.1} & 39.8 & 36.9 & {\bf 44.5} \\
\hline
\end{tabular}
\caption{Comparison of the our approach with the SOTA methods on the YouTube-VIS validation set. ``Ours" represents the model with instance embedding branch trained with IC loss and ME regularization. ``Ours$^+$" stands for the model with the video correspondence module as well. ``Our$^*$ is the model updated by test-time adaptation upon ``Ours$^+$". The best results are highlighted in bold.}
\vspace{-2mm}
\label{sota:vis}
\end{table}

\begin{table}
\setlength{\tabcolsep}{4pt}
\centering
\footnotesize
\begin{tabular} {cl|c|c|c|c|c|c}
\hline
&\multirow{2}{*}{Methods}  & \multicolumn{5}{c|}{MOTA} & \multirow{2}{*}{AP}  \\ 
& & Head & Shou & Wrist & Ankle & Total   \\ \hline
&  \multicolumn{7}{c}{\bf Video + Image Annotations}\\
\parbox[t]{2mm}{\multirow{4}{*}{\rotatebox[origin=c]{90}{Top down}}} & Miracle~\cite{yu2018multi}  & 68.8 & 73.5 & 61.2 & 56.7 & 64.0 & --  \\
& OpenSVAI~\cite{ning2018top} &  -- & -- & -- & -- & 62.4 & 69.7 \\
& LightTrack~\cite{ning2020lighttrack} & -- & -- & -- & -- & 64.6 & 72.4\\
& KeyTrack~\cite{snower202015} & -- & -- & -- & -- & \textcolor{red}{66.6} & 74.3 \\ \cline{2-8}
\parbox[t]{2mm}{\multirow{7}{*}{\rotatebox[origin=c]{90}{Bottom up}}} & MDPN~\cite{guo2018multi}  & 50.9 & 55.5 & 49.0 & 45.1 & 50.6 & 71.7 \\ 
&STAF~\cite{raaj2019efficient} & -- & -- & -- & -- & 60.9 & 70.4 \\
&MIPAL$++$~\cite{hwang2019pose} & 76.0 & 76.9 & 56.4 & 52.4 & 65.7 & \textcolor{red}{74.6} \\ \cdashline{2-8}
&  \multicolumn{7}{c}{\bf Only Image Annotations}\\ 
&Baseline & 64.9 & 70.9 & 56.3 & 55.0 & 62.0 & 69.2 \\
&Ours &  65.8 & 71.6 & 56.3 & 56.6 & 62.8 & 69.3  \\ 
&Ours$^+$ & 67.1 & 72.3 & 58.2 & 57.7 & 64.2 & 69.3 \\ 
&Ours$^{++}$ & 70.4 & 73.3 & 55.9 & 56.3 & \textcolor{blue}{64.7} & \textcolor{blue}{71.4}  \\ 
\hline
\end{tabular}
\caption{Comparison of our approach with the SOTA methods on the PoseTrack2018 validation set.``Baseline" associates poses only by the OKS metrics. ``Ours" and ``Ours$^{+}$" have the same definitions as Table~\ref{sota:vis}. ``Ours$^{++}$" has the same structure as the ``Ours$^+$" model, but is finetuned with the MPII data~\cite{andriluka20142d}. The best results on MOTA and AP for the methods with both image and video annotations and only image annotations are highlighted with red and blue color, respectively.}
\vspace{-5mm}
\label{sota:pose}
\end{table}

\begin{figure*}[t]
\begin{center}
\bgroup 
 \def\arraystretch{0.1} 
 \setlength\tabcolsep{0.5pt}
\begin{tabular}{ccccc}
\includegraphics[width=0.2\linewidth]{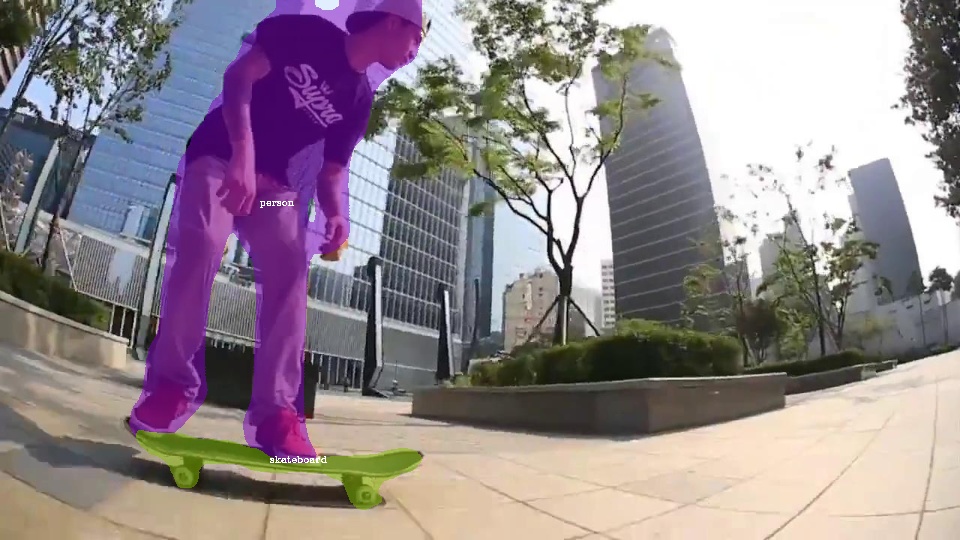} &
\includegraphics[width=0.2\linewidth]{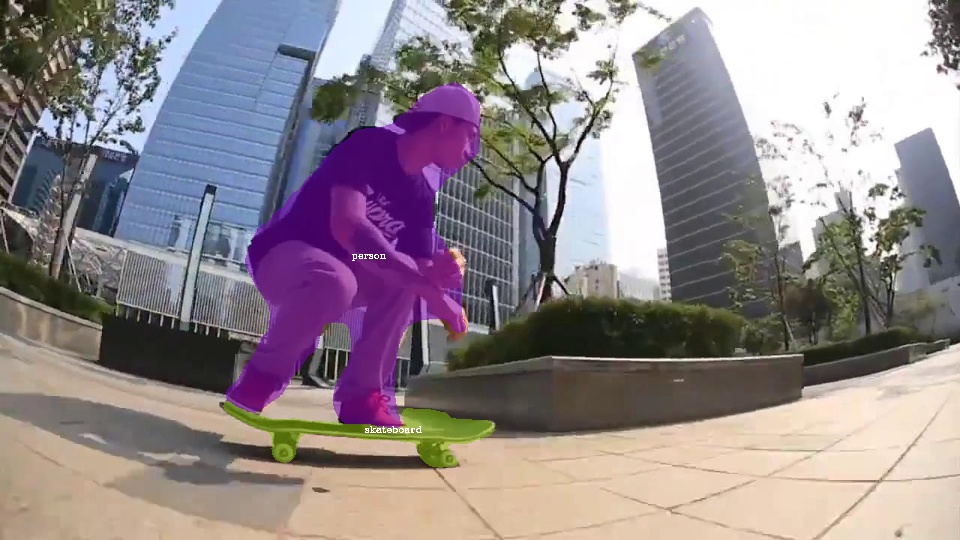} & 
\includegraphics[width=0.2\linewidth]{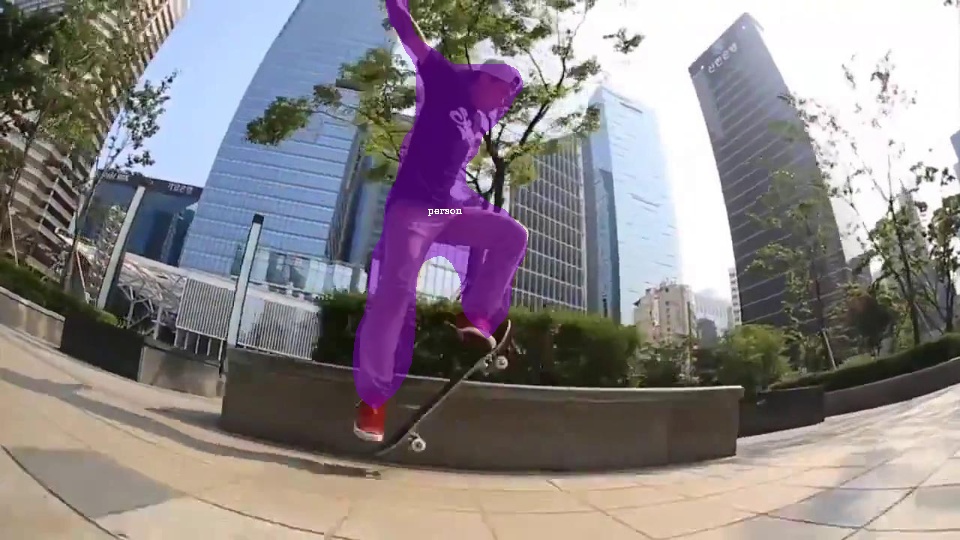} & 
\includegraphics[width=0.2\linewidth]{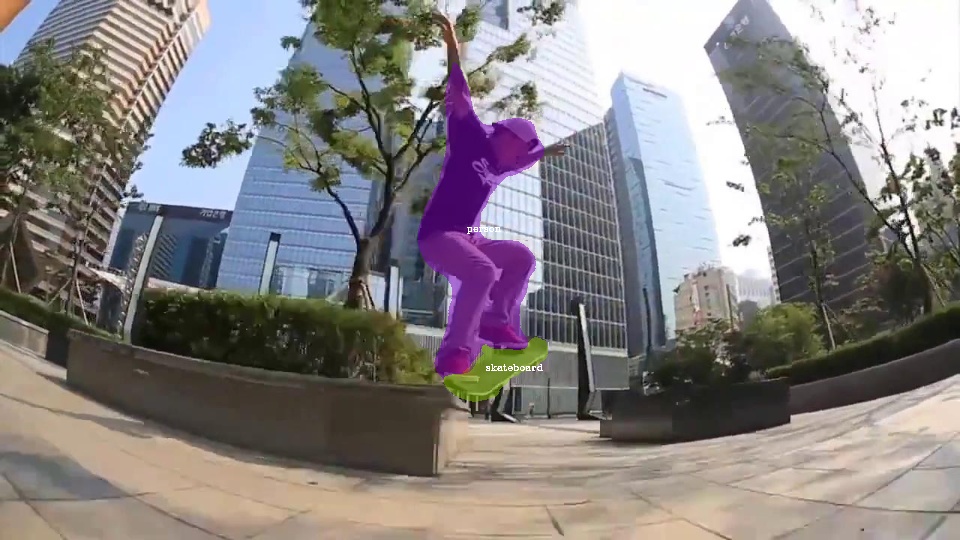} &
\includegraphics[width=0.2\linewidth]{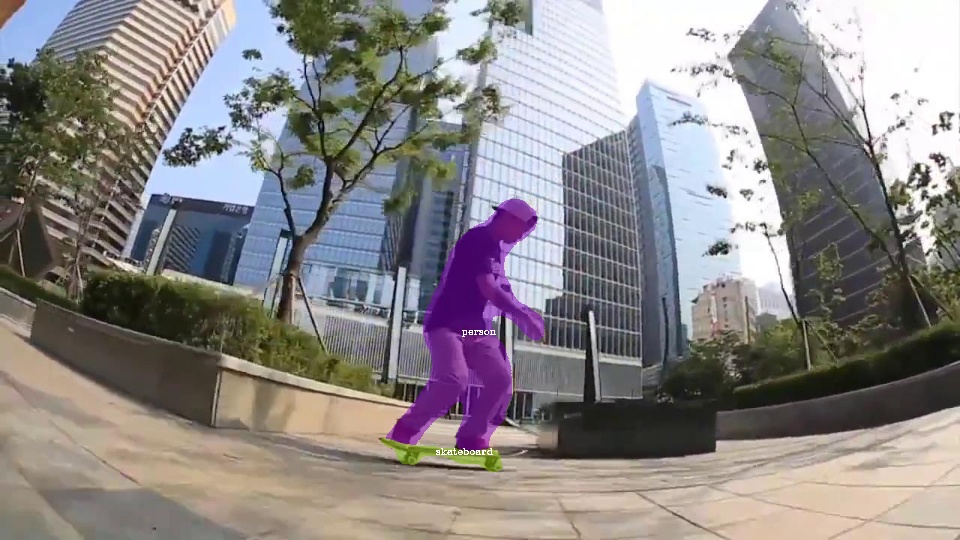} \\
\includegraphics[width=0.2\linewidth]{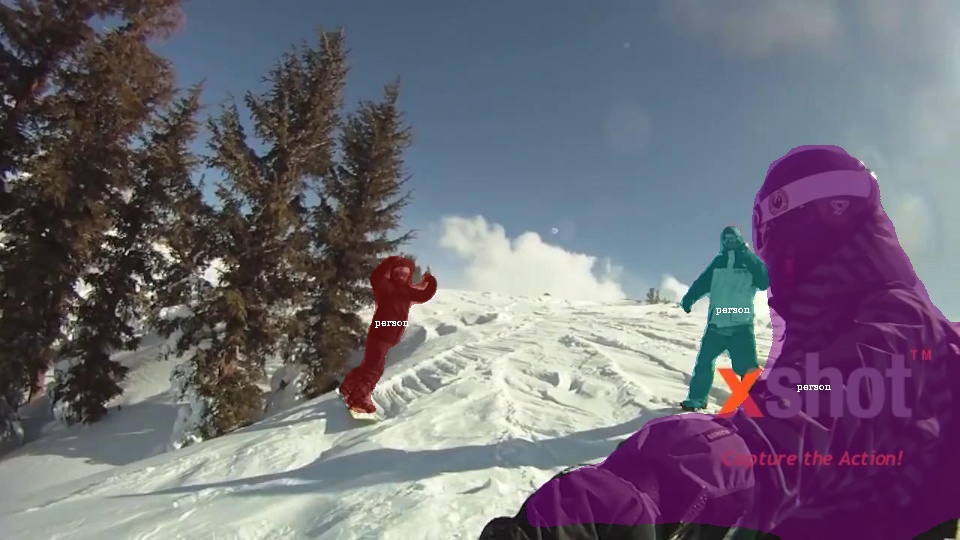} &
\includegraphics[width=0.2\linewidth]{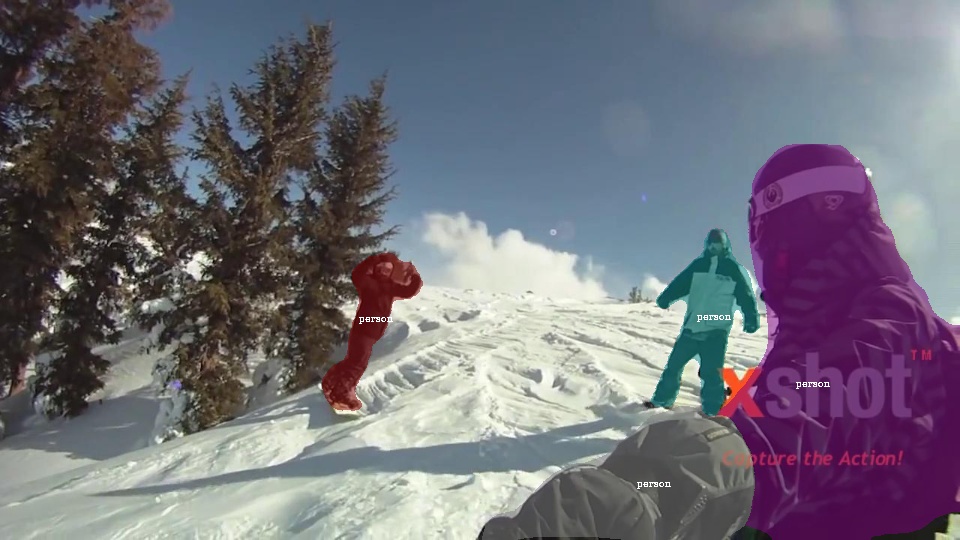} &
\includegraphics[width=0.2\linewidth]{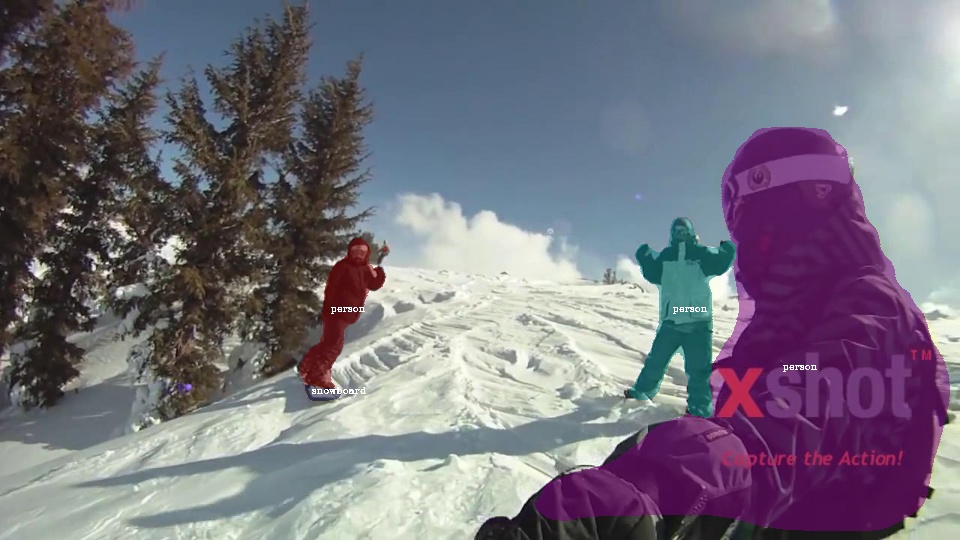} &
\includegraphics[width=0.2\linewidth]{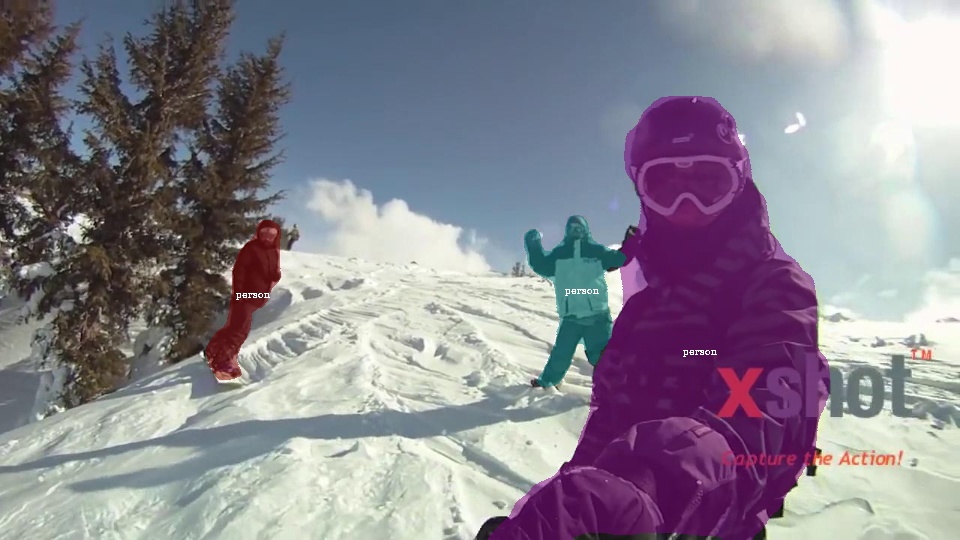} &
\includegraphics[width=0.2\linewidth]{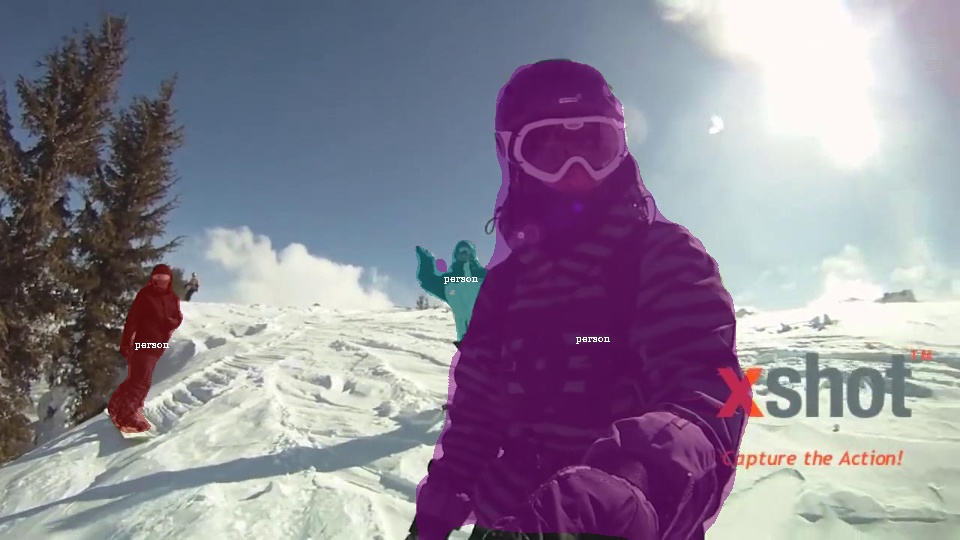} \\
\includegraphics[width=0.2\linewidth]{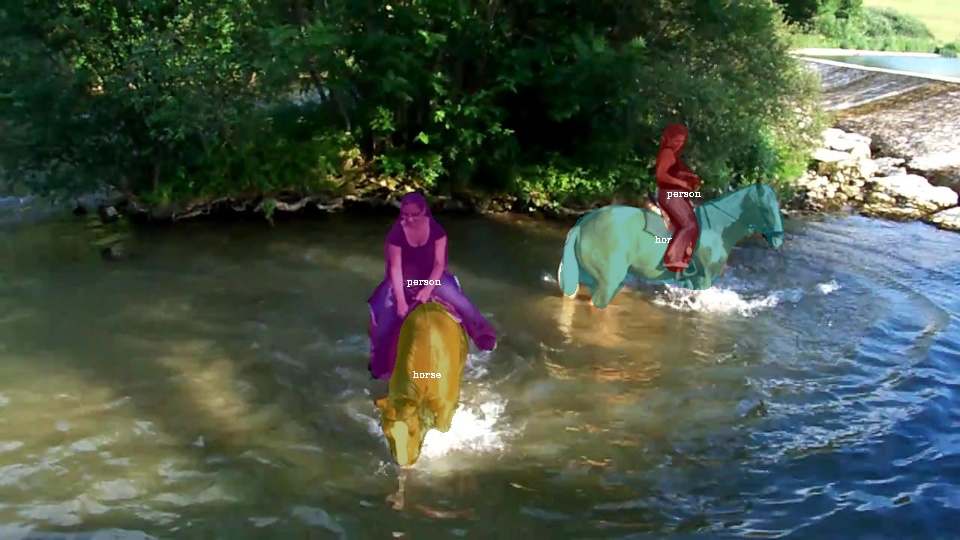} &
\includegraphics[width=0.2\linewidth]{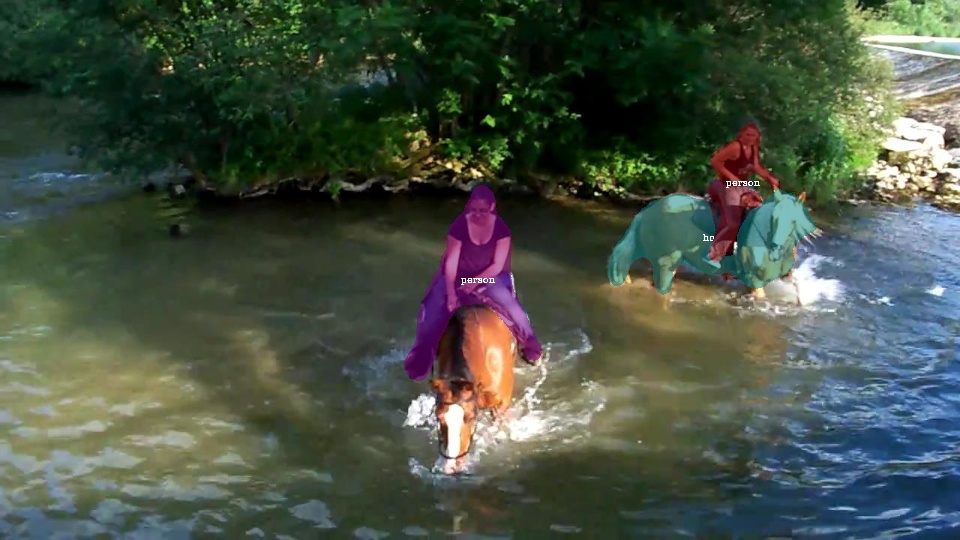} &
\includegraphics[width=0.2\linewidth]{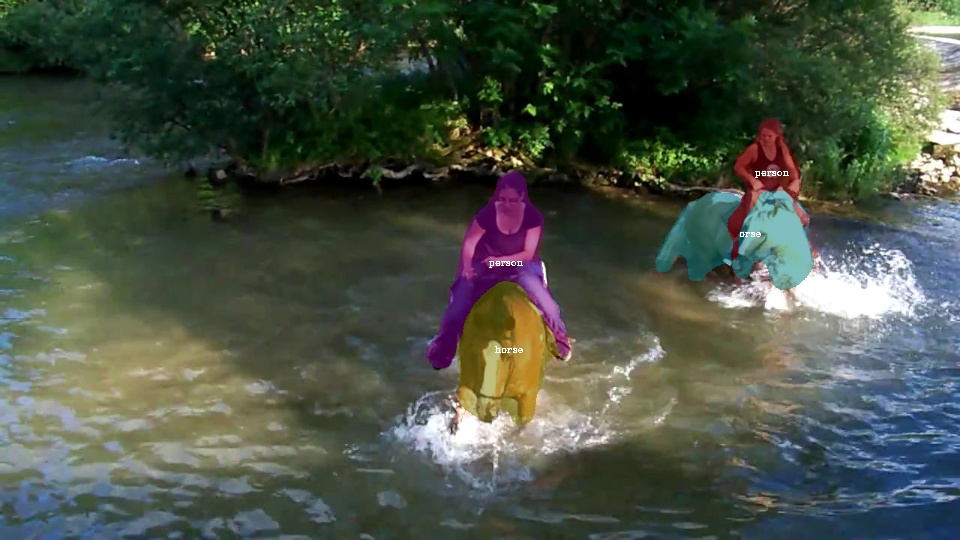} &
\includegraphics[width=0.2\linewidth]{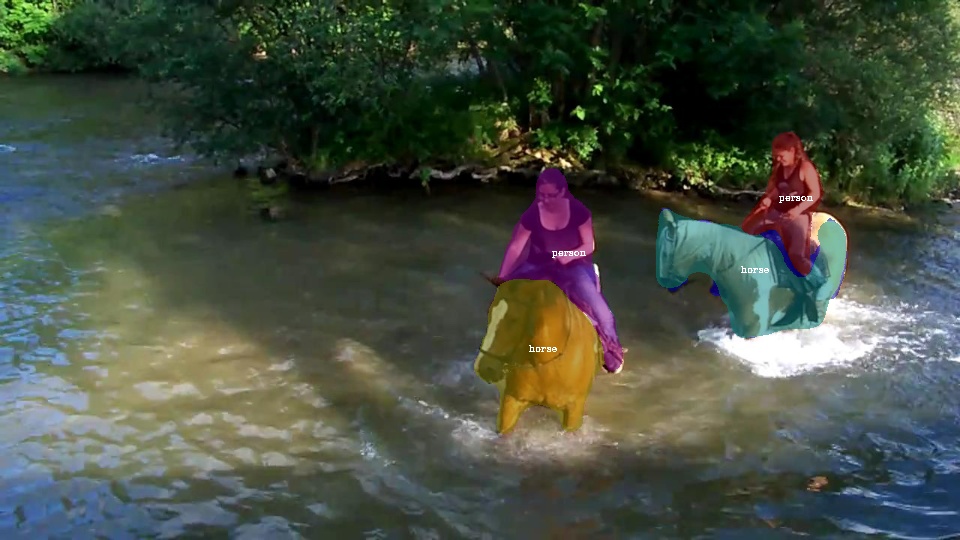} &
\includegraphics[width=0.2\linewidth]{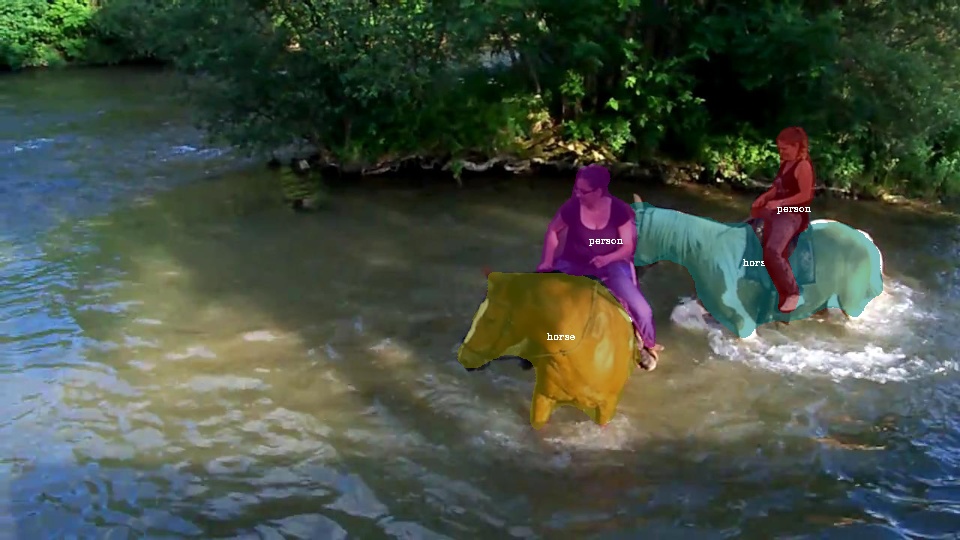} \\
\includegraphics[width=0.2\linewidth]{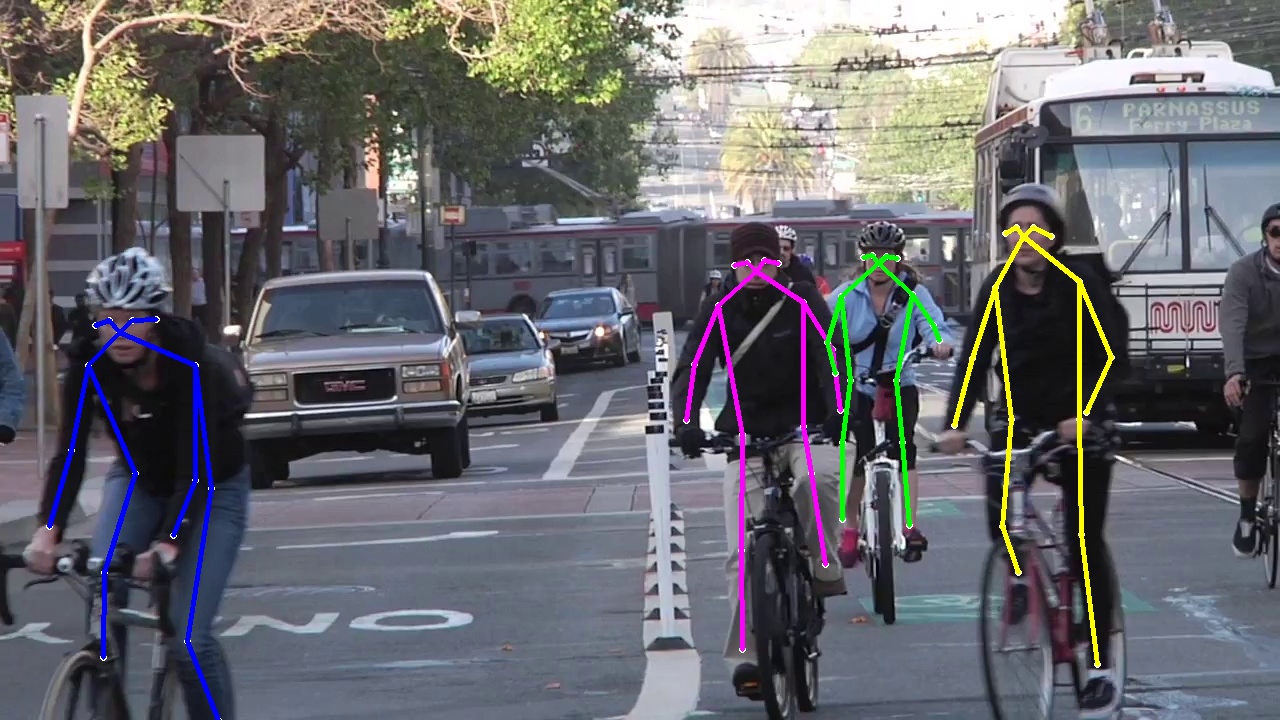} &
\includegraphics[width=0.2\linewidth]{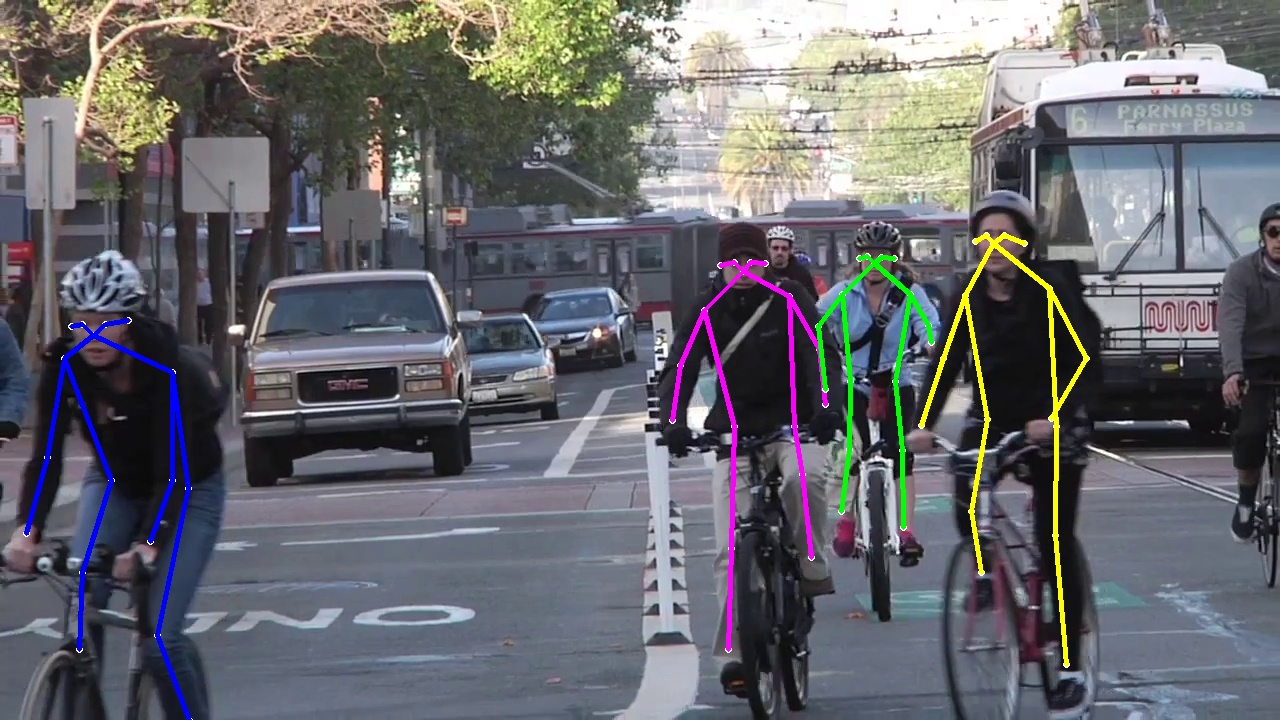} &
\includegraphics[width=0.2\linewidth]{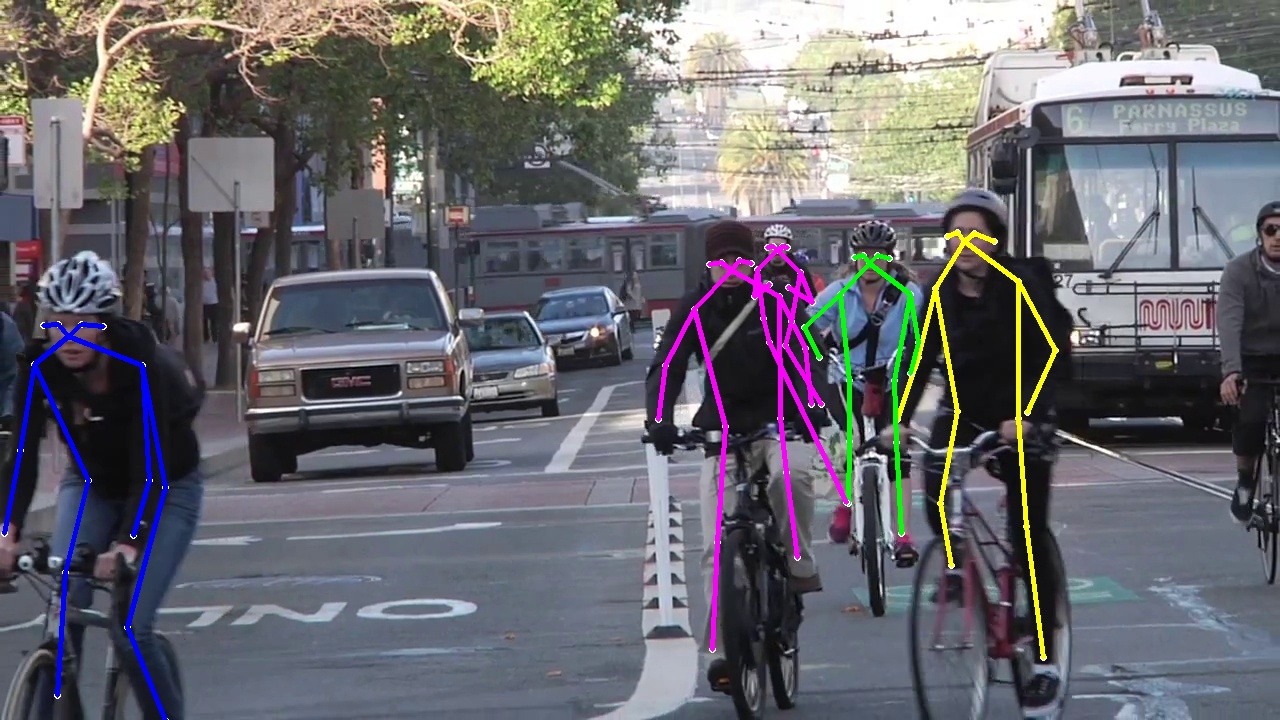} &
\includegraphics[width=0.2\linewidth]{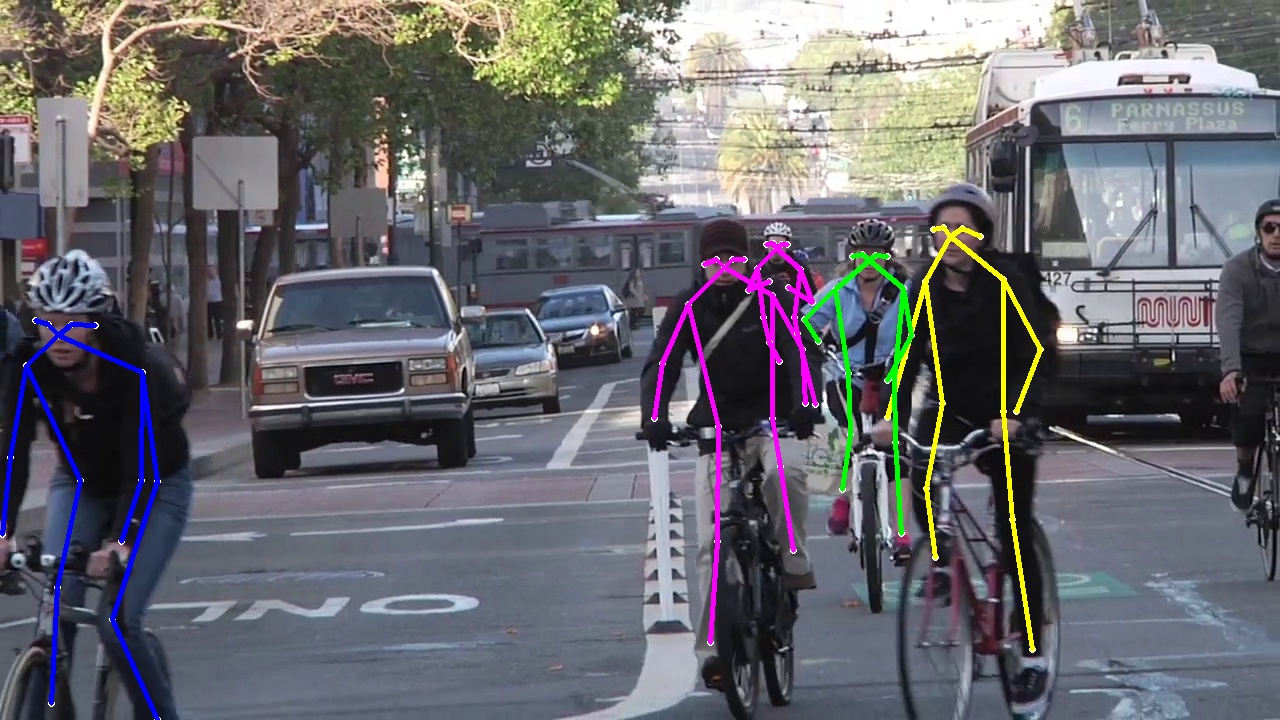} &
\includegraphics[width=0.2\linewidth]{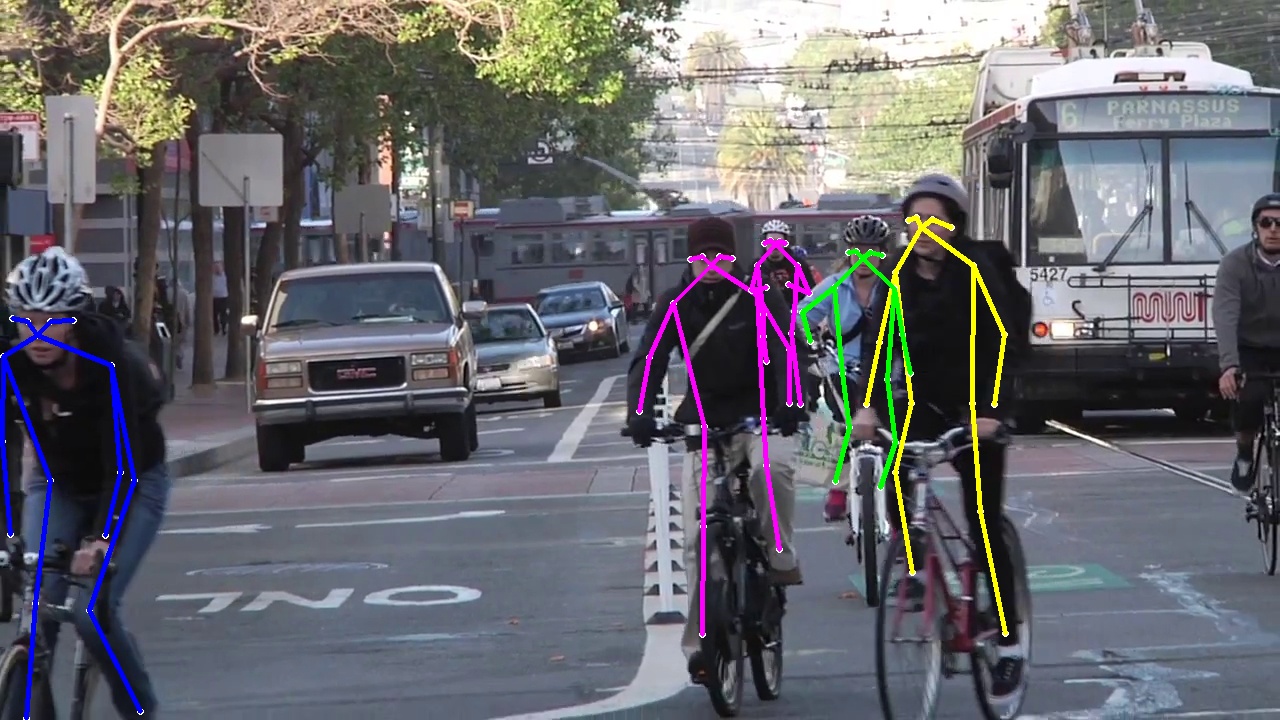} \\
\includegraphics[width=0.2\linewidth]{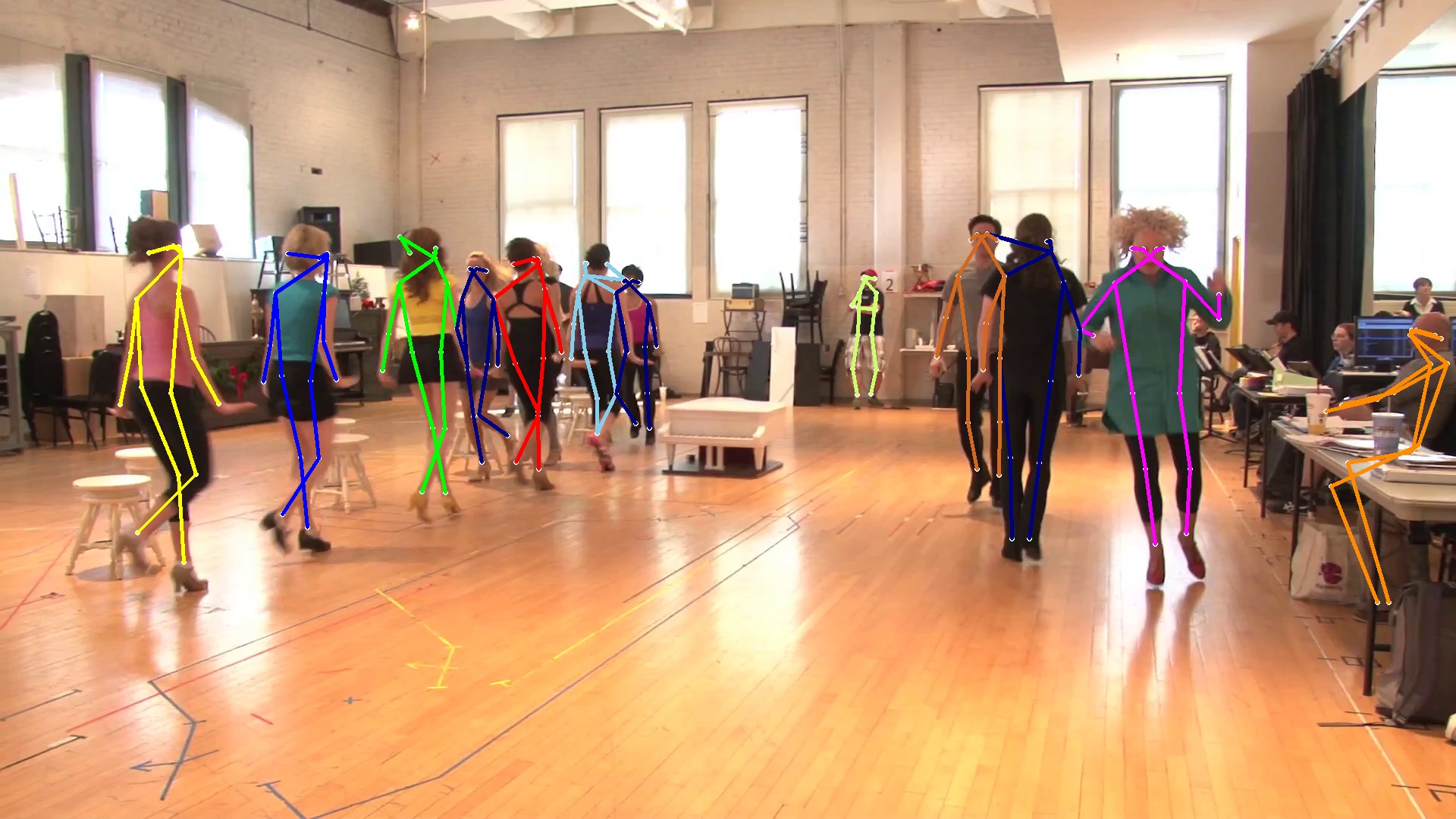} &
\includegraphics[width=0.2\linewidth]{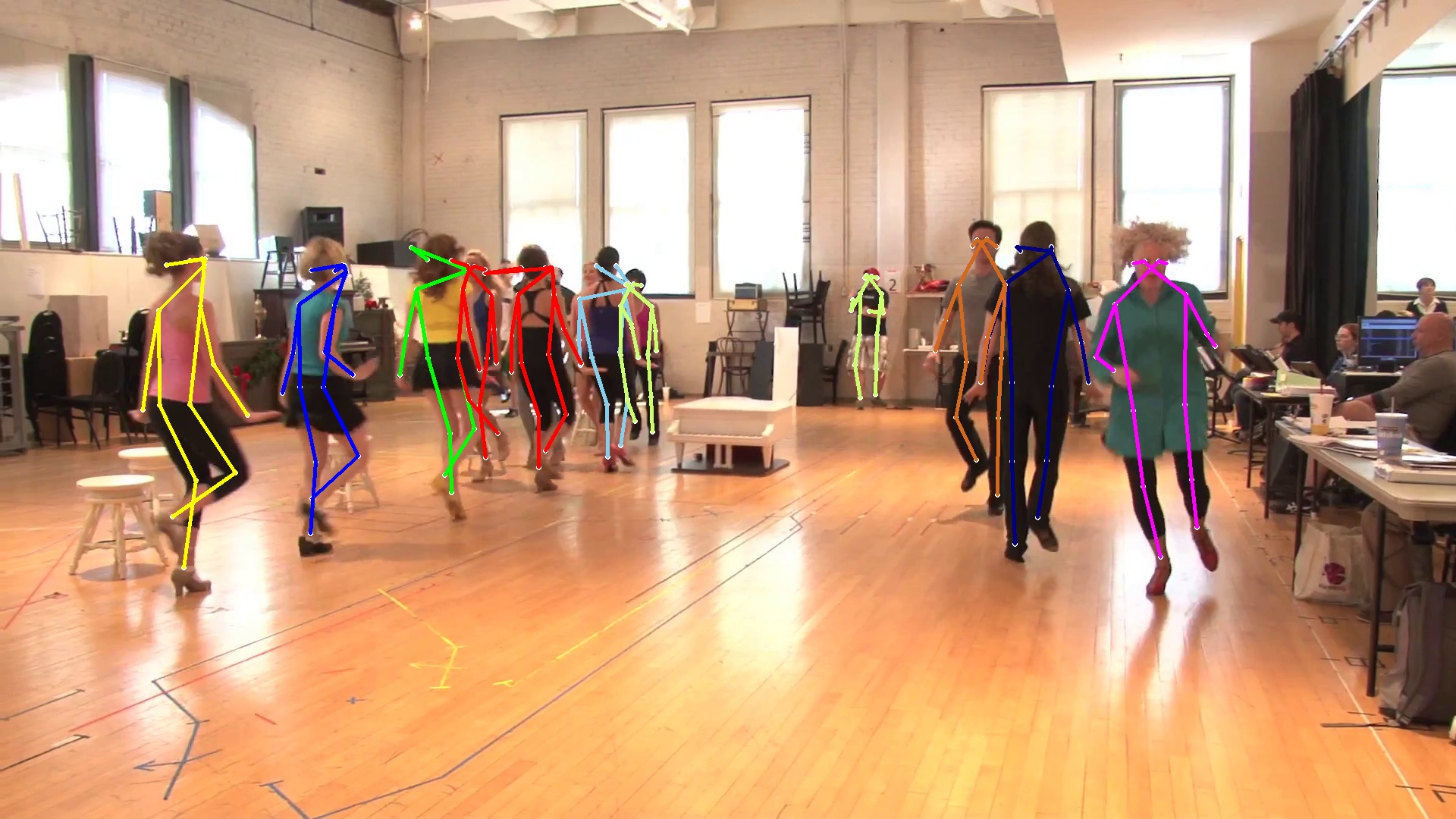} &
\includegraphics[width=0.2\linewidth]{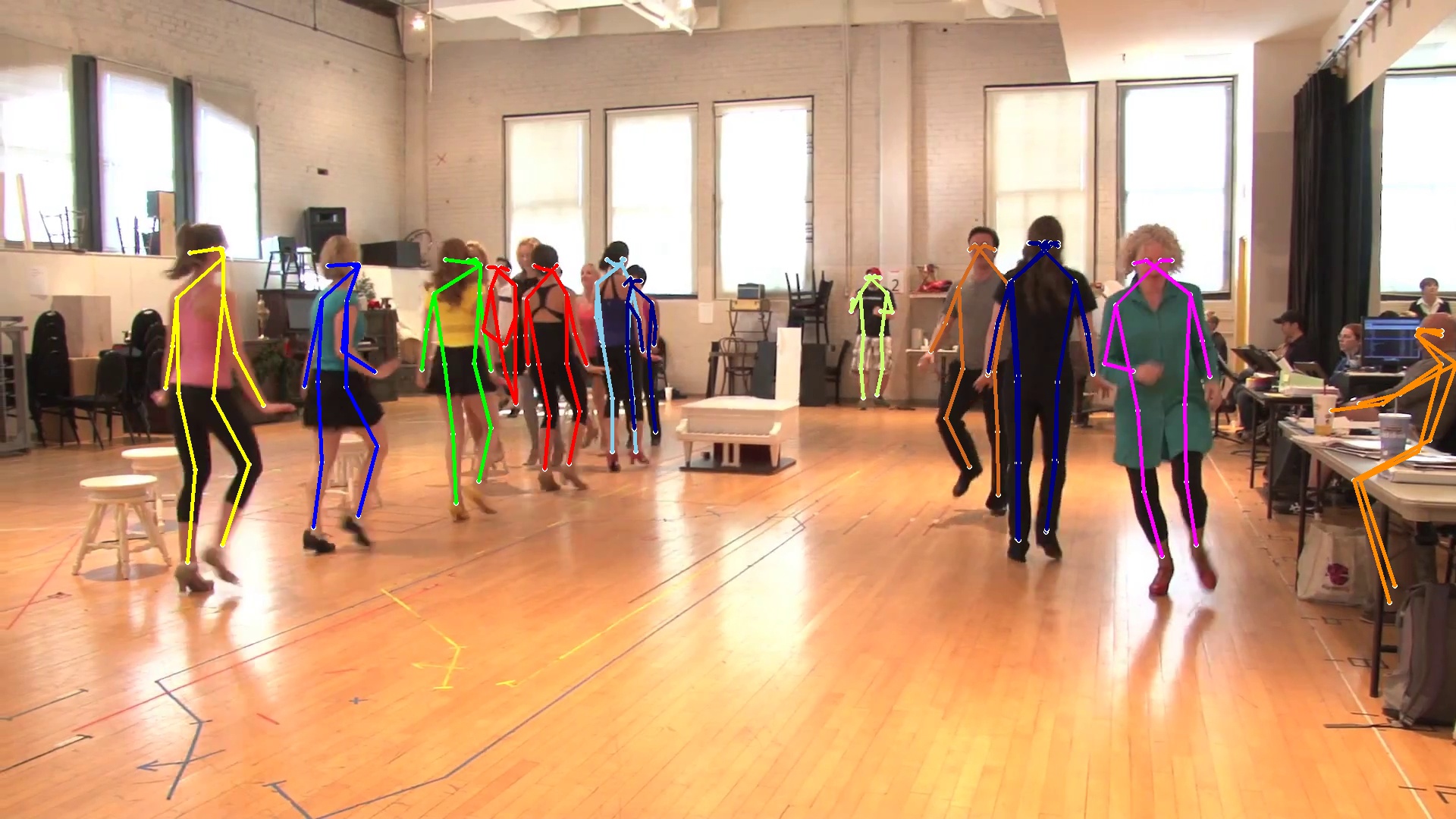} &
\includegraphics[width=0.2\linewidth]{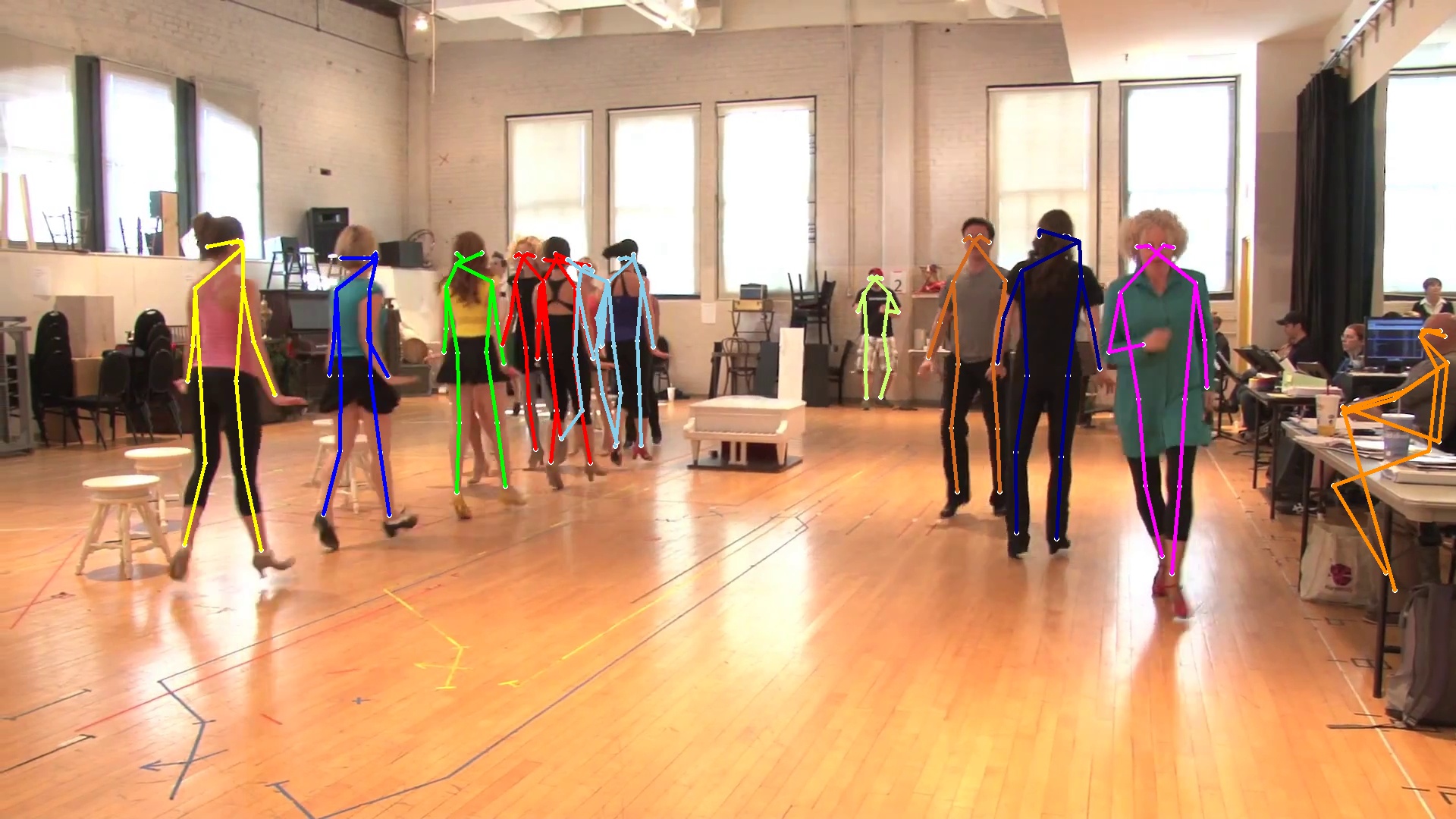} &
\includegraphics[width=0.2\linewidth]{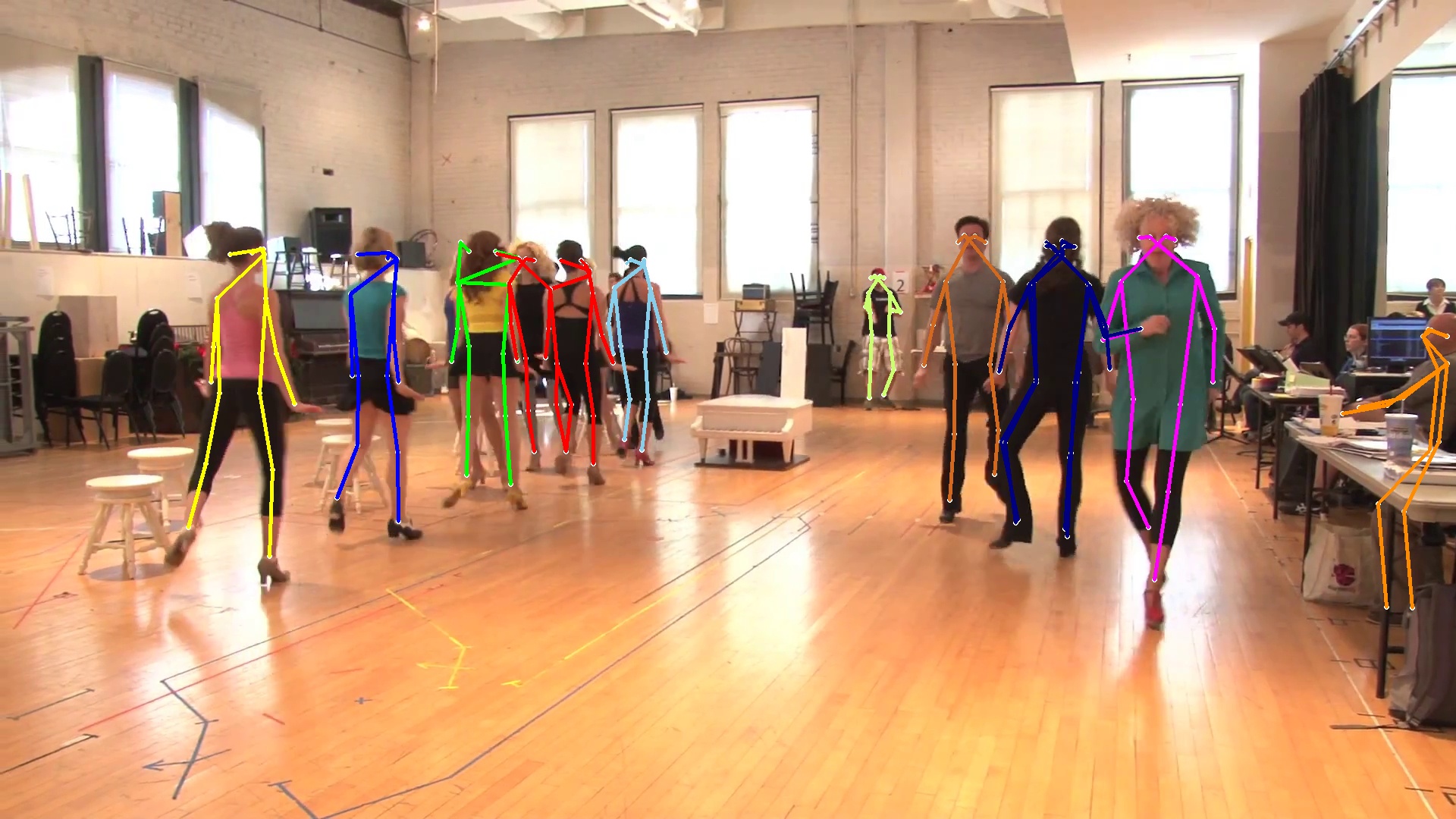} 
\end{tabular} \egroup
\end{center}\vskip-3mm
\caption{Visualization results of our proposed semi-supervised tracking approach on video instance segmentation and pose tracking. Each row has five sampled frames from a video sequence. Categories, bounding boxes and instance masks are shown for each object. Note that objects with the same predicated identity across frames are marked with the same color. Zoom in to see details.}
\vspace{-3mm}
\label{fig:vis}
\end{figure*}

\subsection{Comparison with State-of-the-Art Methods}

{\bf Video Instance Segmentation}. 
Since we test on a subset of the YouTube-VIS validation set, we either evaluate the publicly released models, or our re-implementation of the other approaches.
The comparison results are shown in Table~\ref{sota:vis}. 
Both the MaskTrack RCNN and SipMask have a tracking branch to learn object embedding representation from labeled videos.
Compared to them, our method, although does not involve any annotation of videos, can still achieve comparable performance. Furthermore, with the video instance correspondence module, our approach achieves the best performance across all evaluation metrics. 

In addition, we compare our approach to the methods involving various cues for post processing. IoUTracker+~\cite{bochinski2017high} assigns the instance label with the largest score to a candidate box. Since it does not leverage any visual information, its performance is a little weaker. SeqTracker~\cite{yang2019video} first computes instance segmentation results for all frames of a video, and then searches all possible tracks to find the one with the largest score. 
MaskTrack RCNN and SipMask perform the post-processing proposed by~\cite{yang2019video} to have more comprehensive cues for object association. By adopting a similar post-processing strategy, our approach can achieve comparable or even better performance versus other SOTAs. 
Furthermore, with the help of self-supervised Test-time adaption strategy, we can improve the final performance by more than 1\% on AP and AP$_{0.5}$.
Fig.~\ref{fig:vis} (Row 1-2) shows some qualitative results on YouTube-VIS validation set. Each row represents the predicted results on different frames in a video.

{\bf Analysis of Post-processing}. We notice that similar post-processing steps bring much more improvement to methods that train with video annotations than our approach. For instance, the AP performance of MaskTrack R-CNN~\cite{yang2019video} and SipMask~\cite{Cao_SipMask_ECCV_2020} improves by 7.0\% and 13.6\%, respectively, with category and spatial consistency. However, the improvement of our method is only about 4\%. This is because post-processing takes additional cues from the instance segmentation results, \ie category prediction, bounding box localization and mask prediction. However, due to the obvious domain gap between the training set of COCO, and the testing set of YouTube-VIS, the performance of both modules drops accordingly, and thus the limited improvement after post-processing compared to the others. We note we mainly focus on learning a tracking embedding representation in this work. We leave domain adaptation of the original SOLO heads to the further work.


{\bf Pose Tracking}. Besides video instance segmentation, our approach can also be extended to human body pose tracking. We compare our approach with the SOTAs and report results on the validation set of PoseTrack2018. 
The results are summarized in Table~\ref{sota:pose}. Note that since the number of joints and their definitions are different in COCO~\cite{lin2014microsoft} and PoseTrack~\cite{andriluka2018posetrack}, an additional finetuning step on MPII~\cite{andriluka20142d} is employed (denoted as Our$^{++}$ in Table~\ref{sota:pose}). In general, our proposed method can achieve comparable results to both top-down and bottom-up methods. For instance, comparing with the top-down methods, although our performance on AP is slightly lower, our performance on MOTA is quite competitive. However, the top-down methods always detect the human body first and perform pose estimation and tracking on cropped person images, which are much slower than ours. The analysis of running time is included in the supplementary material. Additionally, our approach even outperforms most of bottom-up methods. For instance, compared to STAF~\cite{raaj2019efficient}, the improvement is substantial: +3.8\% on MOTA and +1.0\% on AP.

%% file: TEX/5_conclusion.tex
\vspace{-1mm}
We introduce a novel semi-supervised framework that can achieve instance tracking without any video annotations. The Instance Contrastive loss and Maximum Entropy regularization are proposed to learn the discriminative representation of different instances capable of tracking via image annotations. Furthermore, in order to leverage the unlabeled videos, which are more accessible in the real-world, we propose to learn video correspondence in a self-supervised manner. Instead of learning a separated network, we integrate all proposed components into existing bottom-up instance segmentation or pose tracking frameworks. Extensive experiments demonstrate that our proposed method performs on par if not better than most STOA approaches.

%% file: TEX/appendix.tex
\appendix
\section*{Appendix}
\section{Architecture Details}
\paragraph{Video Instance Segmentation.} 

As described in the paper, we propose an instance embedding head to learn the discriminative representation of different instances. This head shares a similar structure to the category prediction head in SOLO~\cite{wang2019solo}. Specifically, we use four convolutional layers with $256$ output channels followed by group normalization layers. We add an additional convolutional layer with 128 output channels for dimensionality reduction. This embedding module is adopted for features at different levels in FPN~\cite{lin2017feature}. The video correspondence branch has the exact same structure as the embed branch. 

\begin{figure}[t]
	\centering
	\includegraphics[width=0.45\textwidth]{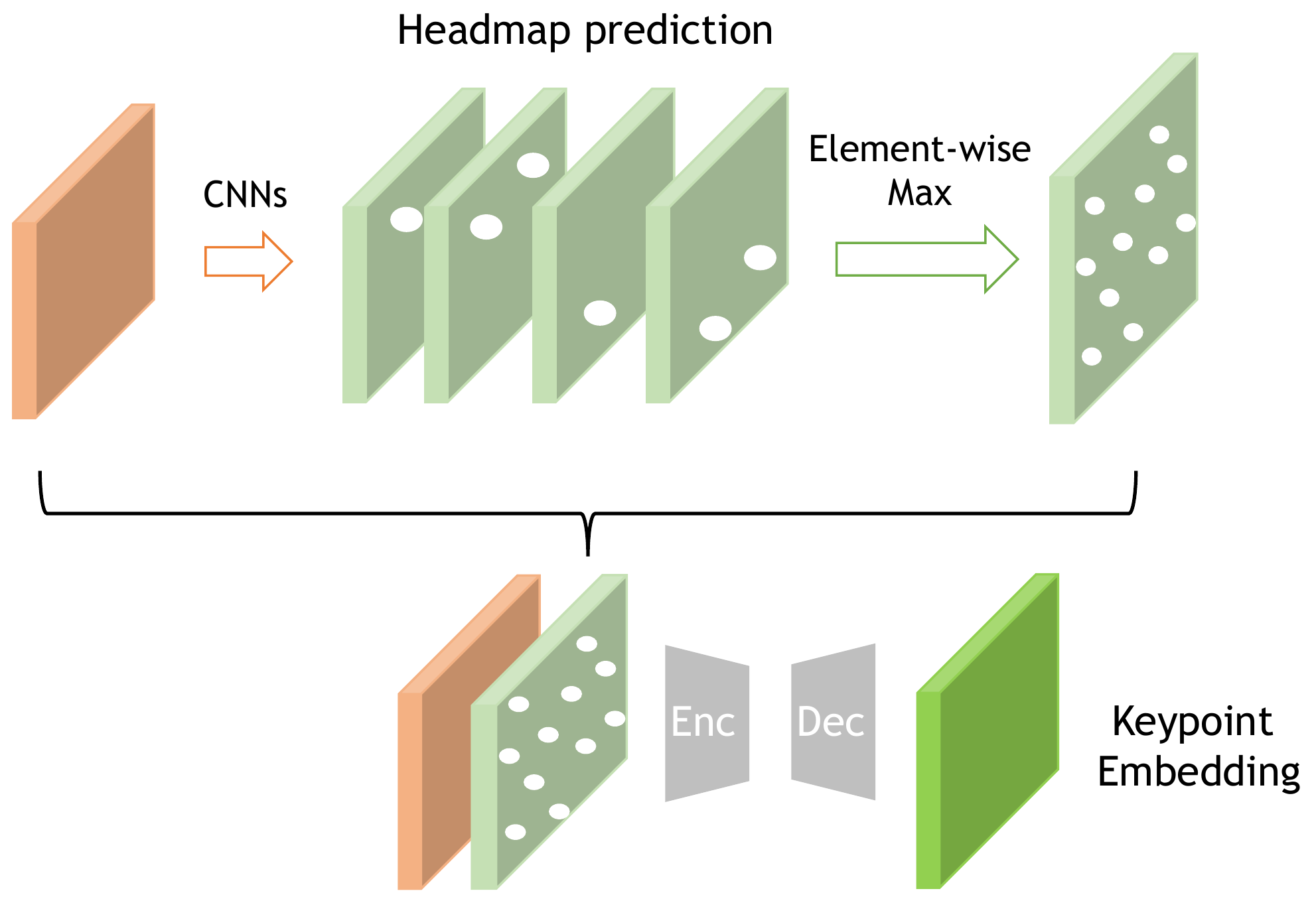}
	\caption{The architecture of the keypoint embedding module (KEM).}
	\label{fig:key}
\end{figure}

\vspace{-2mm}
\paragraph{Pose Tracking.}
Pose tracking~\cite{andriluka2018posetrack} is more challenging for learning a discriminative feature embedding, since it focus on discriminating between different humans, which are instances of the same category. That is, compared to YouTube-VIS~\cite{yang2019video}, pose tracking needs to learn a more fine-grained feature representation to discriminate different human instances across frames. Thus, we propose the keypoint embedding module (KEM) as demonstrated in Fig.~\ref{fig:key}.

Unlike the instance embedding module, the KEM is designed to learn the discriminative features of different joints. In particular, we first concatenate the predicted heatmap, which exists in the original PointSetAnchor~\cite{wei2020point}, see Fig.~\ref{fig:key}, with FPN features as the input to the embedding head. In contrast to designing the head similar to the classification branch in the video instance segmentation framework, we introduced an encoder-decoder with one convolutional layer as the encoder and one de-convolutional layer as the decorder. This encoder-decoder structure is used to obtain the keypoint-level embedding. In addition, the keypoint prediction is also adopted as prior knowledge to indicate the location of each joint on the embedding feature map, to filter out the valid keypoint embeddings of different joint definitions, \ie neck, shoulder, wrist  etc.
We apply the same instance contrastive (IC) objective both at the keypoint and the instance levels. In other words, we repeat the IC loss 17 times since there are 17 joints defined in the COCO~\cite{lin2014microsoft} dataset. Besides these, we also average the embedding of all seventeen joints as a person-level embedding and apply the IC loss on it again. This KEM is added as a branch in parallel to the classification and shape regression branches in PointSetAnchor~\cite{wei2020point}.

\vspace{-2mm}
\paragraph{Difference with associative embedding (AE).} The IC loss correlated to associative embedding approaches \cite{newell2016associative,jin2019multi}. However, both were designed to learn a keypoint embedding for spatial grouping within an individual image, e.g., AE in \cite{jin2019multi} was applied only to the SpatialNet that is independent of pose tracking, which was performed by another TemporalNet. AE in neither of them was utilized for learning cross frame correspondence, that we aim for.

\section{Implementation Details}
\paragraph{Training Details.}
For video instance segmentation, we use ResNet50~\cite{he2016deep} pretrained on ImageNet~\cite{deng2009imagenet} as the backbone network and train the SOLO~\cite{wang2019solo} framework with the proposed instance embedding branch via a classification loss, mask prediction loss and an IC loss on COCO~\cite{lin2014microsoft} instance segmentation annotations. The $\lambda$ parameter in Eq (5) is set to $1$. We further learn video correspondences across frames using unlabeled sequences of YouTube-VOS~\cite{xu2018youtube}. Each sequence is sampled from the same video randomly with random intervals from 2 to 8. 

Similarly, for pose tracking, we use HRNetW48~\cite{sun2019deep} as the backbone network and train PointSetAnchor~\cite{wei2020point} along with the KEM. The other steps are the same as those employed for video instance segmentation. Video correspondence is learnt on the PoseTrack2018~\cite{andriluka2018posetrack} training set without any annotations. Since the joint definitions of COCO~\cite{lin2014microsoft} are different from PoseTrack~\cite{andriluka2018posetrack}, we further fine-tune the model on the MPII~\cite{andriluka20142d} training data.

\paragraph{Inference Details.}
During inference, we associate the objects in an online fashion following a procedure similar to the one proposed in~\cite{yang2019video} for video instance segmentation. A memory bank is established to store all detection results: object category, bounding box location, mask segment and the learned embedding feature. Object association is achieved by cosine similarity of the object embedding feature. 


Different from~\cite{yang2019video}, which has an additional category of ``new object" while training with identity annotations (track ids) across frames, we do not have such a category definition. Thus we make several modifications to the original tracking procedure. Assume $M$ objects are detected in previous frames, and $N$ objected are detected in the current frame. Then the similarity scores should form a $N \times M$ association matrix. To effectively figure out the new objects, we employ a bi-directional softmax~\cite{pang2020quasi} instead of the original softmax. Bi-directional softmax computes the softmax operation along the row and column directions. The new object cannot guarantee good consistency in both directions, resulting in a lower score for new objects. Based on the similarity matrix, we assign every detected object ($1:N$) a unique identity through the row-wise argmax operation. If the similarity score is lower than a threshold, this object is considered as a new object and its embedding feature is concatenated to the memory bank. On the other hand if it is higher than the threshold, it is assigned to an existing object and its embedding is updated by the newly tracked object's with a momentum value of $0.7$. 

\vspace{-5mm}
\paragraph{Post-processing.}
To be consistent with the previous approaches and to improve tracking performance, we also apply the post-processing procedure introduced in~\cite{yang2019video}, which combines category confidence, bounding box Intersection over Union (IoU), embedding similarity and category consistency through a weighted sum.
In particular, the final similarity between newly detected objects and the existing candidates in the memory bank can be computed as:
\begin{equation}
s(n,m) = \mathrm{sim}(n, m)  +  \alpha \mathrm{c(n)}  +  \beta \mathrm{IoU(b_n, b_m)}  +  \gamma \mathrm{\delta(c_n, c_m)}
\end{equation}
where $\mathrm{c(n)}$ is the classification confidence score of the $n\mathrm{th}$ object, $\mathrm{c_n}$ is the predicted category and $\mathrm{\delta(c_n, c_m)}$ is the Kronecker delta function, which returns one if and only if $\mathrm{c_n}$ is equal to $\mathrm{c_m}$, otherwise it returns zero. Note that as discussed in our paper, the current post-processing method can only bring a limited improvement on our approach compared to others, due to the obvious domain gap between the training set of COCO, and the validation set of YouTube-VIS. In this work, we mainly focus on learning a tracking embedding representation while leaving domain adaptation of the original SOLO heads to further work.

\section{More Experiments}

\subsection{Pose Tracking Running Time}
The running time of the proposed semi-supervised tracking approach on PoseTrack2018~\cite{andriluka2018posetrack} is shown in Table~\ref{runtime}. Compared with the top-down methods, LightTrack~\cite{ning2020lighttrack} and AlphaPose~\cite{fang2017rmpe}, our approach performs more efficiently since it estimates all joint locations of different persons at the same time. In addition, LightTrack~\cite{ning2020lighttrack} utilizes pre-computed human detection results and its efficiency can further decrease on considering the detection step as well.

\subsection{Comparison with Associate Embedding}
We also compare our proposed instance contrastive loss with Associate Embedding (AE) loss in~\cite{newell2016associative}. For a fair comparison, we apply AE to our pose tracking experiments. We replace our joint-level embedding with the original form of AE, while keeping all the other settings the same. The AE model achieves {\bf 63.5}\% on MOTA, which is lower than ours, i.e., {\bf 64.7}\%(Table.4 in the paper).

\subsection{More visualization}
We show more qualitative results of our proposed semi-supervised tracking approach on the video instance segmentation and pose tracking tasks and compare them with the baseline model, i.e., image-based instance segmentation/pose estimation models with spatial distance association, as described in our main paper (see Sec. 4.3) in Fig.~\ref{fig:compare}. It can be observed that compared to the baseline model, our proposed semi-supervised tracking can detect instance masks, human poses and associate different instances across frames much more accurately.

\begin{table}\setlength{\tabcolsep}{5pt}
\centering
\small
\begin{tabular} {l|c}
\hline
Method & runtime(fps) \\ \hline
LightTrack~\cite{ning2020lighttrack} & 0.8 \\
AlphaPose~\cite{fang2017rmpe} & 2.2 \\
\hline
Ours & 4.1 \\
Ours (ms) & 1.3 \\ \hline
\end{tabular}
\caption{Average running time of different pose tracking methods on the PoseTrack 2018 validation set. ``ms" represents multi-scale testing.}
\label{runtime}
\end{table}

\begin{figure*}
\begin{center}
\bgroup 
 \def\arraystretch{0.1} 
 \setlength\tabcolsep{0.5pt}
\begin{tabular}{cccc}
\multicolumn{4}{c}{\bf Before} \\ \\ 
\includegraphics[width=0.25\linewidth]{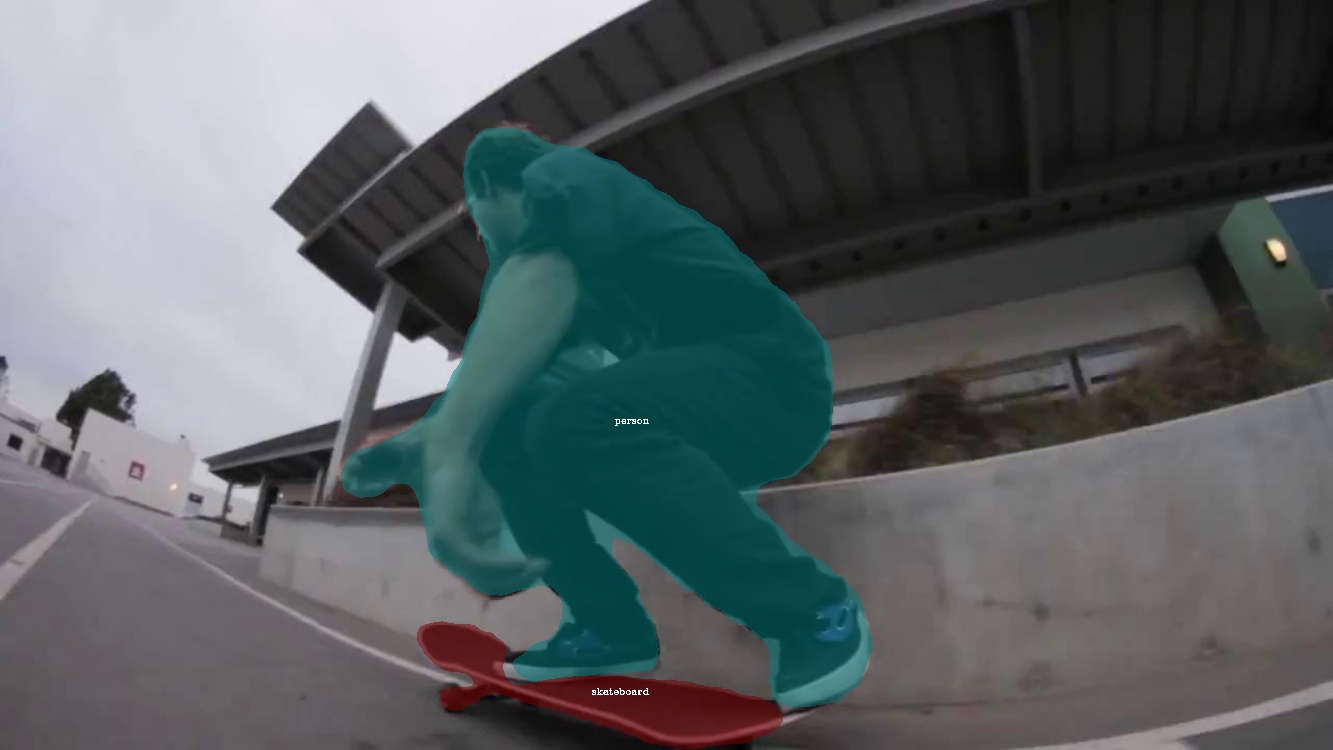} &
\includegraphics[width=0.25\linewidth]{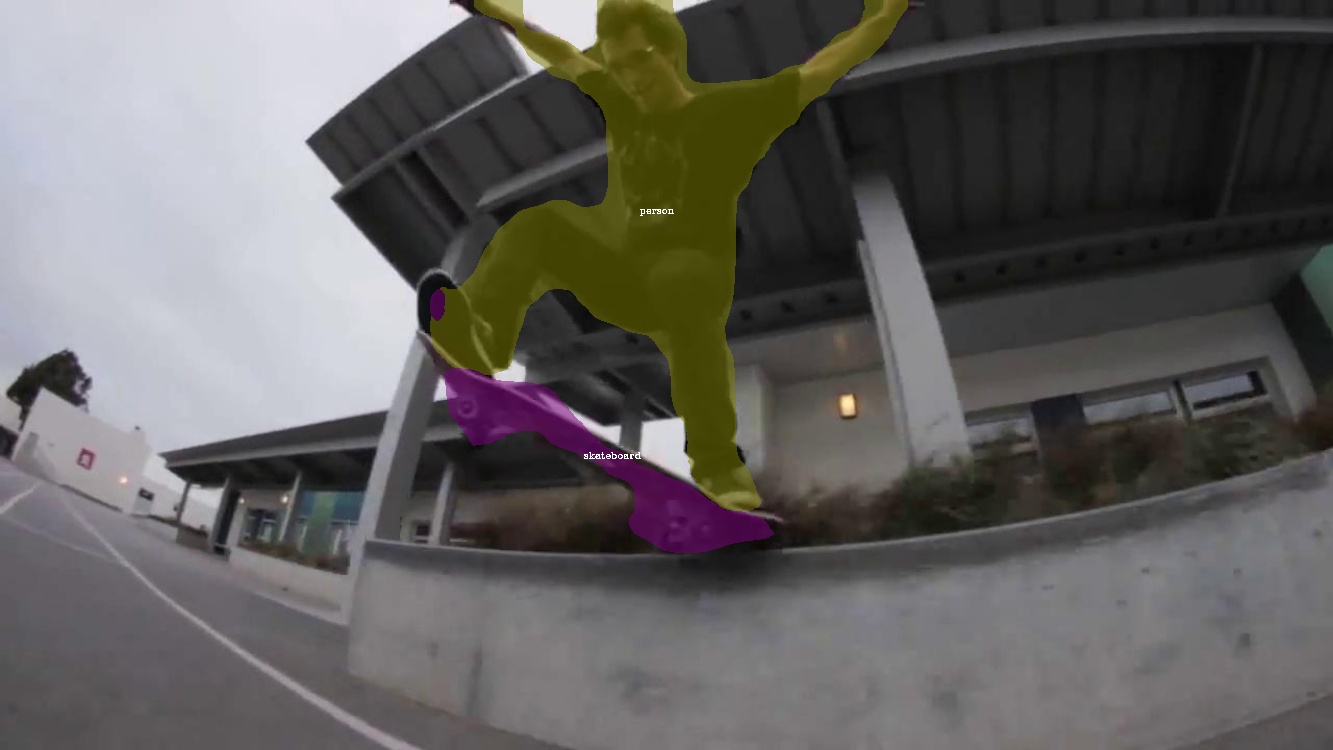} &
\includegraphics[width=0.25\linewidth]{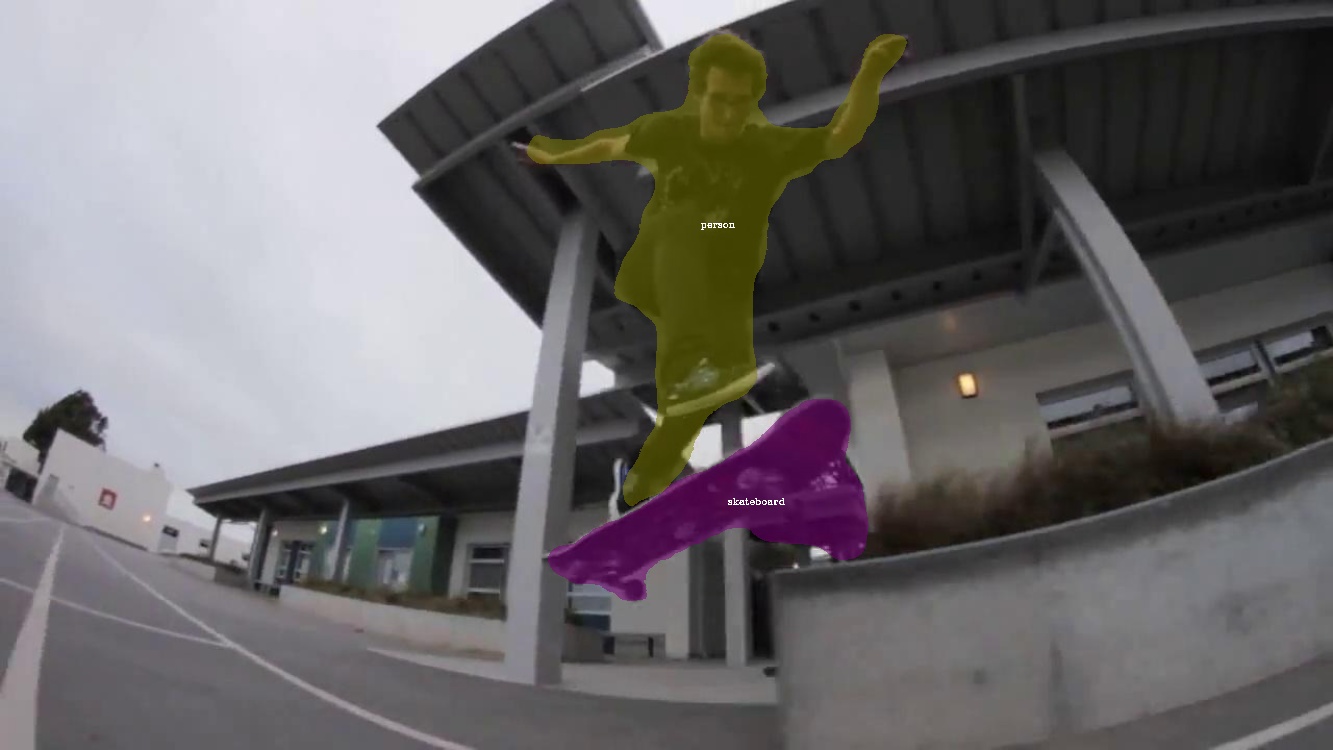} &
\includegraphics[width=0.25\linewidth]{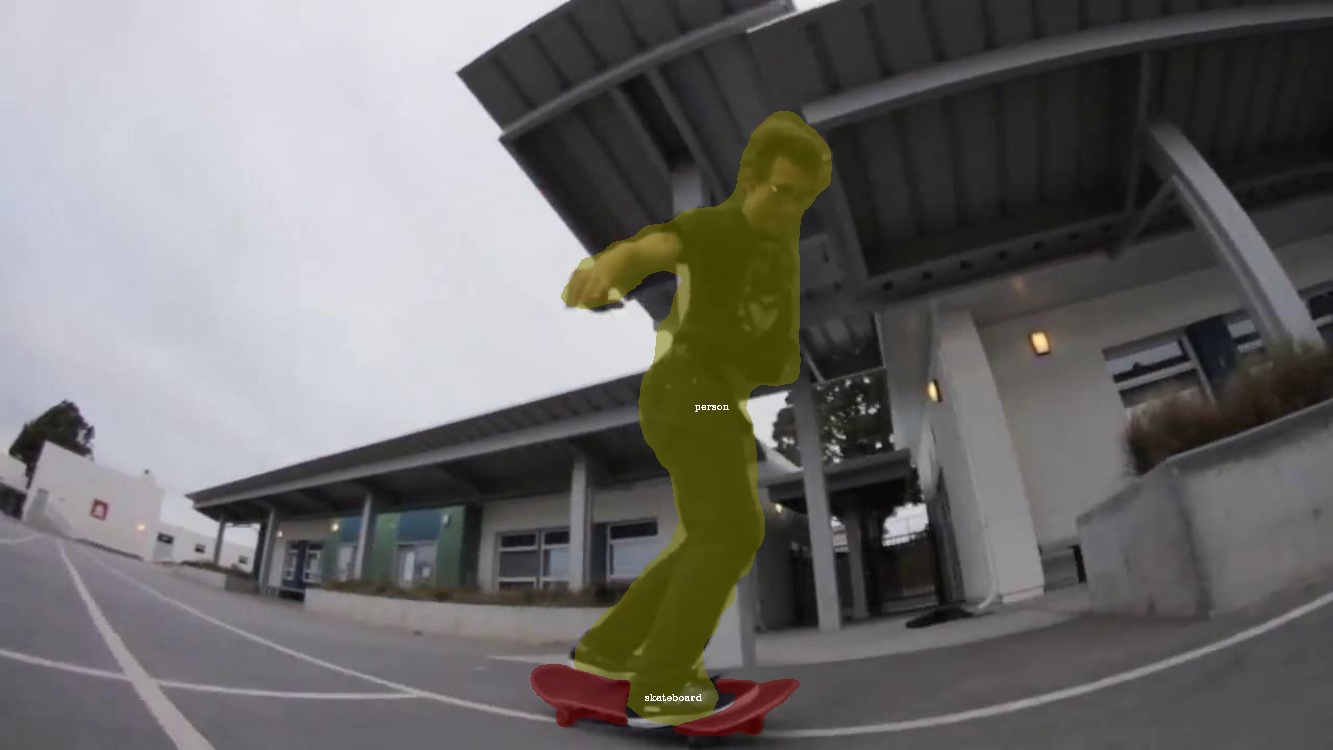} \\
\includegraphics[width=0.25\linewidth]{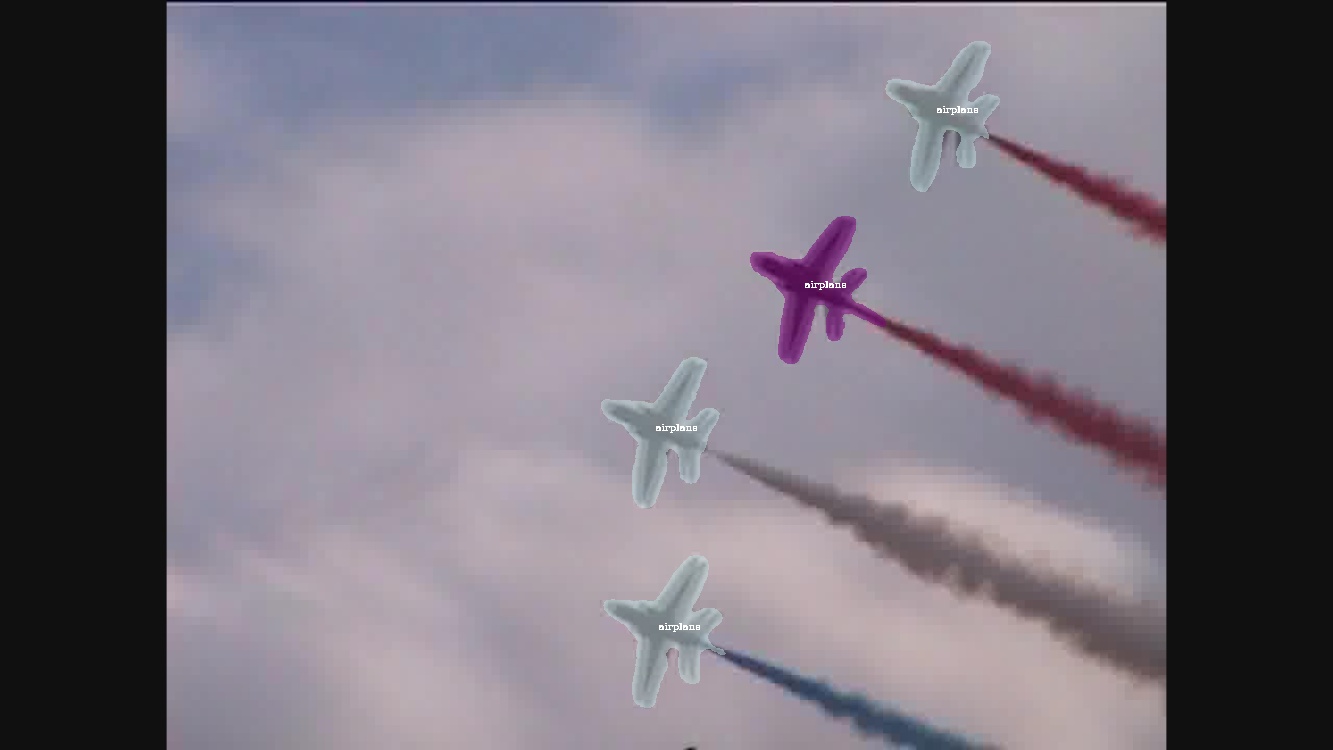} &
\includegraphics[width=0.25\linewidth]{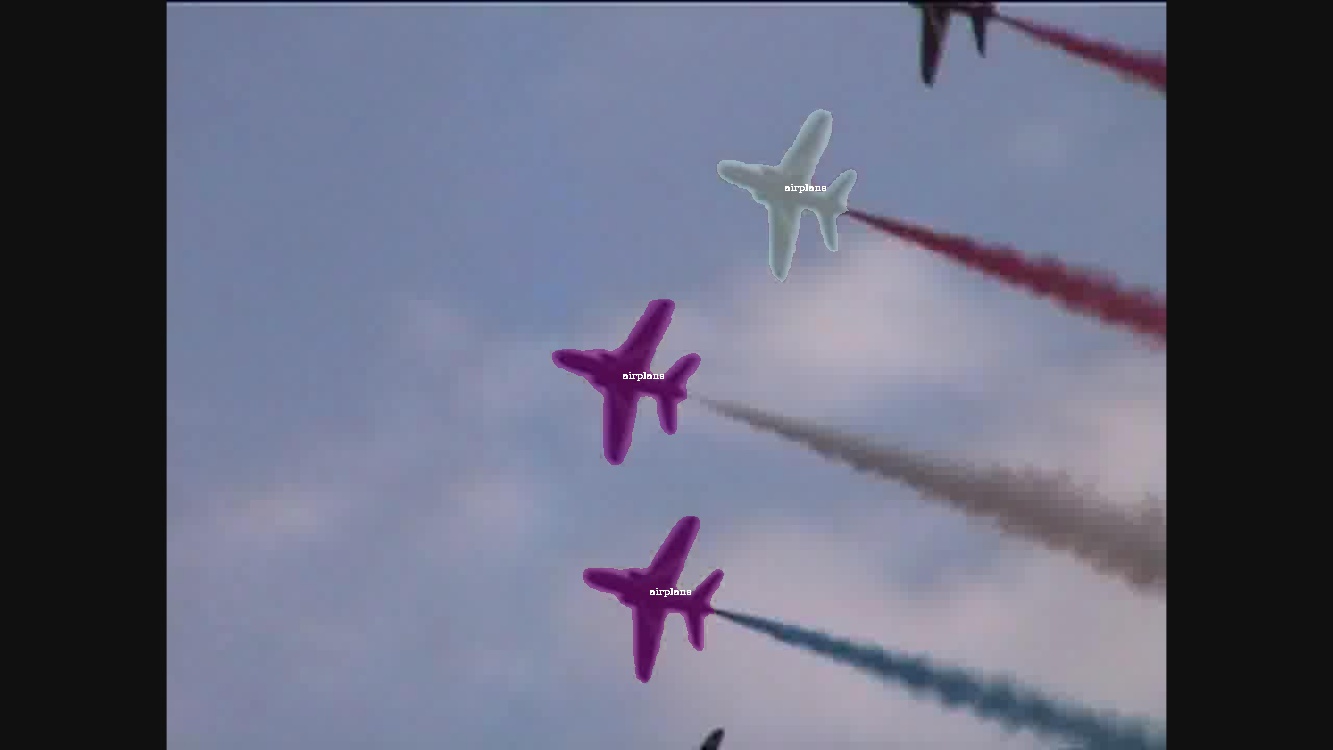} &
\includegraphics[width=0.25\linewidth]{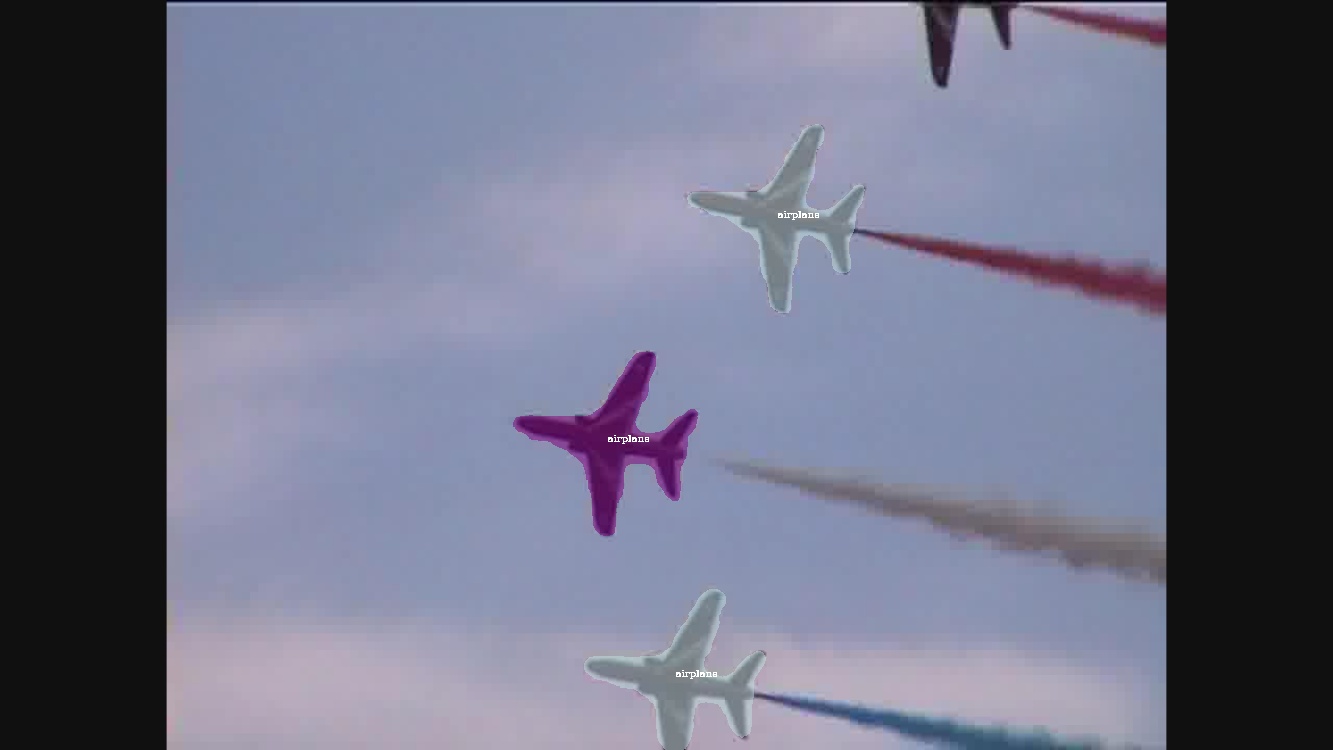} &
\includegraphics[width=0.25\linewidth]{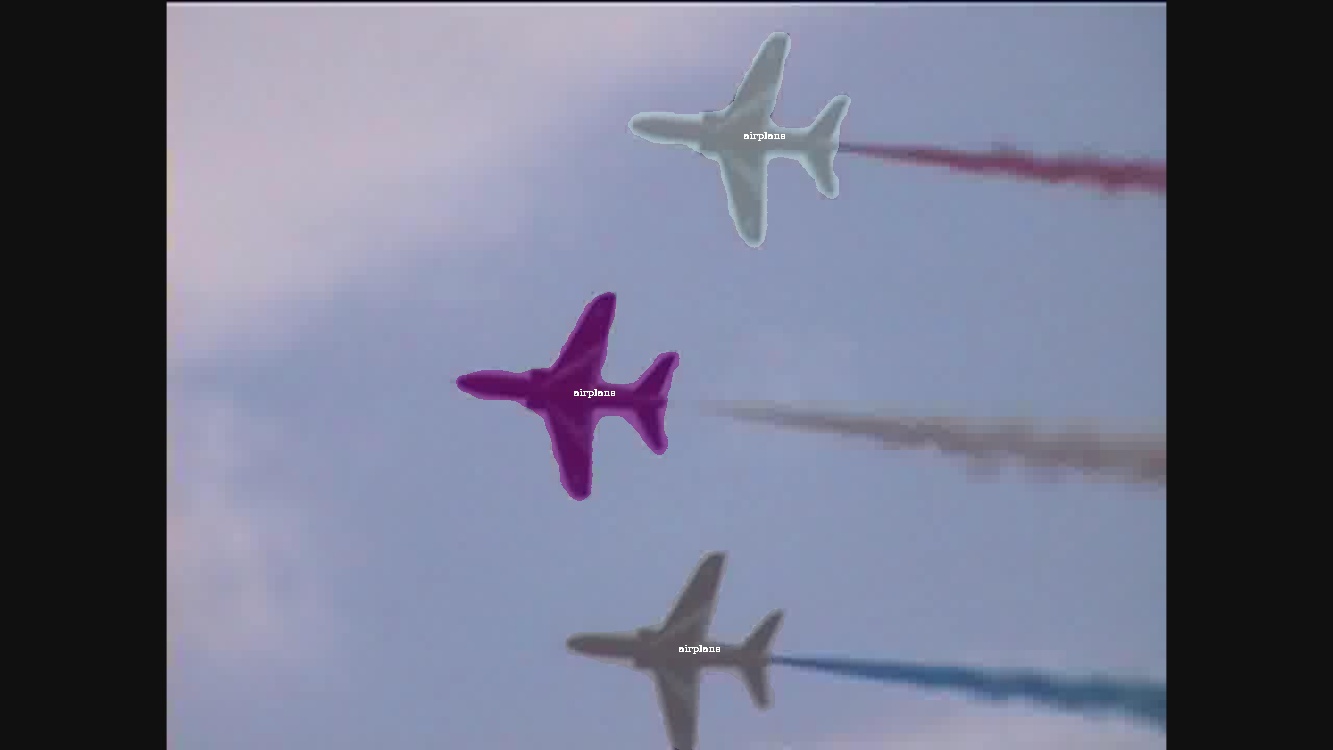} \\ 
\includegraphics[width=0.25\linewidth]{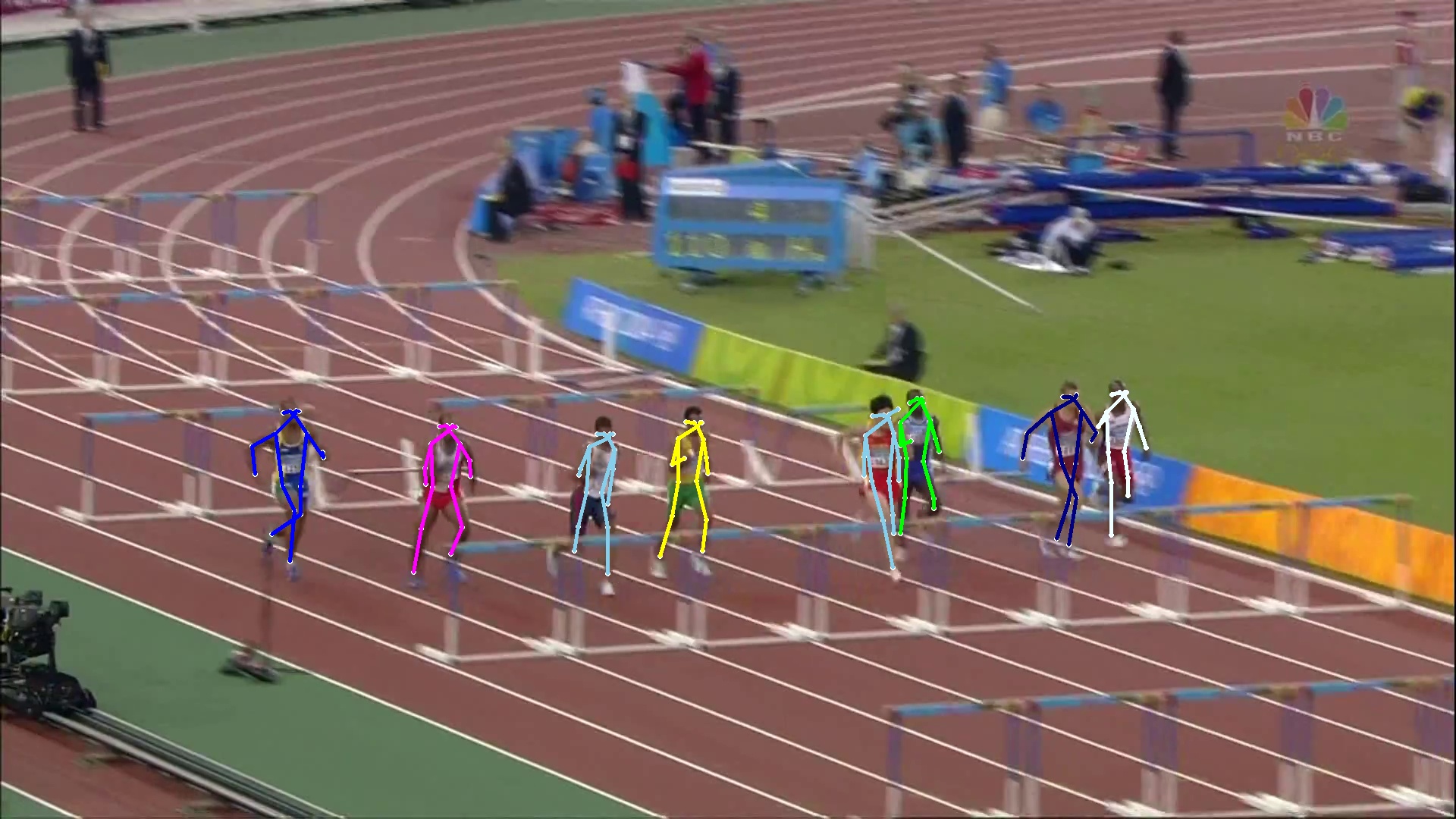} &
\includegraphics[width=0.25\linewidth]{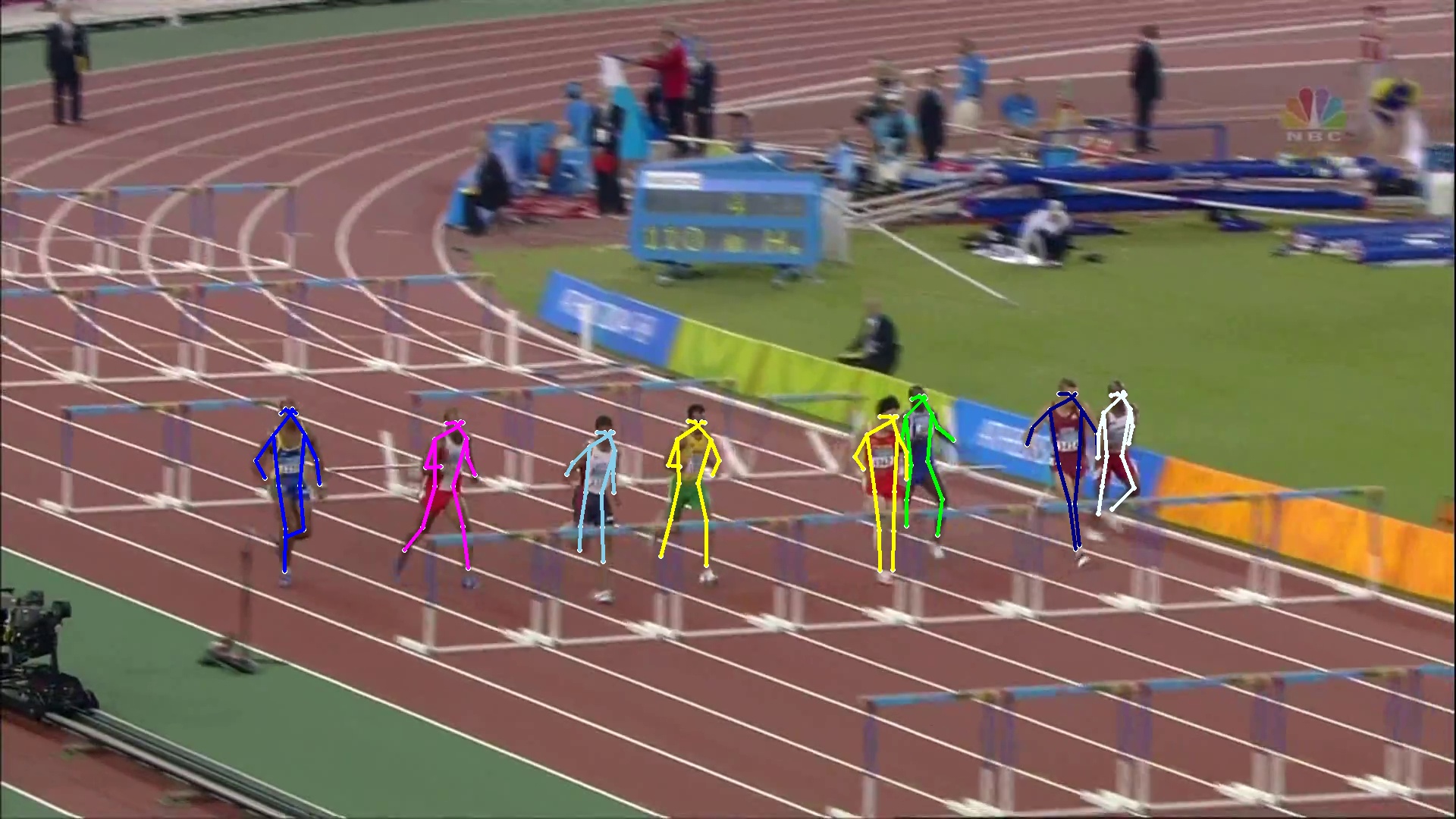} &
\includegraphics[width=0.25\linewidth]{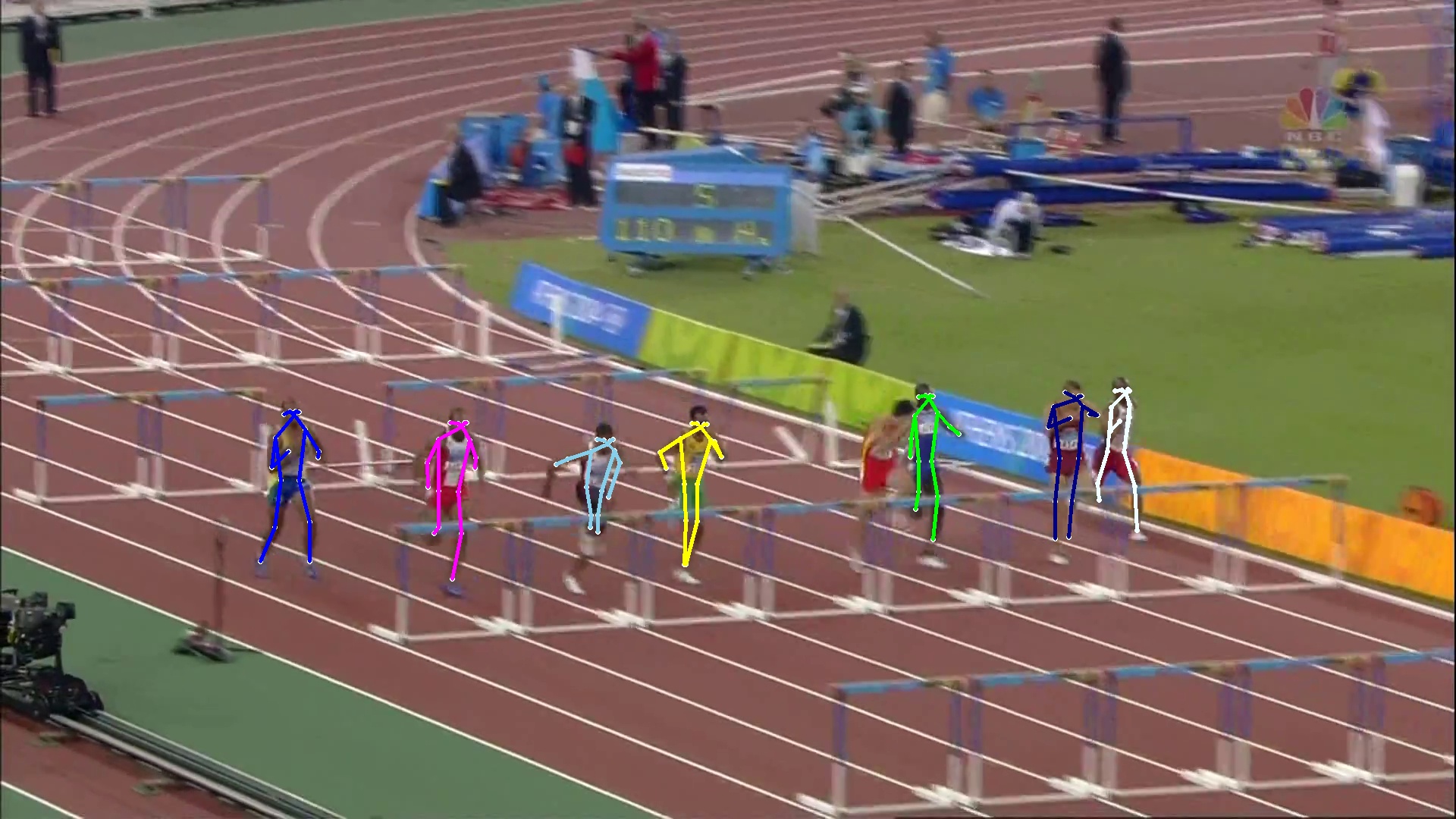} &
\includegraphics[width=0.25\linewidth]{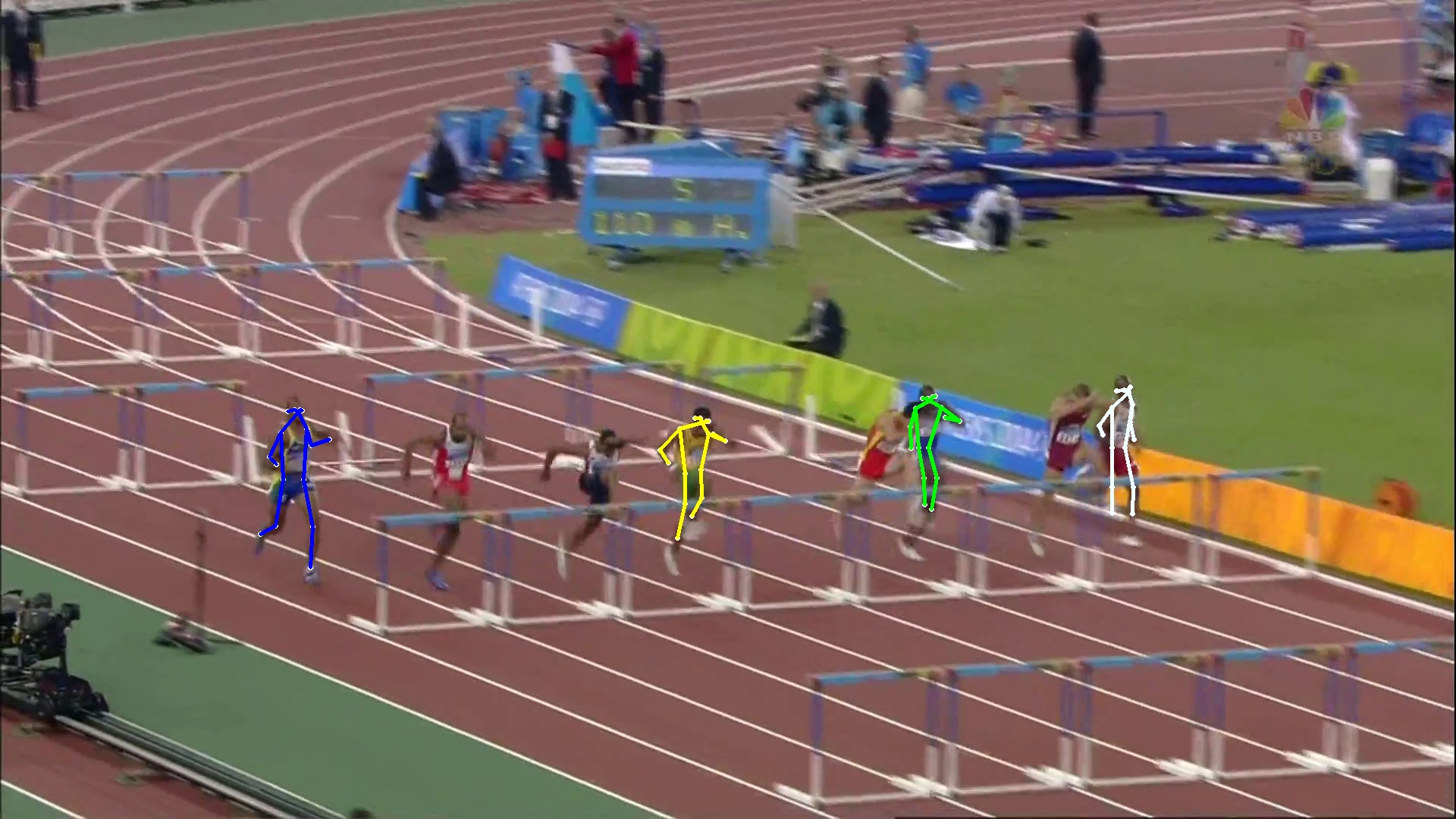} \\
\includegraphics[width=0.25\linewidth]{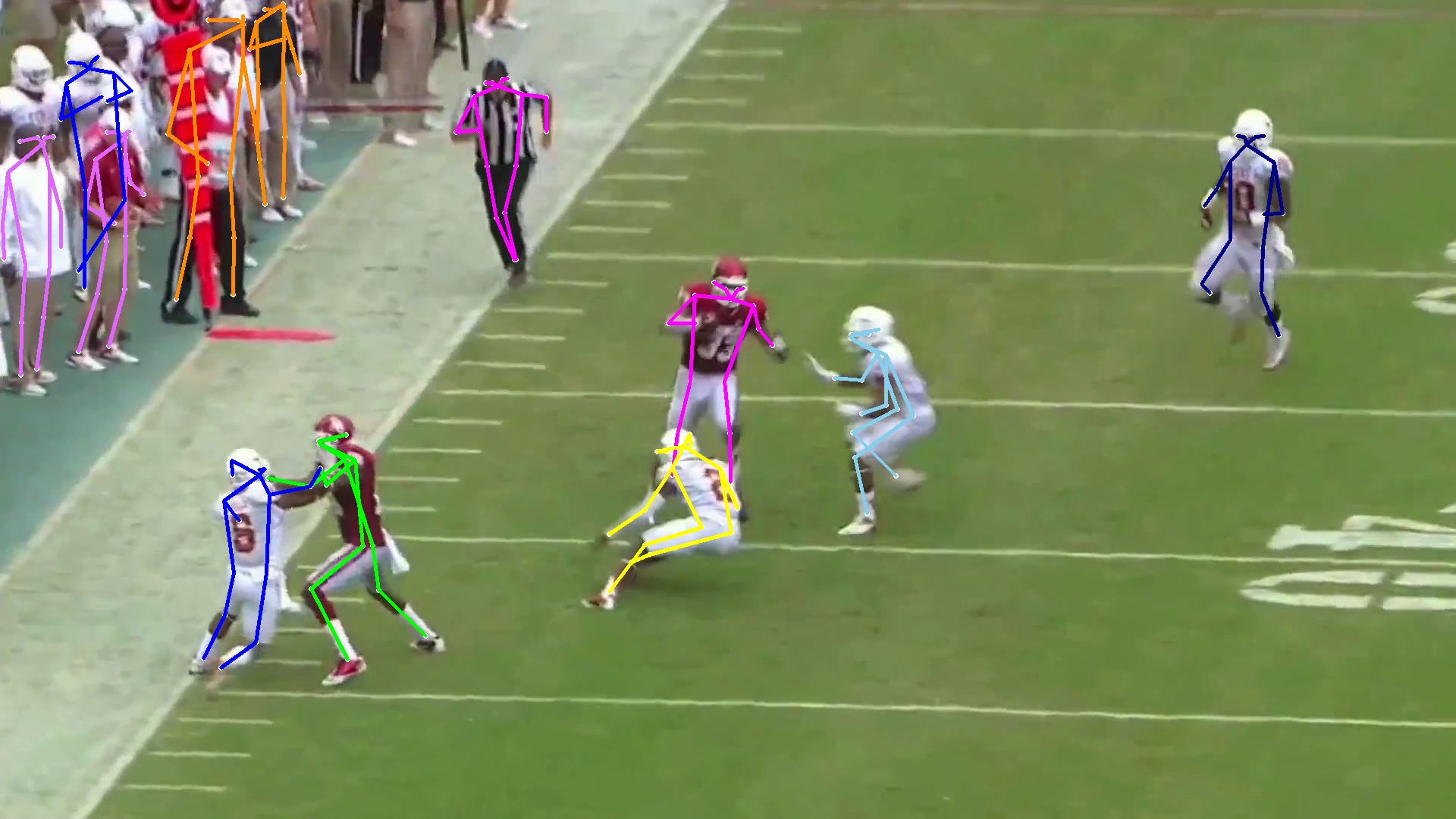} &
\includegraphics[width=0.25\linewidth]{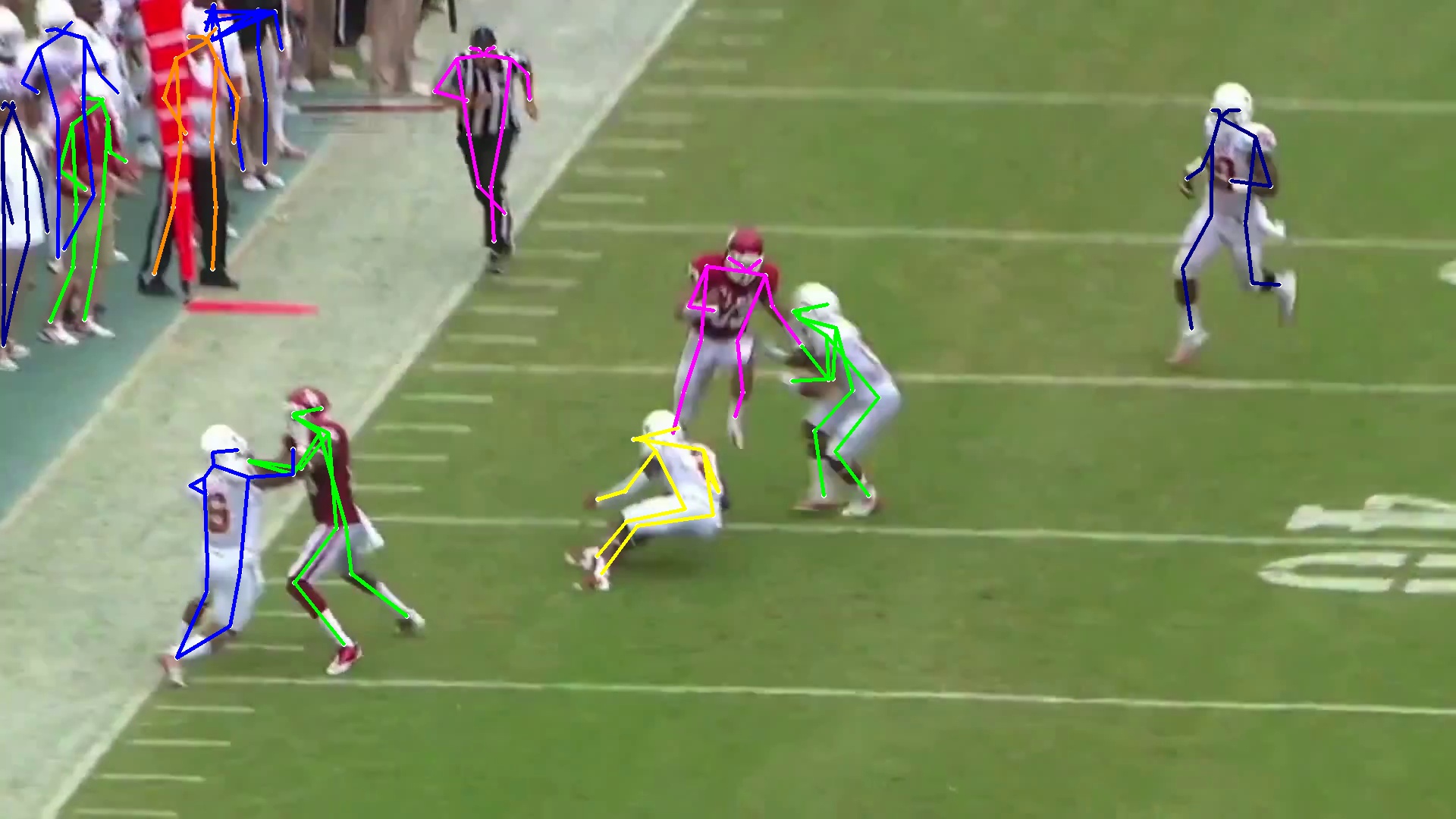} &
\includegraphics[width=0.25\linewidth]{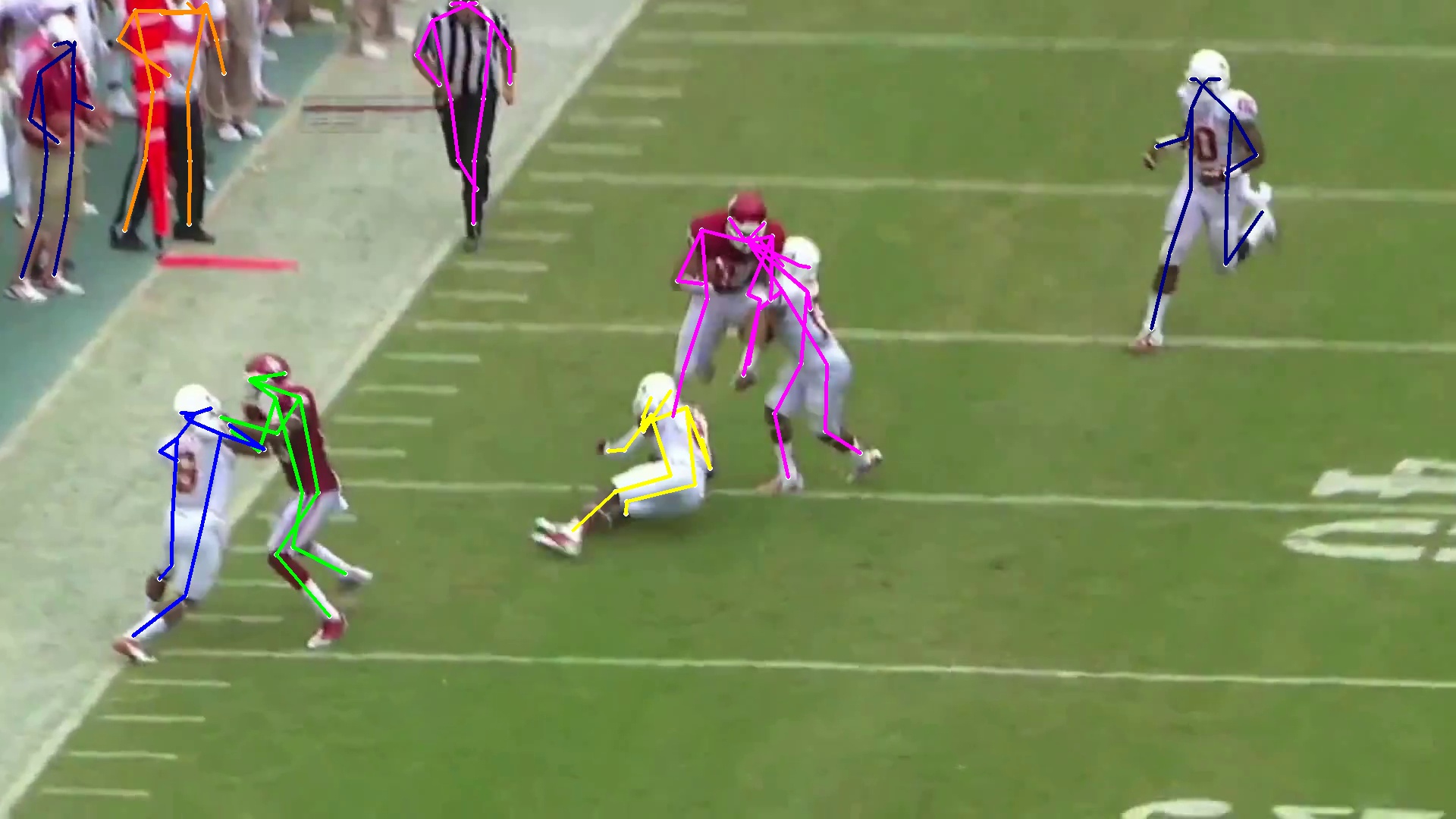} &
\includegraphics[width=0.25\linewidth]{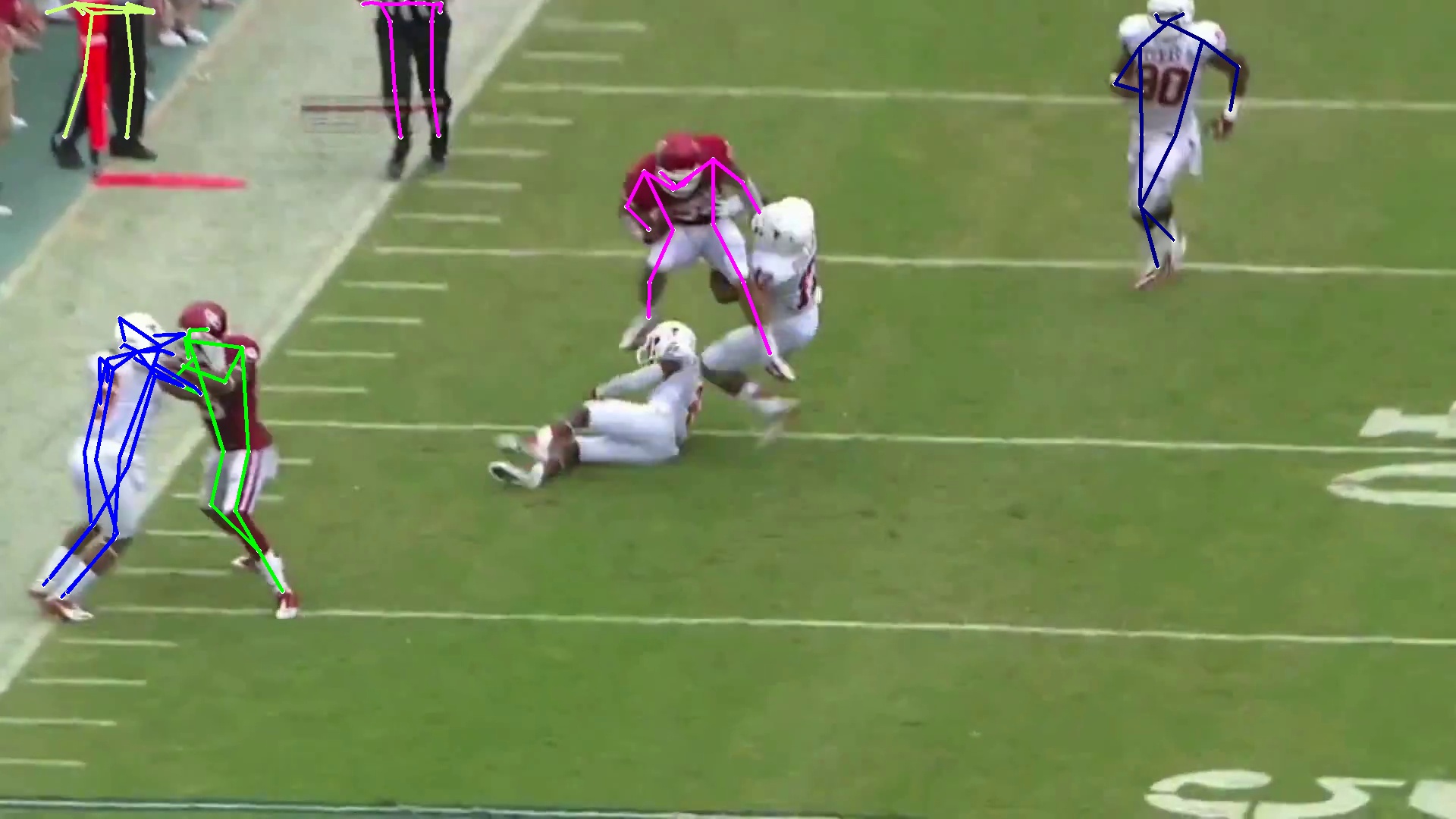} \\ \\
\multicolumn{4}{c}{\bf After} \\ \\ 
\includegraphics[width=0.25\linewidth]{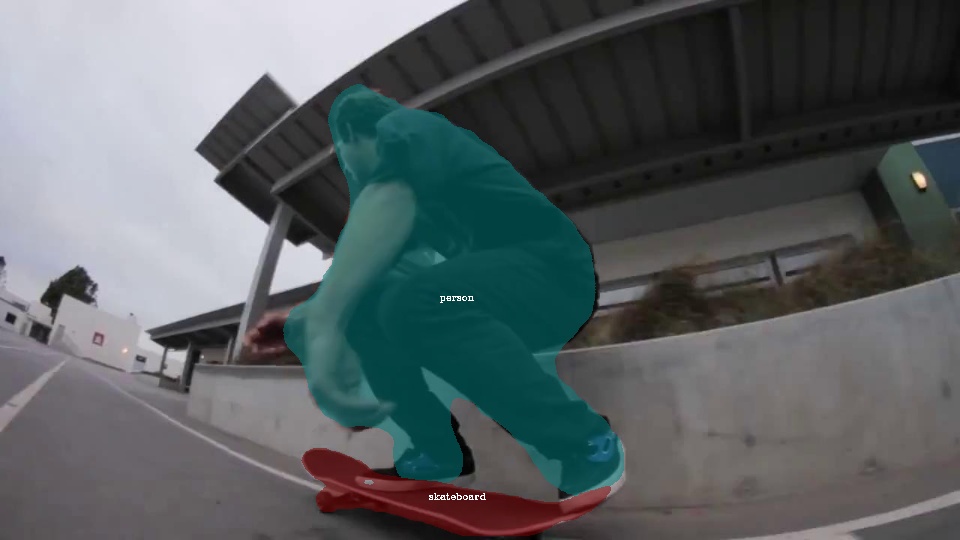} &
\includegraphics[width=0.25\linewidth]{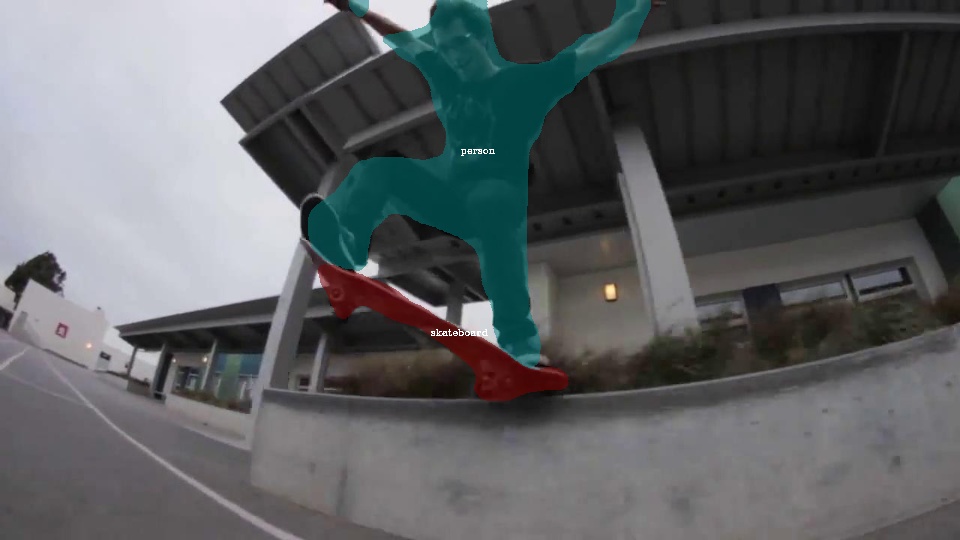} &
\includegraphics[width=0.25\linewidth]{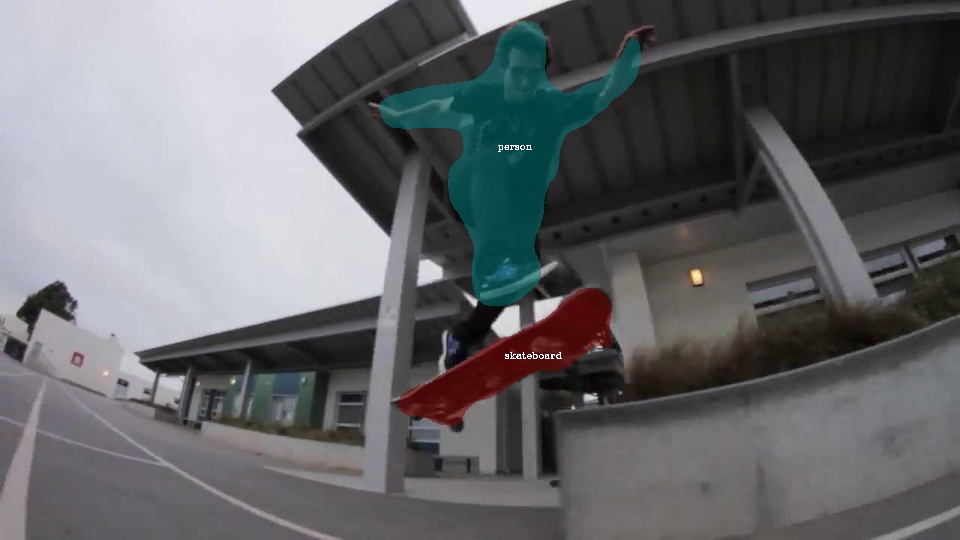} &
\includegraphics[width=0.25\linewidth]{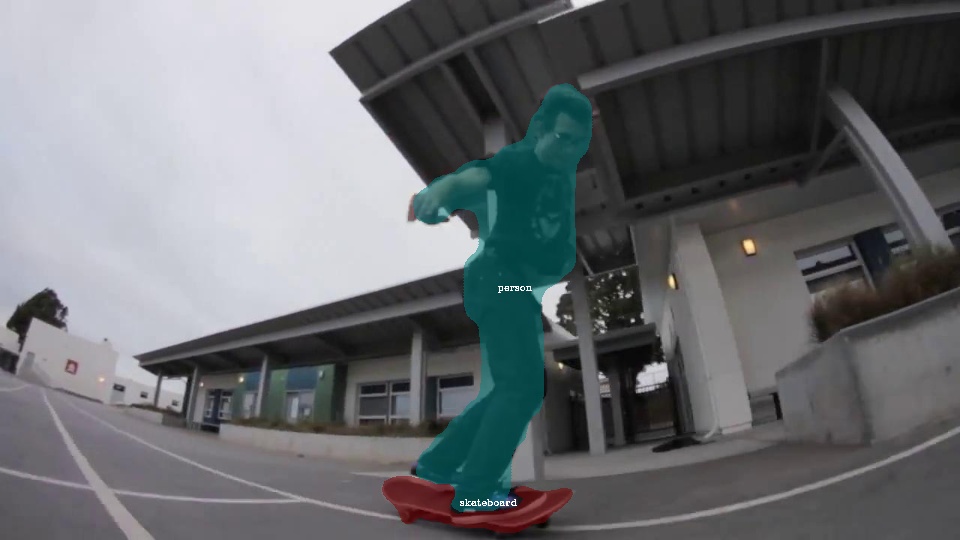} \\
\includegraphics[width=0.25\linewidth]{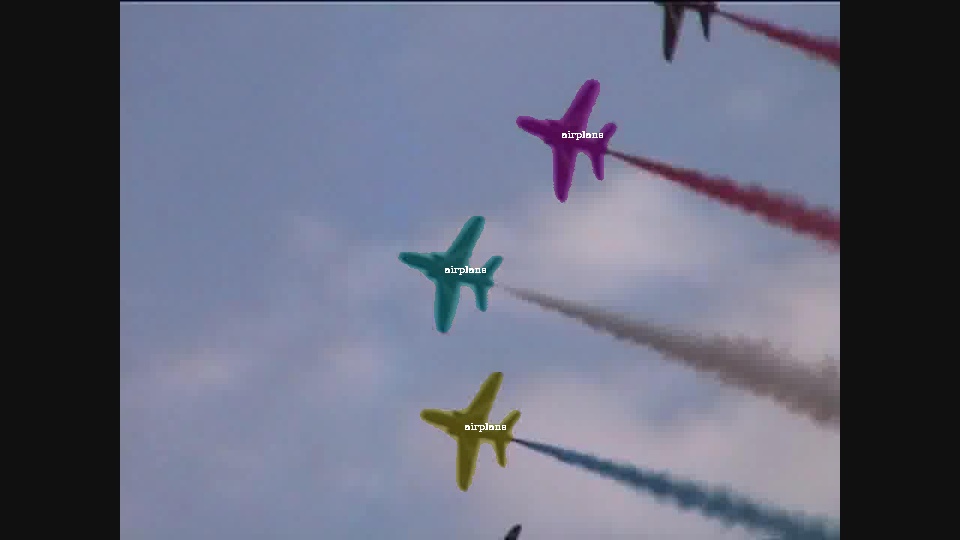} &
\includegraphics[width=0.25\linewidth]{figs/vis_after/2219.jpg} &
\includegraphics[width=0.25\linewidth]{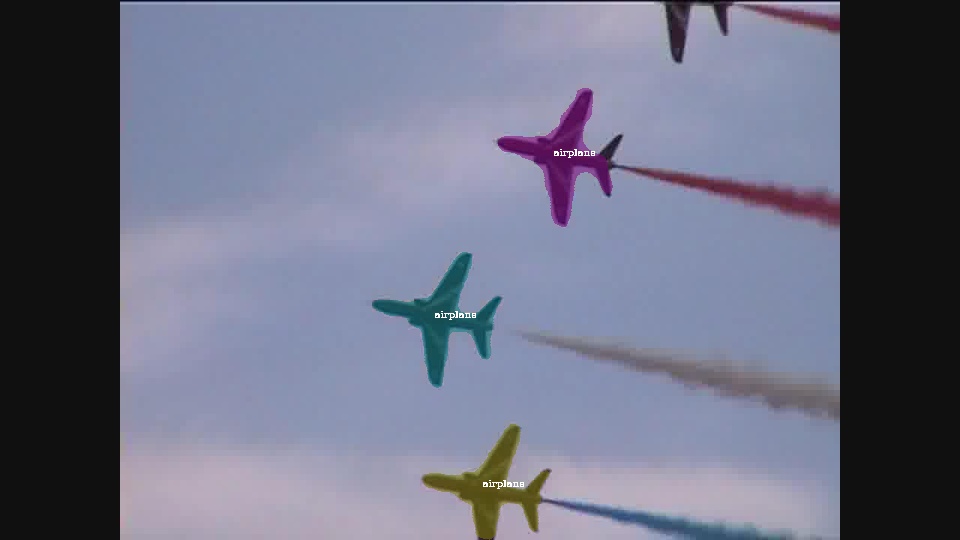} &
\includegraphics[width=0.25\linewidth]{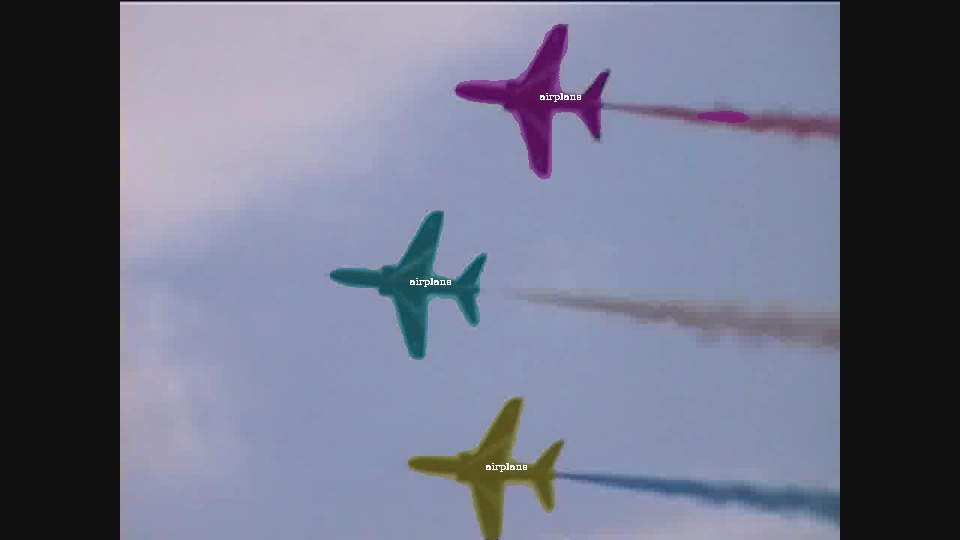} \\
\includegraphics[width=0.25\linewidth]{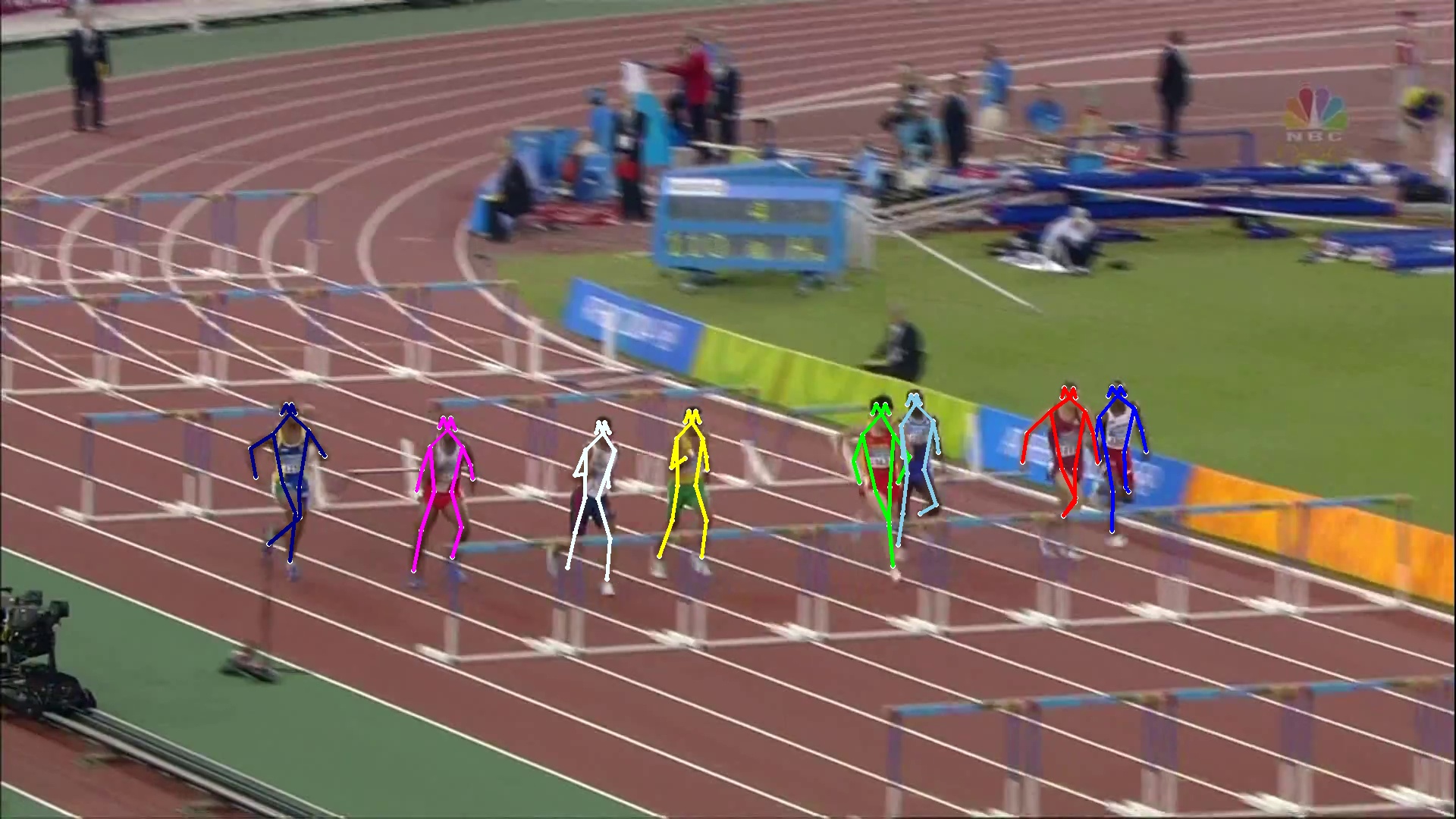} &
\includegraphics[width=0.25\linewidth]{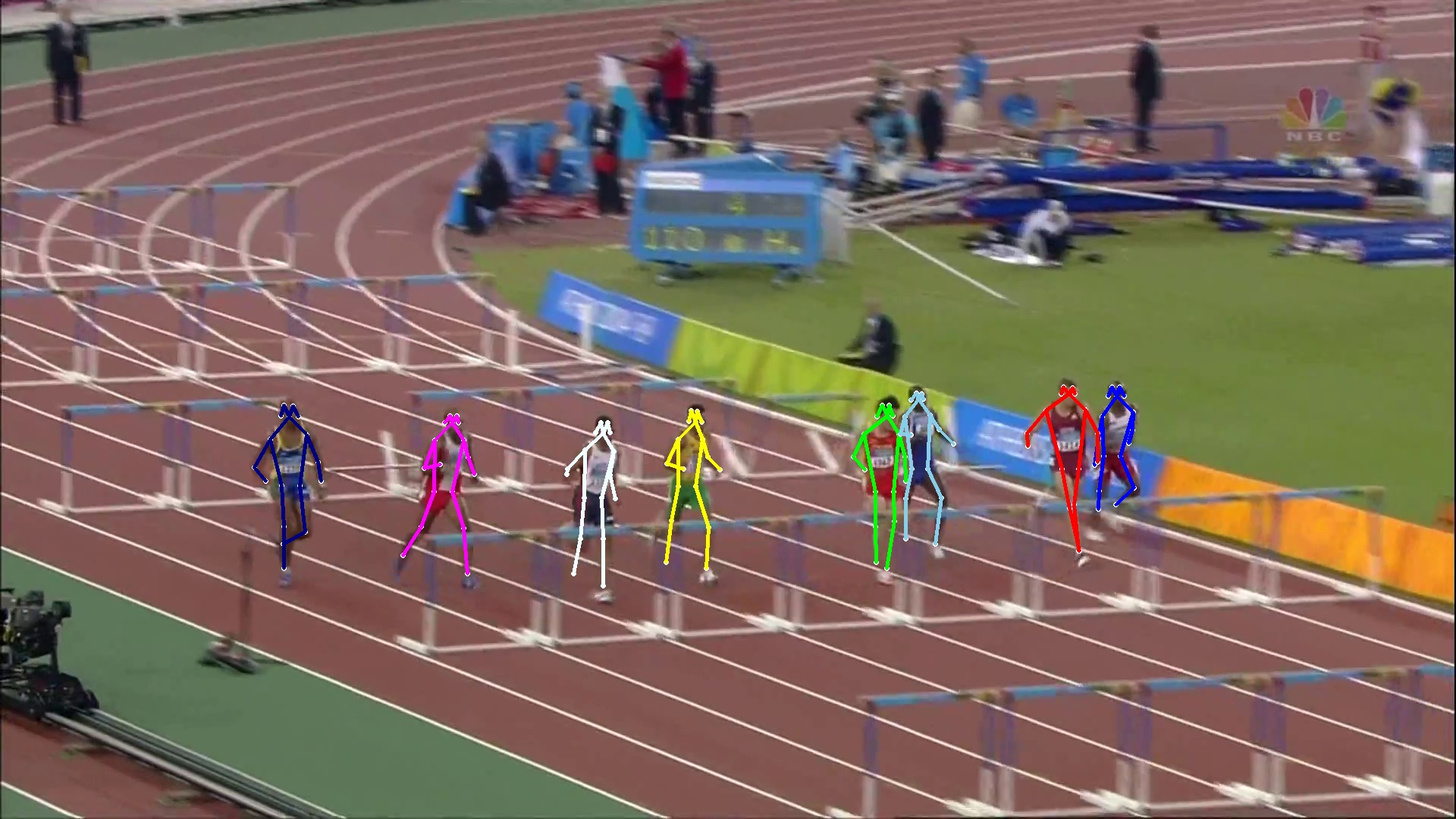} &
\includegraphics[width=0.25\linewidth]{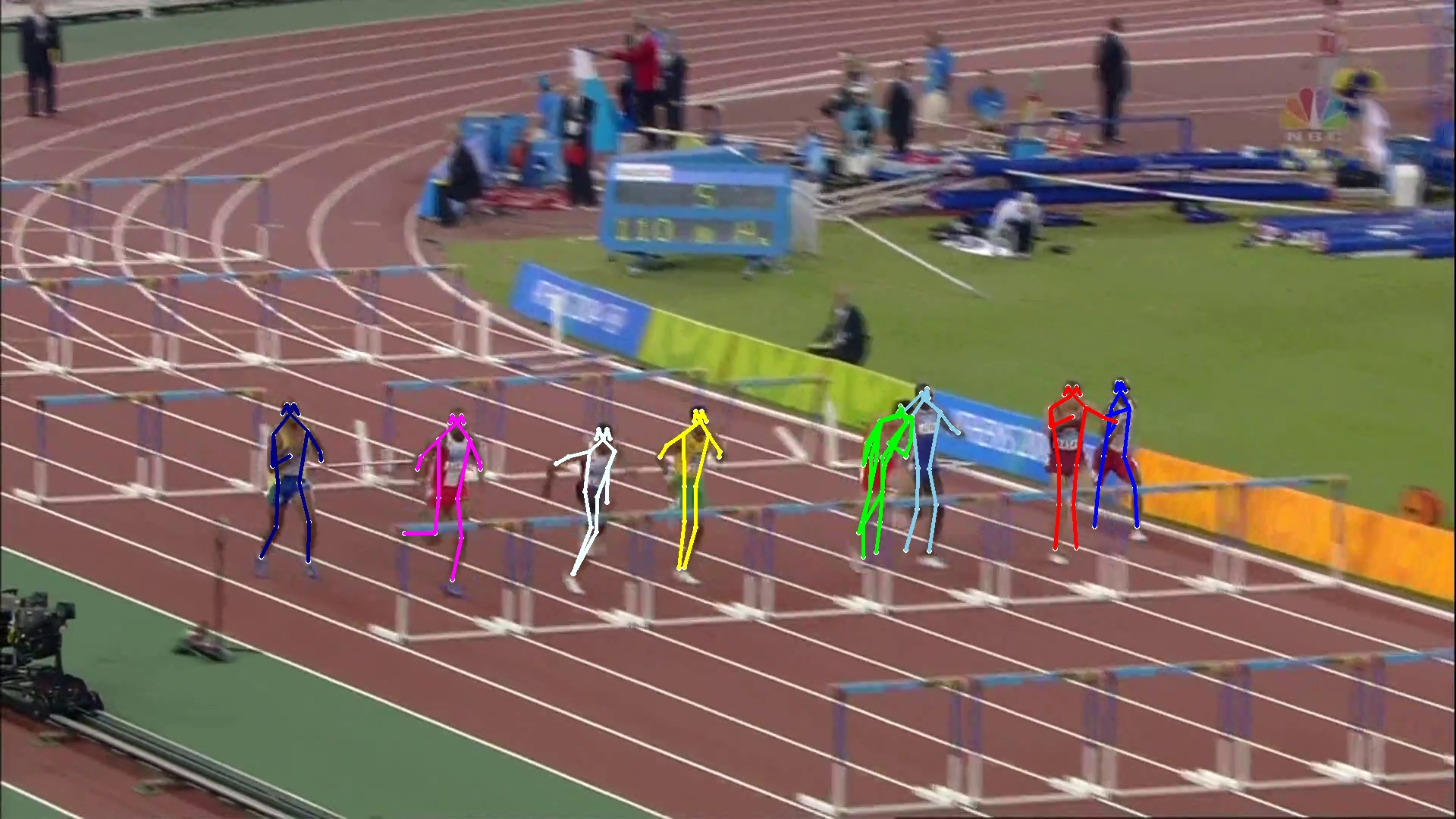} &
\includegraphics[width=0.25\linewidth]{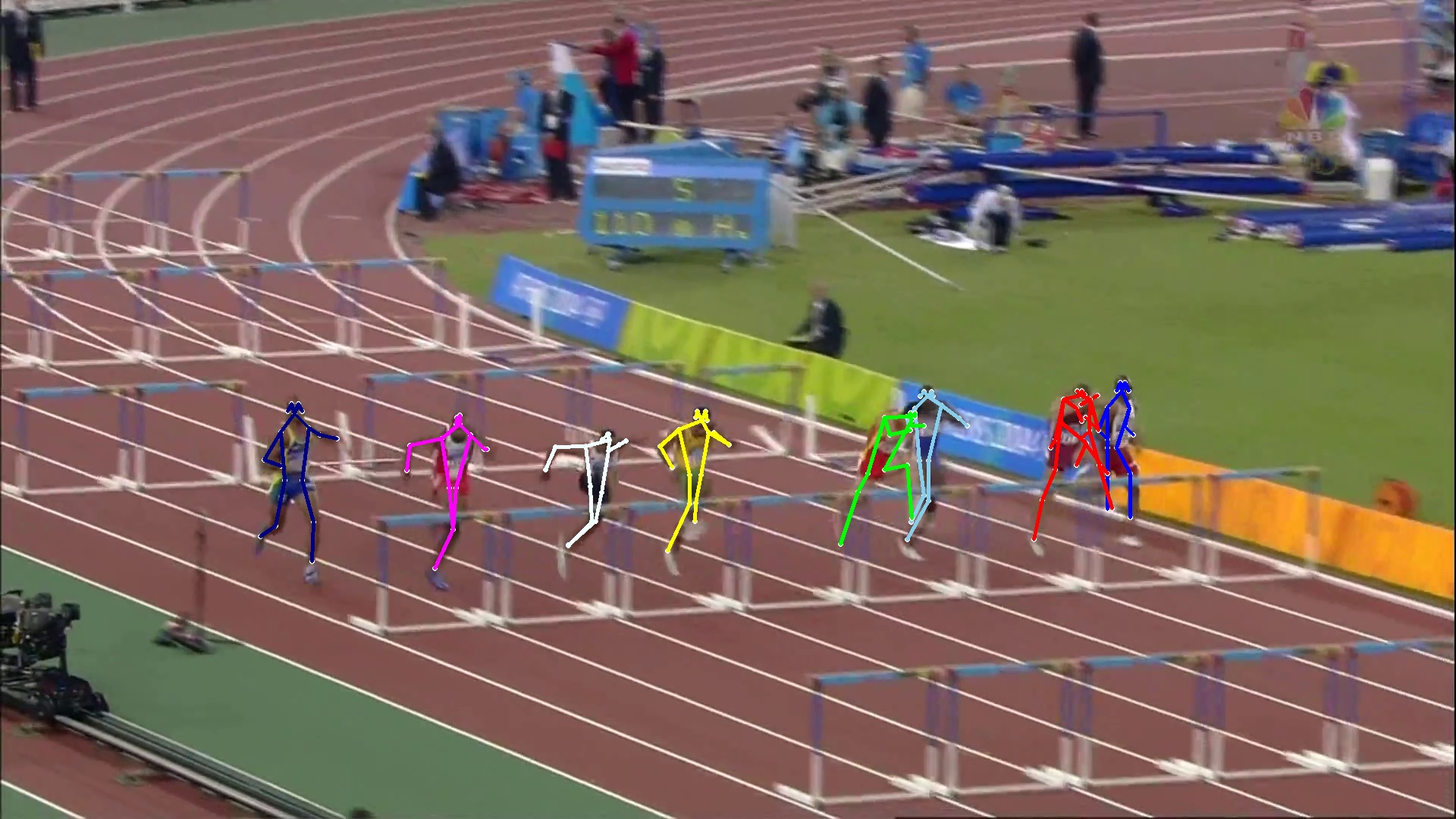} \\
\includegraphics[width=0.25\linewidth]{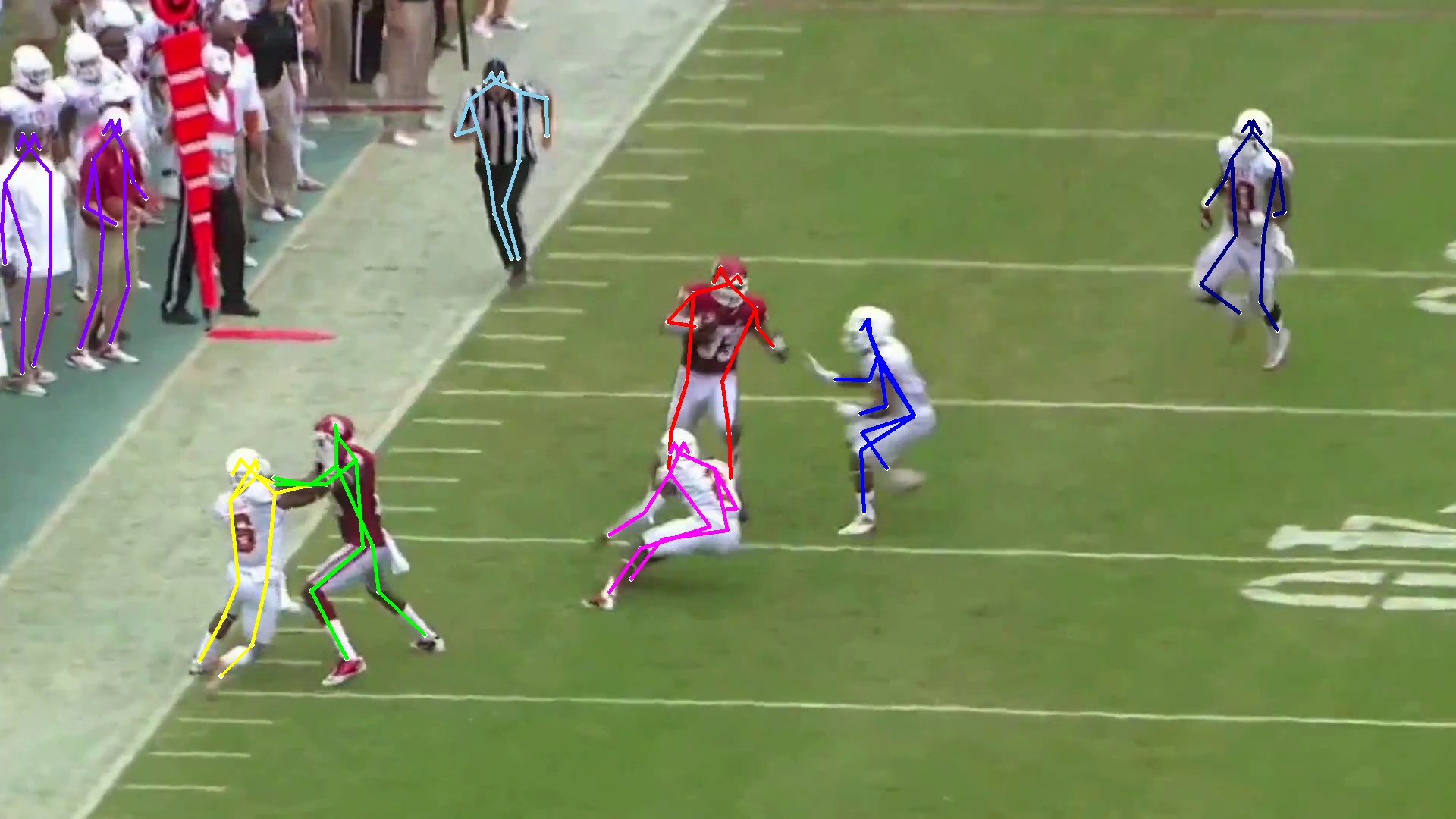} &
\includegraphics[width=0.25\linewidth]{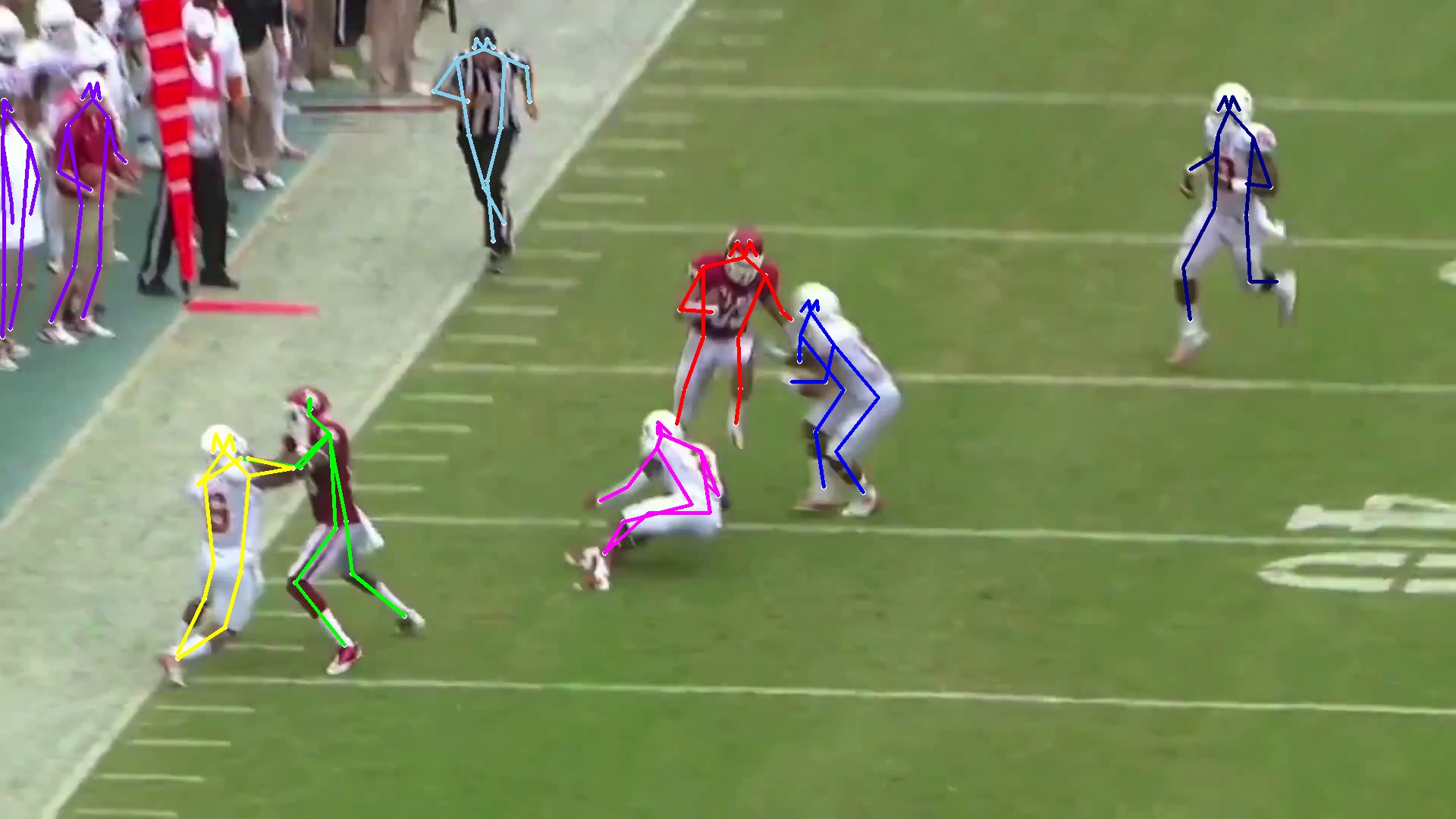} &
\includegraphics[width=0.25\linewidth]{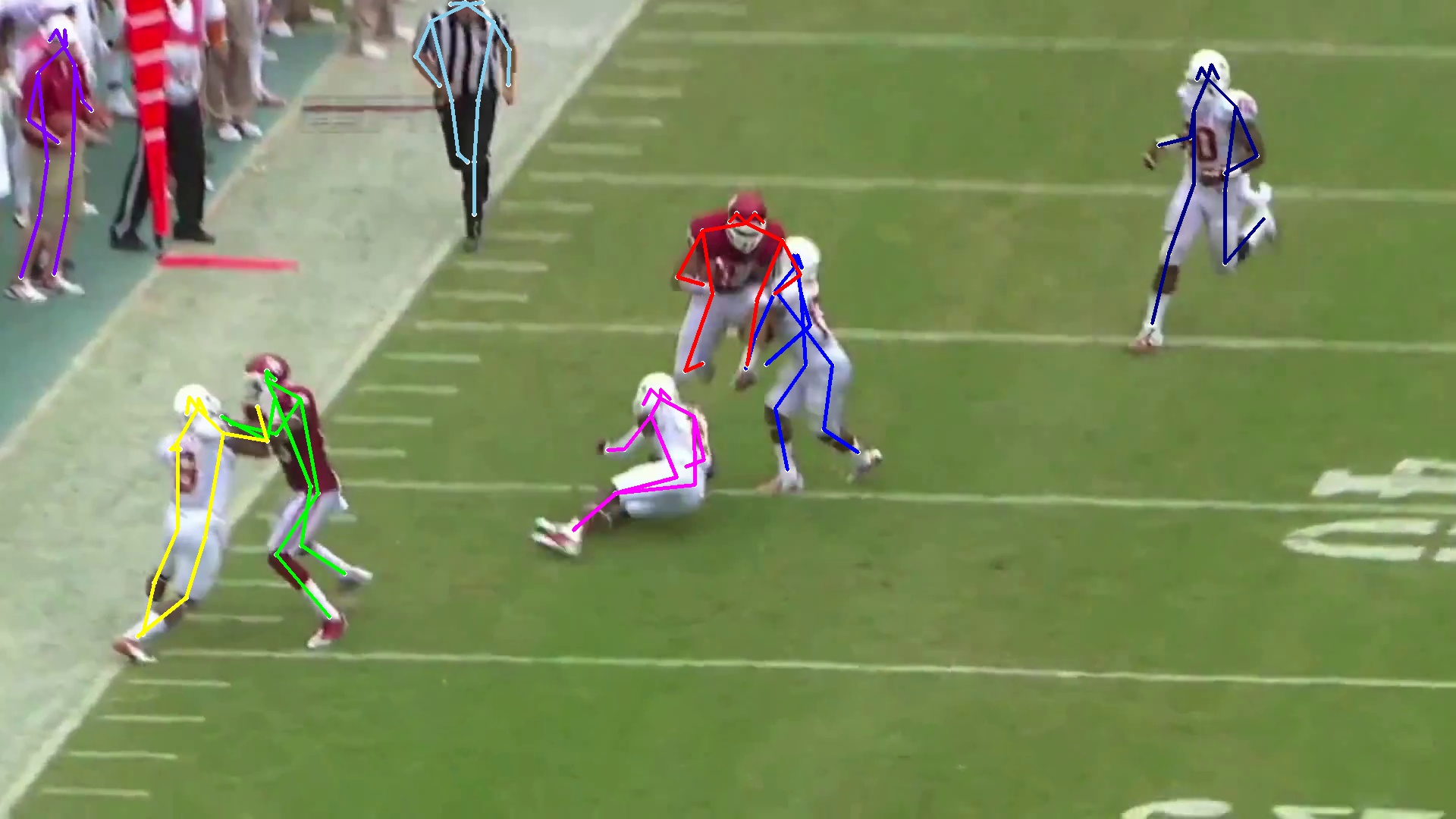} &
\includegraphics[width=0.25\linewidth]{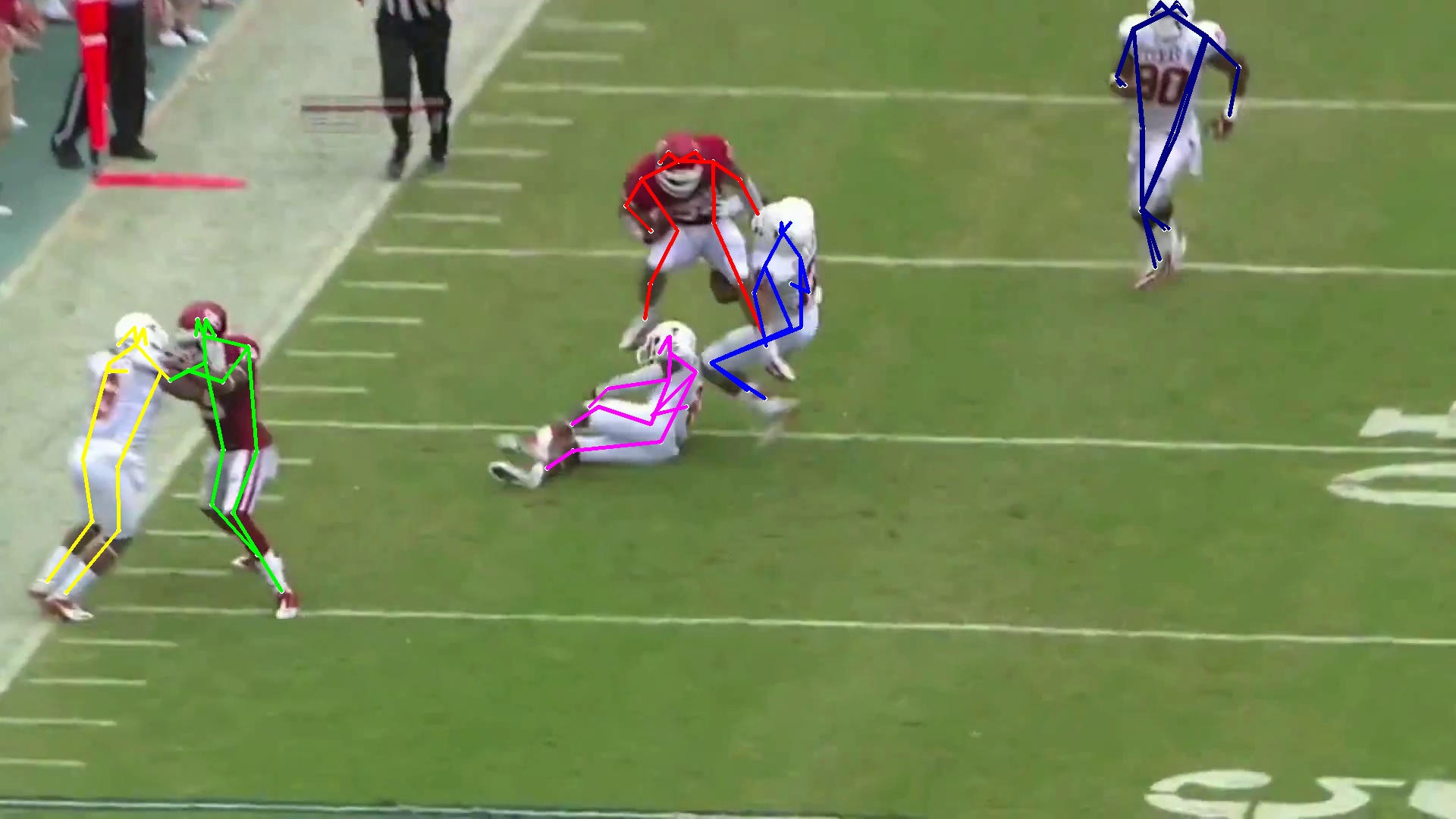} \\
\end{tabular} \egroup
\end{center}\vskip-3mm
\caption{Visualization results of our proposed semi-supervised tracking approach compared to baseline method mentioned in our paper on video instance segmentation and pose tracking. Each row has five sampled frames from a video sequence. Categories and instance masks are shown for each object. Note that objects with the same predicated identity across frames are marked with the same color. Zoom in to see details.}
\vspace{-3mm}
\label{fig:compare}
\end{figure*}